\documentclass{article}

\usepackage[eandd, final]{neurips_2026}

\usepackage[utf8]{inputenc} 
\usepackage[T1]{fontenc}    
\usepackage{hyperref}       
\usepackage{url}            
\usepackage{booktabs}       
\usepackage{amsfonts}       
\usepackage{nicefrac}       
\usepackage{microtype}      
\usepackage{xcolor}         
\usepackage{longtable}
\usepackage{tabularx}
\usepackage{graphicx}
\usepackage{xspace}
\usepackage{multirow}
\usepackage{amsmath}
\usepackage{caption}
\usepackage{enumitem}
\usepackage{wrapfig}
\usepackage[defaultcolor=magenta]{changes}
\usepackage[table]{xcolor}
\usepackage{colortbl}
\usepackage{pgf}
\usepackage{float}
\usepackage[most]{tcolorbox}
\usepackage{titletoc}

\newcommand{\ie}{\textit{i.e.,}\ }
\newcommand{\eg}{\textit{e.g.,}\ } 
\newcommand{\dtname}{\texttt{EHR-ReasonCon}\xspace}
\newcommand{\frname}{\mbox{\textsc{EHR-Inspector}}\xspace}
\newcommand{\equal}[1]{{\hypersetup{linkcolor=black}\thanks{#1}}}

\title{Towards Error-Free EHRs: Reasoning-Intensive Consistency Verification Between Clinical Notes and Structured Tables in Electronic Health Records}

\author{
    Yeonsu Kwon$^{1}$\equal{These authors contributed equally}\;,
    Jiho Kim$^{1}$\footnotemark[1]\;,
    Junseong Choi$^{1}$, 
    Paloma Rabaey$^{2}$, \\
    \textbf{Minseo Kim$^{1}$},
    \textbf{Sujeong Im$^{1}$},
    \textbf{Jeewon Yang$^{1}$}, 
    \textbf{Junmin Lee$^{1}$},
    \textbf{Sangji Lee$^{3}$}, \\ 
    \textbf{Jiwon Kim$^{4}$}, 
    \textbf{Hangyul Yoon$^{1}$},  
    \textbf{Hyunwook Kwon$^{5}$},  
    \textbf{Edward Choi$^{1}$} \\
    $^{1}$KAIST \;
    $^{2}$Ghent University \;
    $^{3}$Samsung Medical Center \; \\
    $^{4}$Samsung Changwon Hospital \;
    $^{5}$Asan Medical Center\\
    \texttt{\{yeonsu.k, jiho.kim, edwardchoi\}@kaist.ac.kr}}

\begin{document}

\maketitle

\begin{abstract}
Data consistency between unstructured clinical notes and structured tables in \mbox{Electronic} Health Records (EHRs) is essential for patient safety and clinical decision-making. However, existing work on note--table consistency verification mainly relies on surface-level matching of numeric values or simple events. Such approaches fail to capture the reasoning underlying real-world EHR documentation, including clinical interpretation, event relations, and temporal changes. To address this gap, we introduce \dtname, a reasoning-intensive benchmark for note--table consistency verification. Built on MIMIC-III with expert-guided annotations, it comprises 8,042 entities derived from clinical notes and provides high-quality ground-truth labels. The annotation protocol is supported by specialized table-exploration tools to ensure systematic evidence retrieval and reliable consistency assessment. We also propose \frname, an LLM-based framework that segments notes, extracts anchor entities and temporal references, and uses table-exploration tools to verify consistency against structured tables. Evaluated using expert-validated LLM-as-a-judge metrics under harsh and lenient criteria, \frname\ achieves state-of-the-art performance across multiple model backbones. Analyses further demonstrate the effectiveness of its components and highlight differences from human verification. The code and dataset are available at \url{https://github.com/dustn1259/EHR-ReasonCon}.
\end{abstract}

\section{Introduction}
\label{intro}
\begin{figure*}[t]
    \centering
    \includegraphics[width=\textwidth]{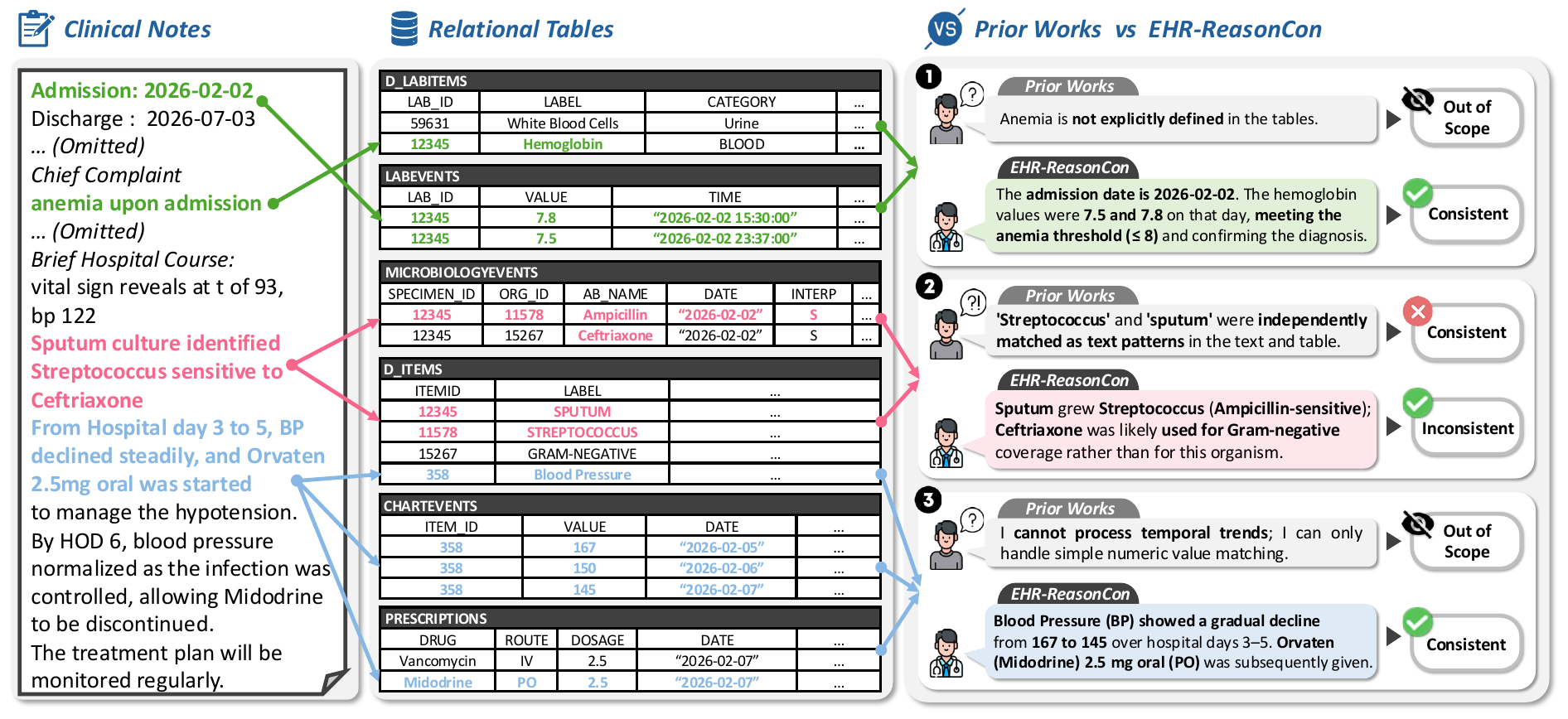}
    \vspace{-4mm}
    \caption{Overview of reasoning-intensive note–table consistency verification. The examples highlight the need for reasoning-intensive verification that goes beyond surface-level alignment between clinical notes and structured tables.
}
    \vspace{-3mm}
    \label{fig:overall_concept}
\end{figure*}

Within Electronic Health Record (EHR) systems, patient information is documented through two primary modalities: structured tables (\eg vital signs, prescriptions) and unstructured clinical notes (\eg physician notes)~\cite{Seinen2024UsingSC}. Clinicians combine objective findings from structured data with contextual information from notes to guide diagnostic and therapeutic decisions, which are then recorded back into the EHR. 
Due to this interdependence, reliability and consistency between these data types are critical. However, in practice, discrepancies frequently arise from administratively driven system architectures and documentation practices~\cite{payne2018using,7001812}, potentially compromising patient safety and creating legal risks~\cite{demsash2023documentation,tsou2017copypaste}.

Detecting these discrepancies is therefore essential, yet manual verification is impractical due to its time and cost demands. This has motivated the development of automated approaches. Prior work has investigated inconsistencies between clinical notes and structured tables, primarily focusing on specific domains such as allergies or medications~\cite{Li2015EndToEndMedDiscrepancy, Lo2022Reconciling, Rinott2012Sarcoma}. More recently, EHRCon~\cite{kwon2024ehrcon} extended this line of work to relational databases. However, these approaches rely on \textit{surface-level} verification, such as checking whether numerical values (\eg WBC 10.0) or discrete events (\eg administration of vancomycin) mentioned in clinical notes are also recorded in structured tables. 
While such approaches provide a useful starting point, they often fail to capture the contextual and nuanced nature of real-world clinical documentation.

Real-world EHR documentation fundamentally requires advanced reasoning for accurate note–table consistency verification beyond surface-level alignment. 
For example, clinical notes often describe interpreted patient states, whereas structured tables record the underlying measurements (see Figure~\ref{fig:overall_concept}-(1))~\cite{Gao2024,Raghavan2014Unstructured}. Verifying these statements therefore requires assessing whether the measurements satisfy the clinical criteria supporting the interpretation. Moreover, clinical notes often describe relationships among multiple clinical events (see Figure~\ref{fig:overall_concept}-(2)). Verifying such statements requires checking whether these event relationships are consistently supported by structured records, rather than validating individual table entries in isolation~\cite{khetan2022mimicause,WANG201834}. Furthermore, clinical notes often describe changes in patient status over time and subsequent interventions (see Figure~\ref{fig:overall_concept}-(3)). Verifying such statements requires assessing trends, time spans, and corresponding treatments rather than relying on a single time point in structured tables~\cite{pan2020,Yu2024Temporal}. However, existing approaches fall short in capturing these reasoning-intensive aspects of note–table consistency.

To bridge this critical gap, we introduce \dtname, a reasoning-intensive benchmark for note–table consistency verification. Built on MIMIC-III~\cite{PhysioNet-mimiciii-1.4}, the dataset comprises 8,042 entities derived from clinical notes. It is designed to reflect real-world clinical documentation practices, guided by a rigorous annotation protocol developed in close collaboration with four clinical experts\footnote{Board-Certified Radiation Oncologist, Board-Certified General Surgeon, Resident in Anesthesiology, EHR technician}. This protocol was iteratively refined through pilot studies to ensure both clarity and robustness. 
To support annotation under this protocol, we developed eight specialized table-exploration tools for systematic retrieval of structured evidence. Using these tools, eight annotators familiar with EHR systems carried out the annotation, consulting authoritative medical references as needed.
To ensure high reliability, we implement a multi-stage quality control process, including dual annotation, disagreement resolution, and final adjudication by physicians. As a result, \dtname\ achieves a high level of inter-annotator agreement (\ie 0.897 for NER, 0.888 for consistency labeling), establishing a reliable ground truth for evaluating reasoning-based consistency verification.

To address this benchmark, we also propose \frname, an LLM-based framework that mirrors the annotation workflow. \frname segments clinical notes into event-centric spans, extracts anchor entities along with their temporal context, and verifies consistency with structured EHR data using table-exploration tools. For evaluation, we adopt LLM-as-a-judge evaluators, validated against clinical expert judgments, and assess the framework under two strictness levels: Harsh and Lenient. Experimental results show that \frname consistently achieves state-of-the-art performance across multiple model backbones, substantially improving recall and precision. Extensive ablation analyses were conducted, and further reasoning-trace evaluations highlighted the differences between LLM and human verification.

\section{Related Works}
\noindent\textbf{Consistency Check Between Clinical Notes and Structured Tables in EHRs}~~~
Discrepancies between clinical notes and tables have long been recognized as a critical issue that can lead to medical errors~\cite{kwon2024ehrcon, Li2015EndToEndMedDiscrepancy, Lo2022Reconciling, Rinott2012Sarcoma}. Early studies on consistency checking primarily focused on reconciliation within specific domains to align information across disparate data sources. For example, \cite{Rinott2012Sarcoma} detected inconsistencies in sarcoma discharge summaries using an ensemble of classifiers, \cite{Li2015EndToEndMedDiscrepancy} proposed a hybrid ML and rule-based approach for pediatric medication discrepancies, and \cite{Lo2022Reconciling} applied NLP methods to reconcile allergy information between clinical notes and structured lists. However, these approaches typically relied on extracting coded entities from notes and comparing them with structured tables, without defining consistency verification as a general task or releasing datasets. To address this limitation, EHRCon~\cite{kwon2024ehrcon} introduced a benchmark for verifying consistency between clinical notes and relational databases, constructed on MIMIC-III~\cite{PhysioNet-mimiciii-1.4}. The dataset includes manual annotations linking entities in clinical notes to entries in multiple tables via SQL query execution. However, EHRCon performs verification in a \textit{surface-level} manner, assessing whether specific values or simple events in notes match structured records. In contrast, \dtname introduces a \textit{reasoning-intensive} benchmark for assessing note–table consistency.

\noindent\textbf{Tool-Augmented Table Reasoning Agents}~~~
Recent work on table QA augments LLM reasoning with external tools for filtering, aggregation, and numerical computation over tabular data~\cite{jiang2025tablemind, lu-etal-2025-tart, wang2025sheetbrain,  xiong2025tablezoomer, zhou2025mixture}. Some approaches integrate program generation and execution into the reasoning process~\cite{lu-etal-2025-tart}, while others develop autonomous agents that perform iterative planning, action, and reflection~\cite{jiang2025tablemind}, or organize these capabilities into modular or multi-agent workflows for complex table reasoning~\cite{xiong2025tablezoomer, zhou2025mixture}. 
Spreadsheet agents further extend tool-based reasoning to large multi-table environments and support both question answering and spreadsheet manipulation~\cite{wang2025sheetbrain}. 
Inspired by these advances, we propose \frname, a tool-augmented framework for reasoning-intensive consistency verification between clinical notes and large-scale structured tables.

\section{EHR-ReasonCon}
\label{sec3}

\begin{minipage}[t]{0.47\textwidth}
\dtname is a high-quality benchmark for reasoning-intensive consistency verification, comprising 8,042 annotated entities linked to 14 tables in MIMIC-III.\footnotemark These entities are derived from 105 clinical notes across three note types: discharge summaries, physician notes, and nursing notes. Table~\ref{tab:ehrcon2-stats} reports statistics for \dtname, with additional details in Appendix~\ref{app:dataset_details}. Figure~\ref{fig:annotation} shows the annotation process, and the steps involved are described below.
\end{minipage}
\hfill
\begin{minipage}[t]{0.51\textwidth}
\centering
\small
\vspace{-2mm}
\captionof{table}{Dataset statistics of \dtname. Con. and Incon. represent the number of entities with consistent and inconsistent labels, respectively.}
\label{tab:ehrcon2-stats}
\resizebox{\textwidth}{!}{%
\begin{tabular}{lcccccc}
\toprule
\multirow{2}{*}{Note Type} & \multicolumn{2}{c}{Entity} & \multicolumn{2}{c}{Labels} & \multicolumn{2}{c}{Note} \\
\cmidrule(lr){2-3} \cmidrule(lr){4-5} \cmidrule(lr){6-7}
& Mean & Total & Con. & Incon. & Total & Mean Length \\
\midrule
Discharge & 92.03 & 3,497 & 2,041 & 1,456 & 38 & 2,789 \\
Physician & 86.09 & 2,841 & 2,388 & 453 & 33 & 1,859 \\
Nursing & 50.12 & 1,704 & 1,418 & 286 & 34 & 1,111 \\
\midrule
Total & 76.59 & 8,042 & 5,847 & 2,195 & 105 & 1,953 \\
\bottomrule
\end{tabular}%
}
\end{minipage}

\footnotetext{Tables include Chartevents, Labevents, Prescriptions, Inputevents\_cv, Inputevents\_mv, Outputevents, Procedureevents\_mv, Microbiologyevents, Diagnoses\_icd, Procedures\_icd, D\_items, D\_icd\_diagnoses, D\_icd\_procedures, and D\_labitems. Although MIMIC-IV~\cite{johnson2020mimic} is newer, we use MIMIC-III because MIMIC-IV lacks diverse note types (\eg physician and nursing notes) and removes note dates instead of shifting them.}




\begin{figure}[t]
    \centering
    \includegraphics[width=\linewidth]{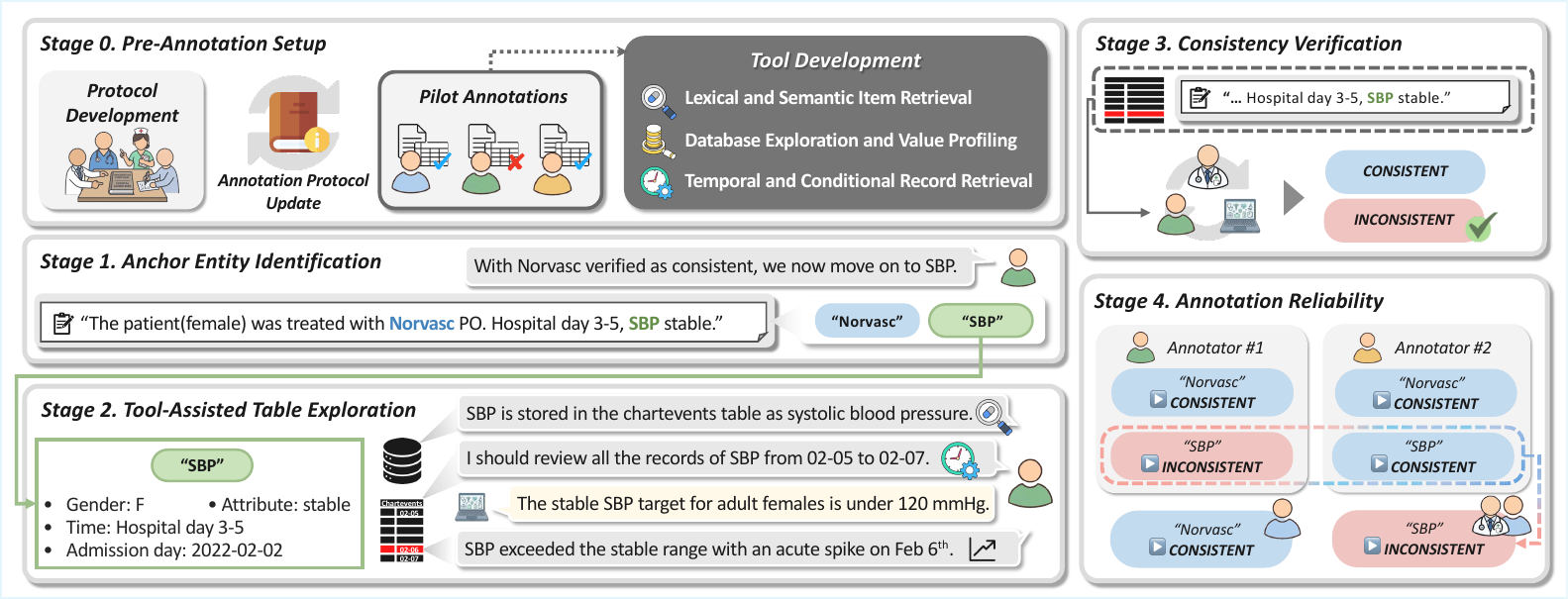}
    \vspace{-4mm}
    \caption{Overview of the reasoning-intensive annotation process for note–table consistency verification. The pipeline includes protocol and tool development (Stage 0), anchor entity identification (Stage 1), tool-assisted evidence retrieval (Stage 2), and entity-level consistency verification (Stage 3). Finally, a dataset-level quality control step (Stage 4) is performed to resolve disagreements and refine annotations after all entities are annotated.}
    \label{fig:annotation}
\end{figure}

\paragraph{Stage 0: Pre-Annotation Setup}
The goal of this stage is to establish the annotation protocol and tools needed to construct a high-quality benchmark reflecting clinical contexts. The protocol, developed with medical practitioners, specifies how narrative expressions in clinical notes are mapped to structured table fields and provides criteria for interpreting temporal trends, handling ambiguous clinical judgments, and ensuring annotation consistency (see Appendix~\ref{app:annotation_protocol} for detailed protocol). However, real-world EHR data contain edge cases not fully anticipated by predefined guidelines. To address this, we conducted a pilot annotation study to refine the protocol. During the pilot phase, we analyzed how annotators searched for relevant evidence in structured tables and formalized recurring search patterns into modular functionalities. This process produced eight table-exploration tools (see Appendix~\ref{app:table_exploration_tools} for further details) that support efficient exploration of complex structured tables and are organized into three functional categories:

\begin{itemize}[leftmargin=1.3em, itemsep=2pt]
\item \textbf{Entity-to-Table Item Alignment}~~
The tools in this category support the alignment of entities mentioned in clinical notes with corresponding items in structured tables. The same clinical information may appear under different names or levels of abstraction (\eg ``White Blood Cells'' and ``WBC''), so these tools retrieve potentially relevant table items based on both lexical similarity and conceptual relatedness.

\item \textbf{Database Exploration and Value Profiling}~~
The tools in this category support exploration of the EHR database schema and content. Since clinical concepts may be distributed across multiple tables, the tools support exploration of relevant table groups and summarize typical values for each item (\eg \textit{Stool Amount: [Small, Medium, Large]} indicates that \textit{Stool Amount} is categorical rather than numerical), enabling annotators to quickly interpret the role of different fields.

\item \textbf{Temporal and Conditional Record Retrieval}~~
The tools in this category support verification of clinical statements that involve temporal changes or specific conditions. The tools allow annotators to retrieve records from structured tables based on time windows and value constraints, enabling inspection of whether the structured data support trends or events described in clinical notes.
\end{itemize}


\textbf{Stage 1: Anchor Entity Identification}~~~
The goal of this stage is to identify anchor entities in clinical notes that correspond to items in structured tables. These anchors serve as entry points for exploring records in tables such as medications and diagnoses. 

\textbf{Stage 2: Tool-Assisted Table Exploration}~~~
The goal of this stage is to retrieve evidence from structured tables corresponding to the anchor entities. Annotators review the anchor entities, their attributes (\eg dosage, route), and their temporal context to understand the clinical trends described in the notes. They then use the predefined tools (in \textbf{Stage 0}) to query structured tables and retrieve records that support the clinical narratives, which are then used as evidence for consistency assessment.



\textbf{Stage 3: Consistency Verification}~~~
The goal of this stage is to verify whether the narrative content associated with each anchor entity is supported by the structured evidence. 
Anchor entities are labeled as \mbox{\textsc{Consistent}} if the corresponding information in the clinical notes is supported by structured records; otherwise, they are labeled as \mbox{\textsc{Inconsistent}}. To ensure reliable labeling, annotators perform reasoning-intensive analysis involving temporal reasoning, commonsense reasoning, and medical interpretation. When necessary, they consult established medical references\footnote{UpToDate, MedlinePlus, Cleveland Clinic, Mayo Clinic} or physicians.

\textbf{Stage 4: Annotation Reliability}~~~
The goal of this stage is to ensure the integrity and reliability of the dataset through a multi-step verification process. The annotation was conducted by eight trained annotators. For each clinical note, two annotators were randomly assigned to independently perform the annotation and then resolve disagreements through mutual reconciliation. For complex clinical judgments, annotators consulted medical professionals. After the initial annotations were completed, an independent reviewer conducted a review of all 105 clinical notes to ensure dataset-wide consistency. The inter-annotator agreement for NER and consistency labeling was 0.897 and 0.888, respectively (see Appendix~\ref{app:IAA} for details).

\begin{figure*}[t]
    \centering
    \includegraphics[width=\textwidth]{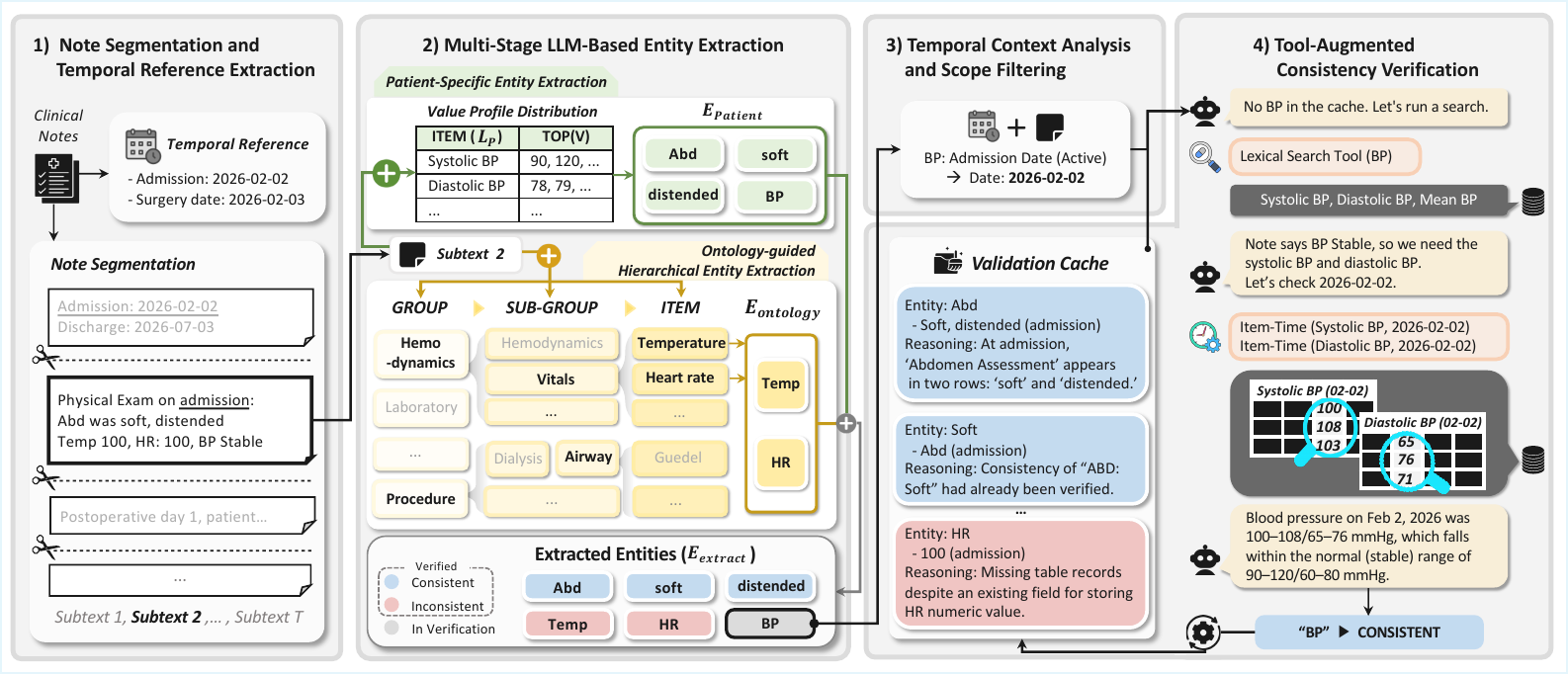}
    \caption{Overview of \frname. The framework (1) segments clinical notes and resolves temporal references, (2) extracts anchor entities using patient records and ontology guidance, (3) filters entities based on temporal context, and (4) verifies note–table consistency using tool-augmented table exploration.}
    \vspace{-2mm}
    \label{fig:ehrinspector}
\end{figure*}

\section{\textsc{EHR-Inspector}}
\label{sec4}
We propose \frname, an LLM-based framework for verifying consistency between clinical notes and structured tables that mirrors the annotation workflow. Figure~\ref{fig:ehrinspector} provides an overview of \frname, operating through a sequence of modules described in the following subsections. The implementation details, including the prompts, are provided in Appendix~\ref{app:detail_framework}.

\subsection{Note Segmentation and Temporal Reference Extraction}
\label{sec4.1}
Clinical notes often contain multiple clinical events within a single document, which can hinder LLM reasoning when the entire note is processed as a single context. To address this, we first perform LLM-based note segmentation. Given a clinical note $x$, the LLM partitions it into topic-coherent segments $X=\{x_1, x_2, ..., x_T\}$, enabling localized reasoning in each segment.
However, segmentation may separate relative temporal expressions (\eg Postoperative Day 3) from their reference events (\eg Surgery Date), which may appear in different segments. To maintain consistent temporal interpretation, the LLM extracts global temporal references from the entire note and resolves relative temporal expressions within segments into absolute time.

\subsection{Multi-Stage LLM-Based Entity Extraction}
\label{sec4.2}
This module aims to extract only note entities that can be aligned with structured table records, rather than all clinical mentions. A simple approach would prompt the LLM with the entire structured item set $L$ and ask it to identify corresponding expressions in the note. However, since $L$ contains over 14,000 items, this is computationally inefficient and may degrade performance. To address this issue, we adopt a two-step extraction strategy to comprehensively identify note entities that can be aligned with structured table items. First, we perform patient-specific extraction using \(L_P \subseteq L\), the subset of structured items recorded for patient \(P\). Given a note segment and \(L_P\), the LLM extracts entities aligned with the patient’s structured records, denoted as \(E_{\text{patient}}\). Second, to capture entities that may be absent from the structured records of $P$ but are valid within the broader EHR ontology, we perform ontology-guided coarse-to-fine extraction. Instead of presenting the large complement \(L \setminus L_P\), the LLM first selects relevant high-level groups and subgroups, and then extracts entities within the selected subgroups, yielding \(E_{\text{ontology}}\). Finally, we obtain the extractable entity set by merging the outputs of the two stages: $E_{\text{extract}} = E_{\text{patient}} \cup E_{\text{ontology}}$. Details are provided below.

\noindent\textbf{Patient-specific Entity Extraction}~~~
In this stage, the LLM identifies note entities corresponding to structured items recorded for patient $P$.
For each note segment $x_t \in X$, the LLM is provided with: (1) the note segment $x_t$, (2) the patient-specific item set $L_P$, and (3) a value distribution profile $TopV(l)$ for each item $l \in L_P$. The profile $TopV(l)$ is defined as the top-$K$ most frequent values of $l$ in the overall structured tables, and serves as an auxiliary signal that helps the LLM better understand the semantic meaning and value type of each item. For example, given $TopV(SBP) = [\textit{110, 120, 130}]$, the LLM can infer from the distribution and format of these values that the item \textit{SBP} is numerical, rather than a categorical or textual attribute. Based on these inputs, the LLM extracts entity names in each segment $x_t$ that correspond to items in $L_P$, and aggregates them across segments to form $E_{\mathrm{patient}} = \bigcup_{x_t \in X} E^{\mathrm{pat}}(x_t)$.

\noindent\textbf{Ontology-guided Hierarchical Entity Extraction}~~~
This stage targets entities outside $L_P$ by narrowing the item set ($L \setminus L_P$) through a pre-constructed \textsc{\mbox{hierarchical} \mbox{ontology}} $O$.
The ontology consists of 13 high-level groups, $O=\{G_1,\dots,G_{13}\}$. Each group $G_i$ (\eg \textit{Hemodynamics}) is divided into subgroups $G_i=\{S_{i,j}\}_{j=1}^{J_i}$, and each subgroup $S_{i,j}$ (\eg \textit{Vitals}) is associated with a set of items $L_{i,j}$ (\eg \{\textit{Heart rate, SpO2, ...}\}). Using the ontology as a coarse-to-fine search space, for each note segment $x_t$, the LLM first identifies relevant groups $G(x_t) \subseteq O$ and subsequently selects subgroups $S(x_t) \subseteq \bigcup_{G_i \in G(x_t)} G_i$.  
The LLM then extracts entities $E^{\mathrm{ont}}(x_t)$ that correspond to the items in $L(x_t)=\bigl(\bigcup_{S_{i,j}\in S(x_t)} L_{i,j}\bigr)\setminus L_P$.
The final ontology-based entity set is obtained by aggregating entities across all segments $x_t \in X$: $E_{\mathrm{ontology}}=\bigcup_{x_t \in X} E^{\mathrm{ont}}(x_t)$. Details on the ontology construction and the full list of groups and subgroups are provided in Appendix~\ref{app:ontology_guided}.

\subsection{Temporal Reasoning and Scope Filtering}
\label{sec4.3}
Structured tables in EHRs typically contain events from the current hospitalization, whereas narrative clinical notes may additionally contain past history or future plans.
To avoid incorrectly flagging past or planned entities as inconsistencies, we retain only entities associated with the current hospitalization from $E_{extract}$. For each entity $e \in E_{extract}$, an LLM extracts temporal cues (\eg occurrence time, duration, end time) and assigns a temporal status $C(e) \in \{{\text{Past}, \text{Active}, \text{Plan}}\}$ using these cues and the global temporal references identified in \S\ref{sec4.1}. The scope-filtered set is $E_{scope} = \{ e \in E_{extract} \mid C(e) = \text{Active} \}$, used for consistency verification.



\subsection{Tool-Augmented Consistency Verification}
\label{sec4.4}
Structured tables in EHRs often contain high-frequency time-series records, making it infeasible to include all relevant structured records in the limited LLM context. To address this, we enable the LLM to interact with the tables through table-exploration tools (defined in \S\ref{sec3}) that retrieve relevant subsets of the data. These tools also expose schema and value representations, allowing the LLM to interpret how clinical concepts are recorded in structured form and perform accurate verification for each entity $e \in E_{scope}$. However, verifying each entity independently can be computationally inefficient, as multiple entities extracted from the same segment $x_t$ often originate from a single clinical observation. For example, the entities `\textit{Abd}', `\textit{soft}', and `\textit{distended}' all refer to the same abdominal exam in the sentence ``\textit{Abd: soft, distended.}'' In such cases, independent verification would lead to redundant tool calls and repeated access to the same table entries. To tackle this issue, we introduce a \textsc{validation cache} that stores the $m$ most recently validated entities and their evidence in a sliding window, enabling reuse for related entities. Finally, each entity $e$ is assigned a \textsc{consistent} or \textsc{inconsistent} label based on its alignment with the structured records.

\section{Experiments}
\subsection{Experimental Setting}
We partitioned the dataset into test and validation sets with a 4:1 ratio. The test set was used for the main experiments, while the validation set was used to develop \frname.


\textbf{Baseline}~~~
We compare \frname with \mbox{CheckEHR}~\cite{kwon2024ehrcon}, the state-of-the-art framework for verifying consistency between clinical notes and structured tables. CheckEHR is an LLM-based eight-stage pipeline that extracts entities from clinical notes and generates SQL queries to validate them against tables. To the best of our knowledge, CheckEHR is the only prior framework that addresses this comparable note--table verification setting, and thus serves as our primary baseline.\footnote{General-purpose table reasoning agents~\cite{jiang2025tablemind,lu-etal-2025-tart} are not directly applicable to our task because they assume an explicit question or claim and a predefined table context. In contrast, our task requires inferring verification units, temporal scope, and supporting or missing evidence from long clinical notes, and then grounding them across relational tables.}


\textbf{Base LLMs}~~~
To ensure a fair comparison, we use four LLMs as base models for both \mbox{CheckEHR} and \frname: Gemini 2.5 Flash~\cite{comanici2025gemini}, a proprietary model known for strong reasoning performance with high cost efficiency; Qwen3-32B~\cite{yang2025qwen3} and GPT-OSS 20B~\cite{agarwal2025gpt}, open-source reasoning models; and MedGemma 27B~\cite{sellergren2025medgemma}, an open-source model specialized for medical-domain tasks.

\definecolor{pastelorange}{RGB}{255,180,140}
\definecolor{pastelblue}{RGB}{160,200,235}

\newcommand{\gradientsetup}[3]{%
  \def\gradcenter{#1}%
  \def\gradgain{#2}%
  \def\gradcap{#3}%
}

\gradientsetup{40}{2.3}{80}

\newcommand{\gradcell}[1]{%
  \pgfmathparse{int(min(\gradcap, round(abs(#1-\gradcenter)*\gradgain)))}%
  \edef\__pct{\pgfmathresult}%
  \ifdim #1pt < \gradcenter pt
    \edef\__col{pastelorange!\__pct}%
  \else
    \edef\__col{pastelblue!\__pct}%
  \fi
  \expandafter\cellcolor\expandafter{\__col}#1%
}

\newcommand{\gradcellfmt}[2]{%
  \pgfmathparse{int(min(\gradcap, round(abs(#1-\gradcenter)*\gradgain)))}%
  \edef\__pct{\pgfmathresult}%
  \ifdim #1pt < \gradcenter pt
    \edef\__col{pastelorange!\__pct}%
  \else
    \edef\__col{pastelblue!\__pct}%
  \fi
  \expandafter\cellcolor\expandafter{\__col}#2%
}

\begin{table*}[t]
\centering
\footnotesize
\setlength{\tabcolsep}{5pt}
\caption{Performance comparison between CheckEHR and \frname. Results are reported as recall (Rec) and precision (Prec) under both Lenient and Harsh evaluation.\protect\footnotemark Values below 40 are shaded orange and above 40 blue, with intensity indicating distance from 40.}
\vspace{-1mm}
\renewcommand{\arraystretch}{1.1}

\resizebox{\textwidth}{!}{
\begin{tabular}{llcccccccccccc}
\toprule
\multirow{4}{*}{\textbf{Base LLM}} & \multirow{4}{*}{\textbf{Method}} &
\multicolumn{6}{c}{\textbf{Lenient}} &
\multicolumn{6}{c}{\textbf{Harsh}} \\
\cmidrule(lr){3-8} \cmidrule(lr){9-14}

 & &
\multicolumn{2}{c}{\textbf{Discharge}} &
\multicolumn{2}{c}{\textbf{Physician}} &
\multicolumn{2}{c}{\textbf{Nursing}} &
\multicolumn{2}{c}{\textbf{Discharge}} &
\multicolumn{2}{c}{\textbf{Physician}} &
\multicolumn{2}{c}{\textbf{Nursing}} \\

\cmidrule(lr){3-4} \cmidrule(lr){5-6} \cmidrule(lr){7-8}
\cmidrule(lr){9-10} \cmidrule(lr){11-12} \cmidrule(lr){13-14}

 & & Rec & Prec & Rec & Prec & Rec & Prec & Rec & Prec & Rec & Prec & Rec & Prec \\

\midrule

\multirow{2}{*}{GPT-OSS 20B}
& CheckEHR
& \gradcell{29.40} & \gradcell{60.19}
& \gradcell{24.91} & \gradcell{39.23}
& \gradcell{24.00} & \gradcell{55.85}
& \gradcell{26.50} & \gradcell{50.70}
& \gradcell{23.00} & \gradcell{35.30}
& \gradcell{22.20} & \gradcell{47.20} \\
& \frname
& \gradcell{30.00} & \gradcell{59.12}
& \gradcell{27.53} & \gradcell{55.01}
& \gradcell{28.38} & \gradcell{61.40}
& \gradcell{26.99} & \gradcell{52.45}
& \gradcell{22.17} & \gradcell{49.33}
& \gradcell{24.49} & \gradcell{51.77} \\

\midrule

\multirow{2}{*}{MedGemma 27B}
& CheckEHR
& \gradcell{17.95} & \gradcell{55.19}
& \gradcell{6.60}  & \gradcell{35.09}
& \gradcell{5.56}  & \gradcell{32.54}
& \gradcell{16.41} & \gradcell{43.23}
& \gradcell{6.10}  & \gradcell{32.71}
& \gradcell{4.80}  & \gradcell{26.09} \\
& \frname
& \gradcell{48.68} & \gradcell{44.99}
& \gradcell{28.63} & \gradcell{33.44}
& \gradcell{50.31} & \gradcell{47.46}
& \gradcell{42.74} & \gradcell{39.51}
& \gradcell{26.93} & \gradcell{30.55}
& \gradcell{45.57} & \gradcell{41.50} \\

\midrule

\multirow{2}{*}{Qwen3 32B}
& CheckEHR
& \gradcell{26.05} & \gradcell{62.57}
& \gradcell{27.55} & \gradcell{48.64}
& \gradcell{24.05} & \gradcell{58.10}
& \gradcell{23.04} & \gradcell{54.02}
& \gradcell{25.31} & \gradcell{43.95}
& \gradcell{22.12} & \gradcell{52.14} \\
& \frname
& \gradcell{56.18} & \gradcell{65.29}
& \gradcell{55.34} & \gradcell{65.21}
& \gradcell{60.36} & \gradcell{62.88}
& \gradcell{50.89} & \gradcell{56.12}
& \gradcell{49.21} & \gradcell{58.98}
& \gradcell{51.58} & \gradcell{52.69} \\

\midrule

\multirow{2}{*}{Gemini 2.5 Flash}
& CheckEHR
& \gradcell{48.80} & \gradcell{65.14}
& \gradcell{28.05} & \gradcell{35.86}
& \gradcell{38.07} & \gradcell{64.48}
& \gradcell{44.89} & \gradcell{55.01}
& \gradcell{26.38} & \gradcell{30.61}
& \gradcell{35.16} & \gradcell{55.59} \\
& \frname
& \gradcellfmt{77.36}{\textbf{77.36}} & \gradcellfmt{71.67}{\textbf{71.67}}
& \gradcellfmt{79.58}{\textbf{79.58}} & \gradcellfmt{71.81}{\textbf{71.81}}
& \gradcellfmt{72.53}{\textbf{72.53}} & \gradcellfmt{71.59}{\textbf{71.59}}
& \gradcellfmt{69.28}{\textbf{69.28}} & \gradcellfmt{62.81}{\textbf{62.81}}
& \gradcellfmt{73.14}{\textbf{73.14}} & \gradcellfmt{66.11}{\textbf{66.11}}
& \gradcellfmt{65.26}{\textbf{65.26}} & \gradcellfmt{62.59}{\textbf{62.59}} \\

\bottomrule
\end{tabular}
}

\vspace{-2mm}
\label{tab:main_res}
\end{table*}

\footnotetext{Due to API and computational costs, our primary evaluation is based on a single inference pass.
To assess robustness, we conducted three independent inference runs on the validation set, each evaluated three times by the LLM-as-a-judge.
Additionally, we performed a paired bootstrap analysis with 10,000 resamples on the 83 held-out test notes to assess the statistical significance of performance differences against CheckEHR.
Detailed results are provided in Appendix~\ref{app:statistical_eval}.}

\subsection{Evaluation}


\textbf{Metrics}~~~
We evaluate the frameworks at the entity level. For each note, we compute Recall, Precision, and F1 and report the average scores. Recall is defined as the number of correctly classified entities divided by the number of ground-truth entities in the note. Precision is defined as the number of correctly classified entities divided by the number of entities recognized by the framework.


\textbf{LLM-as-a-judge}~~~
Direct rule-based comparison between framework outputs and gold annotations is unreliable due to differences in span boundaries and granularity. For example, in the phrase \textit{``lung sounds: clear, no crackles,''} a human annotator may treat \textit{``clear''} and \textit{``no crackles''} as separate verifications for the entity \textit{``lung sounds,''} whereas a framework may extract \textit{``lung''} as the entity and associate both values jointly or separately. Such discrepancies in span boundaries and granularity make rule-based evaluation challenging. To address this issue, we adopt an LLM-as-a-judge evaluator based on Gemini 2.5 Pro~\cite{comanici2025gemini}, which compares framework outputs and gold annotations semantically. 
In addition to assessing agreement with expert-guided gold annotations, we further consider the fact that equivalent clinical findings can be expressed in different ways and mapped to different structured EHR fields~\cite{cohen2019variation, newman2021ambiguity, rosenbloom2011data}.
For example, when a clinical note states \textit{“No rash,”} both the gold annotation and the framework output may extract \textit{“rash”} as the entity, yet map it to different table item–value pairs. The gold annotation may use the table item \textit{“Rash”} with the value \textit{“None,”} directly representing the absence of rash, whereas the framework may use the table item \textit{“Skin integrity”} with the value \textit{“intact,”} interpreting the same expression as a broader skin-status finding. Although these outputs differ structurally, both capture the same clinical information—that no rash was observed. 
To evaluate this, we define two complementary evaluation settings: \textit{Harsh} and \textit{Lenient}. The \textit{Harsh} evaluator enforces exact agreement with gold annotations, while the \textit{Lenient} evaluator accepts clinically plausible variations even when they diverge from gold annotations. To validate the reliability of the LLM-based judgments, a subset of evaluation samples was independently reviewed by human annotators. The \textit{Harsh} setting achieved 99.46\% agreement with author annotations, whereas the \textit{Lenient} setting, involving subjective clinical judgment, reached 95.35\% agreement among four practitioners. We also addressed potential evaluator bias via cross-model validation with GPT-5~\cite{singh2025openai}. Additional details are provided in Appendix~\ref{app:llmjudge}.

\subsection{Main Results}

Table \ref{tab:main_res} shows that \frname generally outperforms CheckEHR across a range of backbone LLMs under both evaluation settings, demonstrating its effectiveness across diverse model families. The performance gap becomes more pronounced when using models with strong reasoning capabilities, such as Qwen3 32B or Gemini 2.5 Flash, indicating that \frname effectively leverages advanced reasoning capabilities. Stronger LLMs also exhibit a larger gap between \textit{Lenient} and \textit{Harsh} evaluation, suggesting that they produce more independent verification rationales that differ from ground-truth annotations, often capturing clinically valid interpretations beyond the annotated labels. Interestingly, although MedGemma 27B is specialized for medical tasks, it does not outperform general-purpose reasoning models and even underperforms in the case of CheckEHR. This suggests that the task requires not only domain-specific medical knowledge but also strong structured reasoning over tabular evidence, where improvements in reasoning capability can play a critical role.
\begin{wrapfigure}{r}{0.40\textwidth}
    \centering
    \vspace{-3.5mm}
    \includegraphics[width=\linewidth]{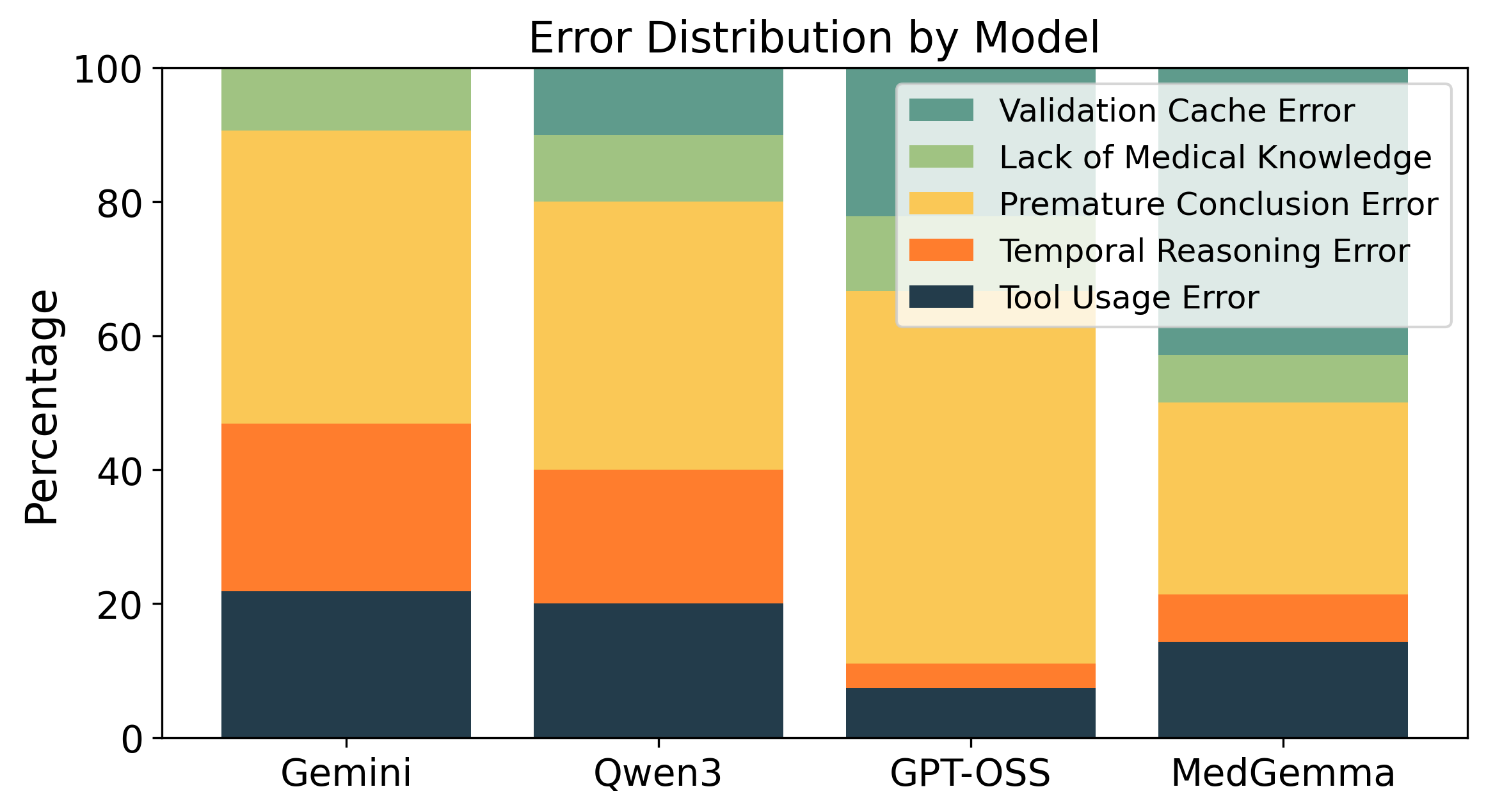}
    \vspace{-4.8mm}
    \caption{Error distribution by model in \frname.}
    \label{fig:errortype}
\end{wrapfigure}
\textbf{Error Analysis by Model}~~~To analyze error cases in \frname across models, we sampled 100 errors per model and categorized them into five types. As shown in Figure~\ref{fig:errortype}, premature conclusions were the most frequent across all models, indicating a systematic under-search issue where models terminate verification before collecting sufficient evidence. This highlights the need for deeper table exploration mechanisms. Additionally, stronger models rarely exhibit validation cache errors, implying more reliable tracking of intermediate verification states. A more detailed error analysis is provided in Appendix~\ref{app:error_analysis}.

\begin{table}[h]
\vspace{-2mm}
\centering
\begin{minipage}[t]{0.52\textwidth}
\centering
\caption{Ablation study on note segmentation and scope filtering. \textbf{Bold} indicates the highest performance.}
\vspace{2mm}
\label{tab:ablation}
\resizebox{\textwidth}{!}{
\begin{tabular}{lcccccc}
\toprule
\multirow{2}{*}{\textbf{Method}} 
& \multicolumn{3}{c}{\textbf{Lenient}} 
& \multicolumn{3}{c}{\textbf{Harsh}} \\
\cmidrule(lr){2-4} \cmidrule(lr){5-7}
 & Rec & Prec & F1 & Rec & Prec & F1 \\
\midrule
Ours & 77.36 & \textbf{72.31} & \textbf{74.75} & 67.84 & \textbf{61.46} & \textbf{64.49} \\
- Segmentation & 69.32 & 70.19 & 69.75 &  61.41 &  60.09 &  60.73 \\
- Filtering & \textbf{79.95} & 62.92 & 70.04 & \textbf{71.09} & 53.52 & 60.63 \\
\bottomrule
\end{tabular}%
}
\end{minipage}
\hfill 
\begin{minipage}[t]{0.46\textwidth}
\centering
\caption{Performance by entity extraction method. \textbf{Bold} and \underline{underline} denote the best and second-best results, respectively.}
\vspace{2mm}
\label{tab:comparison}
\resizebox{\textwidth}{!}{
\begin{tabular}{lcccccc}
\toprule
\multirow{2}{*}{\textbf{Method}} & \multicolumn{3}{c}{\textbf{Lenient}} & \multicolumn{3}{c}{\textbf{Harsh}} \\
\cmidrule(lr){2-4} \cmidrule(lr){5-7}
 & Rec & Prec & F1 & Rec & Prec & F1 \\
\midrule
Ours & \textbf{77.36} & \underline{72.31} & \textbf{74.75} & \textbf{67.84} & \underline{61.46} & \textbf{64.49} \\
Trained NER & 52.64 & \textbf{79.21} & 63.20 & 46.65 & \textbf{71.34} & 56.06 \\
BERT Ensemble & \underline{72.24} & 64.99 & \underline{68.35} & \underline{62.15} & 54.41 & \underline{58.02} \\
CheckEHR & 70.71 & 64.35 & 67.27 & 61.41 & 53.60 & 57.24 \\
\bottomrule
\end{tabular}%
}
\label{tab:ner_results}
\end{minipage}
\vspace{-2mm}
\end{table}

\subsection{Ablation Study}
To analyze the contribution of each component, we conduct an ablation study on the key modules of \frname using Gemini 2.5 Flash as the base LLM. All experiments and further analyses are performed on the validation set.

\textbf{Note Segmentation}~~~
To examine the effect of note segmentation, we remove the segmentation stage and perform entity extraction and verification on the entire clinical note. As shown in Table~\ref{tab:ablation}, the F1 score decreases in both settings. This suggests that segmentation improves reliability by structuring long clinical notes into manageable units. Without segmentation, the model must process longer contexts at once, which reduces extraction stability and leads to more verification errors.

\textbf{Entity Extraction}~~~
To analyze the impact of the entity extraction stage, we compare our method with three NER baselines while keeping the rest of the framework unchanged: Trained NER, BERT Ensemble, and CheckEHR NER. Trained NER fine-tunes MedGemma 27B on note–entity pairs from \dtname, BERT Ensemble aggregates predictions from three biomedical BERT-based models trained on different datasets~\cite{beltagy2019scibert, mattupalli2023clinicalner, uzuner20112010i2b2}, and CheckEHR NER follows the entity extraction method used in the original CheckEHR pipeline. Additional implementation details are provided in Appendix~\ref{app:ner_baselines}. As shown in Table~\ref{tab:ner_results}, our method achieves the highest F1 scores in both evaluation settings. The baselines exhibit a Precision–Recall trade-off: Trained NER achieves high Precision but relatively low Recall due to limited generalization to diverse entity expressions, whereas BERT Ensemble improves Recall by aggregating multiple models but often predicts additional entities not present in the ground-truth, reducing Precision.

\textbf{Scope Filtering}~~~
To examine the contribution of scope filtering, we conduct an ablation experiment by removing the filtering stage from the framework. As shown in Table~\ref{tab:ablation}, removing temporal filtering decreases F1 scores in both evaluation settings.
Without filtering, the framework extracts more entities, increasing Recall but introducing many irrelevant entities that reduce Precision. As a result, the overall F1 score declines. These results highlight the role of scope filtering in controlling excessive entity extraction and maintaining a balance between Recall and Precision.

\textbf{Tool Types}~~~
Similar to prior studies~\cite{qin2023toolllm, xu2025llm}, the toolset of \frname is designed by humans. We evaluate its effectiveness by ablating each tool category (\S\ref{sec3}). 
Interestingly, removing certain tools resulted in performance improvements, as shown in Table~\ref{tab:tool_ablation}.
This can be attributed to the different reasoning processes between human annotators and LLMs. Human annotators, with limited working memory, tend to use tools that search for table items or explore the database structure in order to narrow the search space before examining large tables. In contrast, LLMs, with much larger working memory (process 1M tokens), do not require such space-reduction strategies. 
For LLMs, tools designed to search for items or explore the database structure can introduce unnecessary complexity and lead to premature conclusions. These findings suggest the agents should dynamically manipulate toolsets to optimize performance. Detailed experimental results are provided in Appendix~\ref{app:tool_ablation}.

\begin{table}[t]
    \centering
    \begin{minipage}[t]{0.48\textwidth}
        \vspace{-2mm}
        \centering
        \caption{Tool category ablation results under Lenient evaluation metrics.}
        \label{tab:tool_ablation}
        \vspace{2mm}
        \resizebox{\linewidth}{!}{
        \begin{tabular}{lccc}
        \toprule
        \multirow{2}{*}{\textbf{Method}} 
        & \multicolumn{3}{c}{\textbf{Lenient}} \\
        \cmidrule(lr){2-4}
         & Rec & Prec & F1 \\
        \midrule
        \frname & 77.36 & 72.31 & 74.75  \\
        - Entity-to-Table Item Alignment & 79.00 & 73.23 & 76.00  \\
        - Database Exploration and Value Profiling & 77.41 & 71.70 & 74.45  \\
        - Temporal and Conditional Record Retrieval & 72.58 & 69.96 & 71.25  \\
        \bottomrule
        \end{tabular}}
    \end{minipage}
    \hfill
    \begin{minipage}[t]{0.51\textwidth}
        \vspace{0pt}
        \centering
        \includegraphics[width=\linewidth]{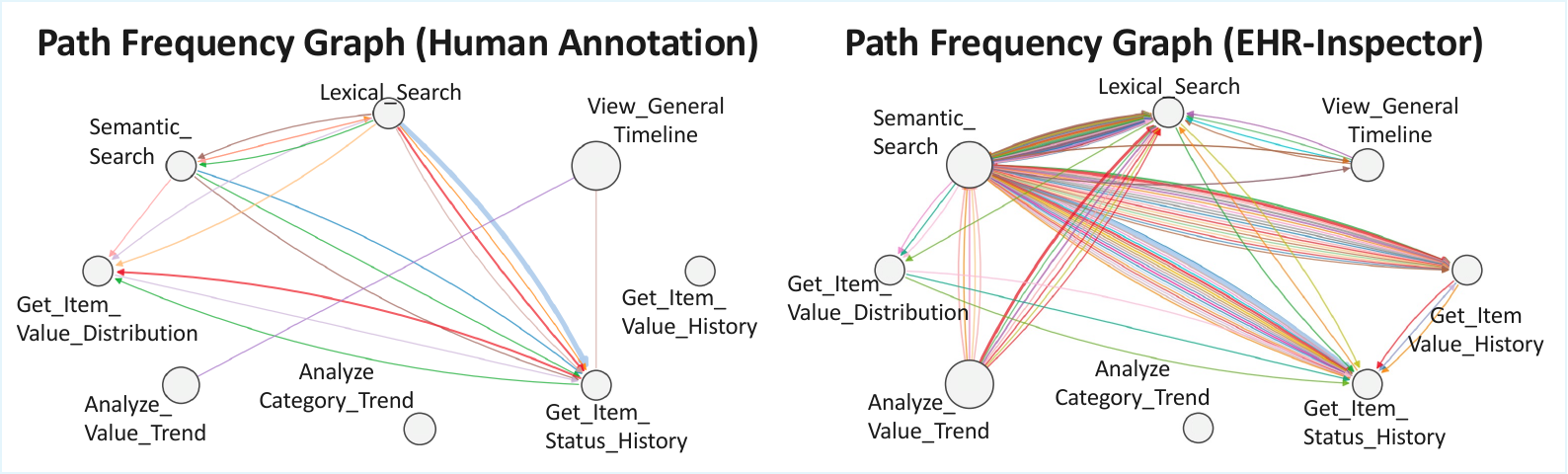}
        \vspace{-3.5mm}
        \captionof{figure}{Tool usage trace comparison between human annotation and \frname.}
        \label{fig:frameworkvshuman}
    \end{minipage}
    \vspace{-7.5mm}
\end{table}

\subsection{Further Analysis}


\textbf{Tool Usage: Annotators vs. \frname}~~~
We compare the tool usage paths of human annotators and the Gemini 2.5 Flash-based \frname. The results, plotted for frequently occurring tool usage traces during consistency verification, are shown in Figure~\ref{fig:frameworkvshuman}. Human annotators exhibit relatively limited and less varied tool usage traces, suggesting that they identify optimized tool usage traces and apply them conservatively to minimize risks~\cite{inbook}. In contrast, \frname exhibits a broader but less controlled exploration of tool usage paths, resulting in more complex and dispersed patterns. Therefore, to enhance the framework, it should transition from less controlled exploration to a data-driven approach that optimizes tool usage traces based on past outcomes. Detailed trace analysis by model is described in Appendix~\ref{app:tool_trace_model}.

\textbf{Generalization Across Datasets and Database Schemas}~~~
We evaluate \frname's generalization ability across both different datasets and database schemas. On EHRCon~\cite{kwon2024ehrcon}, a \textit{surface-level} consistency verification dataset, \frname achieves a Lenient F1 score of 86.41, outperforming \mbox{CheckEHR} by 24.99, indicating strong cross-dataset generalization. We further assess robustness to both different database schemas and note structures using MIMIC-IV discharge summaries~\cite{johnson2020mimic}. 
Despite increased complexity from additional tables (\ie \textit{emar} and \textit{emar\_detail}), Lenient F1 drops only modestly from 71.03\footnote{For MIMIC-IV, we annotated discharge summaries and compared model performance with the same setting in MIMIC-III.} to 68.67.
Additionally, to examine reliance on prior knowledge of the MIMIC database (see Appendix~\ref{app:prior_mimiciii}), we construct a perturbed schema with unseen table and column names. In this setting, Lenient F1 decreases from 74.44 to 69.73, suggesting some benefit from prior familiarity; nevertheless, \frname maintains reasonable performance, demonstrating robustness to both dataset shifts and schema variations. Detailed results are provided in Appendix~\ref{app:ehrcon_res}.

\section{Conclusion and Limitations}
\label{limitation}
In this study, we introduce \dtname, a reasoning-intensive benchmark to verify consistency between clinical notes and structured tables by addressing real-world EHR documentation challenges. To tackle this benchmark, we propose \frname, an LLM-based framework that mirrors the human annotation workflow through tool-augmented table reasoning. Experimental results show that \frname achieves state-of-the-art performance across diverse model backbones.

While our dataset curation and evaluation design were rigorous, certain limitations exist. The limited number of clinical notes used to build \dtname may constrain its representativeness across varied clinical scenarios. Furthermore, its reliance on preprocessed MIMIC data could fail to capture the true complexity of real-world clinical errors. Finally, the LLM-as-a-judge paradigm, though highly scalable and consistent with expert consensus, inherently risks introducing model-driven biases. To address these limitations, future work should focus on expanding data scale and diversity while incorporating scalable expert-in-the-loop evaluation frameworks to improve overall robustness.

\begin{ack}
This work was supported by the Institute of Information \& Communications Technology Planning \& Evaluation (IITP) grants (No.RS-2019-II190075, No.RS-2025-02304967), the Korea Health Industry Development Institute (KHIDI) grant (No.HR21C0198), the InnoCORE Program (N10250156), and the National Research Foundation of Korea (NRF) grant (NRF-2020H1D3A2A03100945), and the Medical Scientist Training Program funded by the Korean government (MSIT, MOHW).
\end{ack}

\bibliographystyle{plain}
\bibliography{custom}

@article{singh2025openai,
  title={Openai gpt-5 system card},
  author={Singh, Aaditya and Fry, Adam and Perelman, Adam and Tart, Adam and Ganesh, Adi and El-Kishky, Ahmed and McLaughlin, Aidan and Low, Aiden and Ostrow, AJ and Ananthram, Akhila and others},
  journal={arXiv preprint arXiv:2601.03267},
  year={2025}
}

@article{qin2023toolllm,
  title={Toolllm: Facilitating large language models to master 16000+ real-world apis},
  author={Qin, Yujia and Liang, Shihao and Ye, Yining and Zhu, Kunlun and Yan, Lan and Lu, Yaxi and Lin, Yankai and Cong, Xin and Tang, Xiangru and Qian, Bill and others},
  journal={arXiv preprint arXiv:2307.16789},
  year={2023}
}

@article{xu2025llm,
  title={LLM-based agents for tool learning: A survey},
  author={Xu, Weikai and Huang, Chengrui and Gao, Shen and Shang, Shuo},
  journal={Data Science and Engineering},
  volume={10},
  pages={533--563},
  year={2025},
  publisher={Springer},
  doi={10.1007/s41019-025-00296-9}
}

@inproceedings{beltagy2019scibert,
  title     = {SciBERT: A Pretrained Language Model for Scientific Text},
  author    = {Beltagy, Iz and Lo, Kyle and Cohan, Arman},
  booktitle = {Proceedings of EMNLP-IJCNLP},
  year      = {2019},
  pages     = {3615--3620}
}

@article{uzuner20112010i2b2,
  title   = {2010 i2b2/VA challenge on concepts, assertions, and relations in clinical text},
  author  = {Uzuner, Ozlem and South, Brett R. and Shen, Shuying and DuVall, Scott L.},
  journal = {Journal of the American Medical Informatics Association},
  volume  = {18},
  number  = {5},
  pages   = {552--556},
  year    = {2011}
}

@misc{mattupalli2023clinicalner,
  title        = {Clinical NER Model},
  author       = {Mattupalli, Saketh},
  year         = {2023},
  howpublished = {\url{https://huggingface.co/blaze999/clinical-ner}},
  note         = {HuggingFace model for clinical named entity recognition, fine-tuned from DeBERTa-v3-base}
}

@article{sellergren2025medgemma,
  title={Medgemma technical report},
  author={Sellergren, Andrew and Kazemzadeh, Sahar and Jaroensri, Tiam and Kiraly, Atilla and Traverse, Madeleine and Kohlberger, Timo and Xu, Shawn and Jamil, Fayaz and Hughes, C{\'\i}an and Lau, Charles and others},
  journal={arXiv preprint arXiv:2507.05201},
  year={2025}
}

@article{agarwal2025gpt,
  title={gpt-oss-120b \& gpt-oss-20b model card},
  author={Agarwal, Sandhini and Ahmad, Lama and Ai, Jason and Altman, Sam and Applebaum, Andy and Arbus, Edwin and Arora, Rahul K and Bai, Yu and Baker, Bowen and Bao, Haiming and others},
  journal={arXiv preprint arXiv:2508.10925},
  year={2025}
}

@article{yang2025qwen3,
  title={Qwen3 technical report},
  author={Yang, An and Li, Anfeng and Yang, Baosong and Zhang, Beichen and Hui, Binyuan and Zheng, Bo and Yu, Bowen and Gao, Chang and Huang, Chengen and Lv, Chenxu and others},
  journal={arXiv preprint arXiv:2505.09388},
  year={2025}
}

@article{comanici2025gemini,
  title={Gemini 2.5: Pushing the frontier with advanced reasoning, multimodality, long context, and next generation agentic capabilities},
  author={Comanici, Gheorghe and Bieber, Eric and Schaekermann, Mike and Pasupat, Ice and Sachdeva, Noveen and Dhillon, Inderjit and Blistein, Marcel and Ram, Ori and Zhang, Dan and Rosen, Evan and others},
  journal={arXiv preprint arXiv:2507.06261},
  year={2025}
}

@article{wang2025sheetbrain,
  title={SheetBrain: A Neuro-Symbolic Agent for Accurate Reasoning over Complex and Large Spreadsheets},
  author={Wang, Ziwei and Su, Jiayuan and Zhou, Mengyu and Zeng, Huaxing and Jia, Mengni and Lv, Xiao and Dong, Haoyu and Ma, Xiaojun and Han, Shi and Zhang, Dongmei},
  journal={arXiv preprint arXiv:2510.19247},
  year={2025}
}

@article{zhou2025mixture,
  title={Mixture-of-minds: Multi-agent reinforcement learning for table understanding},
  author={Zhou, Yuhang and Zhang, Mingrui and Li, Ke and Wang, Mingyi and Liu, Qiao and Wang, Qifei and Liu, Jiayi and Liu, Fei and Li, Serena and Li, Weiwei and others},
  journal={arXiv preprint arXiv:2510.20176},
  year={2025}
}

@article{xiong2025tablezoomer,
  title={TableZoomer: a collaborative agent framework for large-scale table question answering},
  author={Xiong, Sishi and He, Ziyang and He, Zhongjiang and Zhao, Yu and Pan, Changzai and Zhang, Jie and Song, Shuangyong and Li, Yongxiang},
  journal={Vicinagearth},
  volume={2},
  number={1},
  pages={1--23},
  year={2025},
  publisher={Springer}
}

@article{jiang2025tablemind,
  title={Tablemind: An autonomous programmatic agent for tool-augmented table reasoning},
  author={Jiang, Chuang and Cheng, Mingyue and Tao, Xiaoyu and Mao, Qingyang and Ouyang, Jie and Liu, Qi},
  journal={arXiv preprint arXiv:2509.06278},
  year={2025}
}

@inproceedings{lu-etal-2025-tart,
    title = "{TART}: An Open-Source Tool-Augmented Framework for Explainable Table-based Reasoning",
    author = "Lu, Xinyuan  and
      Pan, Liangming  and
      Ma, Yubo  and
      Nakov, Preslav  and
      Kan, Min-Yen",
    editor = "Chiruzzo, Luis  and
      Ritter, Alan  and
      Wang, Lu",
    booktitle = "Findings of the Association for Computational Linguistics: NAACL 2025",
    month = apr,
    year = "2025",
    address = "Albuquerque, New Mexico",
    publisher = "Association for Computational Linguistics",
    url = "https://aclanthology.org/2025.findings-naacl.244/",
    doi = "10.18653/v1/2025.findings-naacl.244",
    pages = "4323--4339",
    ISBN = "979-8-89176-195-7"
}

@article{PhysioNet-mimiciii-1.4,
  author = {Johnson, Alistair and Pollard, Tom and Mark, Roger},
  title = {{MIMIC-III Clinical Database}},
  journal = {{PhysioNet}},
  year = {2016},
  month = sep,
  note = {Version 1.4},
  doi = {10.13026/C2XW26},
  url = {https://doi.org/10.13026/C2XW26}
}

@article{Li2015EndToEndMedDiscrepancy,
  author       = {Li, Qi and Spooner, Stephen Andrew and Kaiser, Megan and Lingren, Nataline and Robbins, Jessica and Lingren, Todd and Tang, Huaxiu and Solti, Imre and Ni, Yizhao},
  title        = {An end-to-end hybrid algorithm for automated medication discrepancy detection},
  journal      = {BMC Medical Informatics and Decision Making},
  year         = {2015},
  volume       = {15},
  number       = {1},
  pages        = {37},
  date         = {2015-05-06},
  doi          = {10.1186/s12911-015-0160-8},
  url          = {https://pmc.ncbi.nlm.nih.gov/articles/PMC4427951/},
  pmid         = {25943550},
  pmcid        = {PMC4427951}
}

@inproceedings{Rinott2012Sarcoma,
  author    = {Ruty Rinott and Michele Torresani and Rossella Bertulli and Abigail Goldsteen and Paolo Casali and Boaz Carmeli and Noam Slonim},
  title     = {Automatic Detection of Inconsistencies between Free Text and Coded Data in Sarcoma Discharge Letters},
  booktitle = {Studies in Health Technology and Informatics},
  volume    = {180},
  pages     = {661--666},
  year      = {2012},
  doi       = {10.3233/978-1-61499-101-4-661}
}

@article{Lo2022Reconciling,
  title        = {Reconciling Allergy Information in the Electronic Health Record After a Drug Challenge Using Natural Language Processing},
  author       = {Lo, Ying-Chih and Varghese, Sheril and Blackley, Suzanne and Seger, Diane L. and Blumenthal, Kimberly G. and Goss, Foster R. and Zhou, Li},
  journal      = {Frontiers in Allergy},
  year         = {2022},
  volume       = {3},
  pages        = {904923},
  doi          = {10.3389/falgy.2022.904923},
  pmid         = {35769562},
  pmcid        = {PMC9234873}
}

@inproceedings{Raghavan2014Unstructured,
  title     = {How essential are unstructured clinical narratives and information fusion to clinical trial recruitment?},
  author    = {Raghavan, Preethi and Chen, James L. and Fosler-Lussier, Eric and Lai, Albert M.},
  booktitle = {AMIA Joint Summits on Translational Science Proceedings},
  volume    = {2014},
  pages     = {218--223},
  year      = {2014},
  month     = {Apr},
  pmid      = {25717416},
  pmcid     = {PMC4333685}
}

@inproceedings{Yu2024Temporal,
  title     = {A Systematic Temporal Extraction Pipeline for Medical Concepts in Clinical Notes},
  author    = {Yu, Deahan and Stidham, Ryan W. and Vydiswaran, V. G. Vinod},
  booktitle = {AMIA Annual Symposium Proceedings},
  volume    = {2023},
  pages     = {1314--1323},
  year      = {2024},
  month     = {Jan},
  pmid      = {38222360},
  pmcid     = {PMC10785919}
}

@inproceedings{khetan2022mimicause,
  title={MIMICause: Representation and automatic extraction of causal relation types from clinical notes},
  author={Khetan, Vivek and Rizvi, Md Imbesat and Huber, Jessica and Bartusiak, Paige and Sacaleanu, Bogdan and Fano, Andrew},
  booktitle={Findings of the association for computational linguistics: ACL 2022},
  pages={764--773},
  year={2022}
}

@article{Gao2024,
  title   = {Clinical natural language processing for secondary uses},
  author  = {Gao, Yanjun and Mahajan, Diwakar and Uzuner, {\"O}zlem and Yetisgen, Meliha},
  journal = {Journal of Biomedical Informatics},
  volume  = {150},
  pages   = {104596},
  year    = {2024},
  month   = {Feb},
  doi     = {10.1016/j.jbi.2024.104596},
  pmid    = {38278312},
  pmcid   = {PMC11212507},
  publisher = {Elsevier}
}

@Article{pan2020,
author="Pan, Xiaoyi
and Chen, Boyu
and Weng, Heng
and Gong, Yongyi
and Qu, Yingying",
title="Temporal Expression Classification and Normalization From Chinese Narrative Clinical Texts: Pattern Learning Approach",
journal="JMIR Med Inform",
year="2020",
month="Jul",
day="27",
volume="8",
number="7",
pages="e17652",
issn="2291-9694",
doi="10.2196/17652",
url="https://medinform.jmir.org/2020/7/e17652",
url="https://doi.org/10.2196/17652",
url="http://www.ncbi.nlm.nih.gov/pubmed/32716307"
}

@article{WANG201834,
title = {Clinical information extraction applications: A literature review},
journal = {Journal of Biomedical Informatics},
volume = {77},
pages = {34-49},
year = {2018},
issn = {1532-0464},
doi = {https://doi.org/10.1016/j.jbi.2017.11.011},
url = {https://www.sciencedirect.com/science/article/pii/S1532046417302563},
author = {Yanshan Wang and Liwei Wang and Majid Rastegar-Mojarad and Sungrim Moon and Feichen Shen and Naveed Afzal and Sijia Liu and Yuqun Zeng and Saeed Mehrabi and Sunghwan Sohn and Hongfang Liu}
}

@INPROCEEDINGS{7001812,
  author={Villa, Luis Bernardo and Cabezas, Ivan},
  booktitle={2014 IEEE 16th International Conference on e-Health Networking, Applications and Services (Healthcom)}, 
  title={A review on usability features for designing electronic health records}, 
  year={2014},
  volume={},
  number={},
  pages={49-54},
  doi={10.1109/HealthCom.2014.7001812}}

@article{payne2018using, title={Using voice to create inpatient progress notes: effects on note timeliness, quality, and physician satisfaction}, author={Payne, Thomas H and Alonso, W David and Markiel, J Andrew and Lybarger, Kevin and Lordon, Ross and Yetisgen, Meliha and Zech, Jennifer M and White, Andrew A}, journal={JAMIA open}, volume={1}, number={2}, pages={218--226}, year={2018}, publisher={Oxford University Press} }

@article{tsou2017copypaste,
  title={Safe Practices for Copy and Paste in the EHR: Systematic Review, Recommendations, and Novel Model for Health IT Collaboration},
  author={Tsou, Amy Y and Lehmann, Christoph U and Michel, Jeremy and Solomon, Ronni and Possanza, Lorraine and Gandhi, Tejal},
  journal={Applied Clinical Informatics},
  volume={8},
  number={1},
  pages={12--34},
  year={2017},
  doi={10.4338/ACI-2016-09-R-0150}
}

@article{demsash2023documentation,
  title={Health professionals' routine practice documentation and its associated factors in a resource-limited setting: a cross-sectional study},
  author={Demsash, Addisalem Workie and Kassie, Sisay Yitayih and others},
  journal={BMJ Health \& Care Informatics},
  volume={30},
  number={1},
  pages={e100699},
  year={2023},
  doi={10.1136/bmjhci-2022-100699}
}

@inproceedings{kwon2024ehrcon,
  title={EHRCon: Dataset for Checking Consistency between Unstructured Notes and Structured Tables in Electronic Health Records},
  author={Kwon, Yeonsu and Kim, Jiho and Lee, Gyubok and Bae, Seongsu and Kyung, Daeun and Cha, Wonchul and Pollard, Tom and Johnson, Alistair and Choi, Edward},
  booktitle={Advances in Neural Information Processing Systems (NeurIPS) Datasets and Benchmarks Track},
  year={2024}
}

@article{Seinen2024UsingSC,
  title={Using Structured Codes and Free-Text Notes to Measure Information Complementarity in Electronic Health Records: Feasibility and Validation Study},
  author={Tom M. Seinen and Jan A. Kors and Erik M. van Mulligen and Peter R. Rijnbeek},
  journal={Journal of Medical Internet Research},
  year={2024},
  volume={27},
  url={https://api.semanticscholar.org/CorpusID:274208819}
}

@article{johnson2020mimic,
  title={Mimic-iv},
  author={Johnson, Alistair and Bulgarelli, Lucas and Pollard, Tom and Horng, Steven and Celi, Leo Anthony and Mark, Roger},
  journal={PhysioNet. Available online at: https://physionet. org/content/mimiciv/1.0/(accessed August 23, 2021)},
  pages={49--55},
  year={2020}
}

@inbook{inbook,
author = {Machina, Mark and Siniscalchi, Marciano},
year = {2014},
month = {12},
pages = {},
title = {Ambiguity and Ambiguity Aversion},
volume = {1},
isbn = {9780444536853},
journal = {Handbook of the Economics of Risk and Uncertainty},
doi = {10.1016/B978-0-444-53685-3.00013-1}
}

@inproceedings{hsu2020characterizing,
  title={Characterizing the value of information in medical notes},
  author={Hsu, Chao-Chun and Karnwal, Shantanu and Mullainathan, Sendhil and Obermeyer, Ziad and Tan, Chenhao},
  booktitle={Findings of the Association for Computational Linguistics: EMNLP 2020},
  pages={2062--2072},
  year={2020}
}

@article{styler2014temporal,
  title={Temporal annotation in the clinical domain},
  author={Styler IV, William F and Bethard, Steven and Finan, Sean and Palmer, Martha and Pradhan, Sameer and De Groen, Piet C and Erickson, Brad and Miller, Timothy and Lin, Chen and Savova, Guergana and others},
  journal={Transactions of the association for computational linguistics},
  volume={2},
  pages={143--154},
  year={2014}
}

@article{zhang2018text,
  title={A text structuring method for chinese medical text based on temporal information},
  author={Zhang, Runtong and Chu, Fuzhi and Chen, Donghua and Shang, Xiaopu},
  journal={International journal of environmental research and public health},
  volume={15},
  number={3},
  pages={402},
  year={2018},
  publisher={MDPI}
}

@article{mathiesen2024effect,
  title={Effect of integrated medicines management on quality of discharge medication information—a secondary endpoint in a randomized controlled trial},
  author={Mathiesen, Liv and Nguyen, Tram Bich Michelle and D{\ae}hlen, Ingrid and Mow{\'e}, Morten and Lea, Marianne},
  journal={International Journal for Quality in Health Care},
  volume={36},
  number={4},
  pages={mzae100},
  year={2024},
  publisher={Oxford University Press UK}
}

@article{patel2019communication,
  title={Communication at transitions of care},
  author={Patel, Shilpa J and Landrigan, Christopher P},
  journal={Pediatric Clinics},
  volume={66},
  number={4},
  pages={751--773},
  year={2019},
  publisher={Elsevier}
}

@article{rajkomar2022deciphering,
  title={Deciphering clinical abbreviations with a privacy protecting machine learning system},
  author={Rajkomar, Alvin and Loreaux, Eric and Liu, Yuchen and Kemp, Jonas and Li, Benny and Chen, Ming-Jun and Zhang, Yi and Mohiuddin, Afroz and Gottweis, Juraj},
  journal={Nature communications},
  volume={13},
  number={1},
  pages={7456},
  year={2022},
  publisher={Nature Publishing Group UK London}
}

@inproceedings{wolf-etal-2020-transformers,
    title = "Transformers: State-of-the-Art Natural Language Processing",
    author = "Wolf, Thomas  and
      Debut, Lysandre  and
      Sanh, Victor  and
      Chaumond, Julien  and
      Delangue, Clement  and
      Moi, Anthony  and
      Cistac, Pierric  and
      Rault, Tim  and
      Louf, Remi  and
      Funtowicz, Morgan  and
      Davison, Joe  and
      Shleifer, Sam  and
      von Platen, Patrick  and
      Ma, Clara  and
      Jernite, Yacine  and
      Plu, Julien  and
      Xu, Canwen  and
      Le Scao, Teven  and
      Gugger, Sylvain  and
      Drame, Mariama  and
      Lhoest, Quentin  and
      Rush, Alexander",
    editor = "Liu, Qun  and
      Schlangen, David",
    booktitle = "Proceedings of the 2020 Conference on Empirical Methods in Natural Language Processing: System Demonstrations",
    month = oct,
    year = "2020",
    address = "Online",
    publisher = "Association for Computational Linguistics",
    url = "https://aclanthology.org/2020.emnlp-demos.6/",
    doi = "10.18653/v1/2020.emnlp-demos.6",
    pages = "38--45"
}

@article{rosenbloom2011data,
  title={Data from clinical notes: a perspective on the tension between structure and flexible documentation},
  author={Rosenbloom, S Trent and Denny, Joshua C and Xu, Hua and Lorenzi, Nancy and Stead, William W and Johnson, Kevin B},
  journal={Journal of the American Medical Informatics Association},
  volume={18},
  number={2},
  pages={181--186},
  year={2011},
  publisher={BMJ Group BMA House, Tavistock Square, London, WC1H 9JR}
}

@article{cohen2019variation,
  title={Variation in physicians’ electronic health record documentation and potential patient harm from that variation},
  author={Cohen, Genna R and Friedman, Charles P and Ryan, Andrew M and Richardson, Caroline R and Adler-Milstein, Julia},
  journal={Journal of general internal medicine},
  volume={34},
  number={11},
  pages={2355--2367},
  year={2019},
  publisher={Springer}
}

@article{newman2021ambiguity,
  title={Ambiguity in medical concept normalization: An analysis of types and coverage in electronic health record datasets},
  author={Newman-Griffis, Denis and Divita, Guy and Desmet, Bart and Zirikly, Ayah and Ros{\'e}, Carolyn P and Fosler-Lussier, Eric},
  journal={Journal of the American Medical Informatics Association},
  volume={28},
  number={3},
  pages={516--532},
  year={2021},
  publisher={Oxford University Press}
}

\newpage
\appendix
\newpage
\startcontents[supplementary]
\printcontents[supplementary]{l}{1}{\section*{Supplementary Contents}}

\newpage
\section{Dataset Details}
\label{app:dataset_details}
\dtname is a reasoning-intensive consistency verification dataset built based on the MIMIC-III database. In this section, we will provide detailed information about the tables and columns we handle with this dataset.

\subsection{Tables and Columns in \dtname}
\dtname performs consistency checks between three types of clinical notes—Discharge summaries, Physician notes, and Nursing notes—and 14 tables from MIMIC-III. The tables used for this process are Chartevents, Labevents, Prescriptions, Inputevents\_cv, Inputevents\_mv, Outputevents, Procedureevents\_mv, Microbiologyevents, Diagnoses\_icd, Procedures\_icd, D\_items, D\_icd\_diagnoses, D\_icd\_procedures, and D\_labitems. To ensure a comprehensive verification process, \dtname does not impose any restrictions on the scale or types of columns when verifying consistency between the notes and the tables. The final labeling results in a dataset that includes specific column entries from each of the tables, as detailed in Table~\ref{apptab:table_columns}.

\subsection{Three Types of Notes in \dtname}
We construct \dtname using 105 clinical notes, following the setting of EHRCon~\cite{kwon2024ehrcon}. To closely reflect real-world clinical practice, we selected three widely used types of inpatient clinical documentation: discharge summaries, physician notes, and nursing notes \cite{hsu2020characterizing}, capturing the characteristics of diverse clinical settings. In addition, we performed data filtering to improve analytical accuracy. The original notes may include information outside the scope of this study, such as family history or prior visit records. Therefore, these elements were excluded, and the dataset was refined to focus only on core information directly related to the current hospitalization process.

\subsection{Comparison with Surface-Level Consistency Verification Benchmark}
Going beyond a surface-level approach, \dtname is designed to faithfully reflect the complex context inherent in actual EHR documentation by integrating three key characteristics: (1) Compositional, (2) Time-series, and (3) Interpretive Information. To emphasize the strengths of this approach, we compare it with EHRCon, the only study that clearly defines both the task and a benchmark for verifying consistency between clinical notes and tables. In the following sections, we will explain how \dtname captures these characteristics in a more reasoning-intensive manner.

\paragraph{Compositional Clinical Information}
\dtname goes beyond simple numerical values or discrete event comparisons by incorporating a broader range of complex attributes. Existing benchmarks like EHRCon have primarily focused on fragmented data such as time (CHARTTIME) or numerical values (VALUENUM). In contrast, \dtname expands the validation scope to 14 tables and 58 columns (see Table~\ref{apptab:table_columns}). This includes complex attributes that were previously overlooked, such as event endtime (ENDTIME), procedure location (LOCATION), route of administration (ROUTE), and qualitative interpretations (INTERPRETATION) stored in the microbiologyevents table, which capture susceptibility results for antibiotic names (AB\_NAME). As a result, \dtname captures not just simple matches but the deeper consistency across the entire clinical documentation.

\paragraph{Temporally Evolving Information}
EHR data is inherently time-series data, and time information is a key factor in understanding a patient's clinical trajectory. EHRCon recognized this importance and utilized time information to validate the consistency between notes and table data. However, it simplified the time aspect into three categories: standard format, narrative expression (\eg admission date), and missing time information.
While this approach may be useful for verifying data consistency, it falls short of capturing the complex temporal context embedded in clinical records. Medical records often contain ambiguous time references such as ``post-surgery'' or ``early morning,'' as well as more difficult-to-define time spans or intervals, making standardization challenging.
In response, \dtname has adopted a reasoning-intensive approach, moving away from rigid classification systems. By analyzing the overall context of the document, it captures the subtle timing and sequential relationships between individual clinical events, accurately and comprehensively reconstructing fragmented information into a patient's complete time-series record.


\begin{table}[H]
\centering
\renewcommand{\arraystretch}{1.2} 
\caption{Tables and columns used in \dtname.}
\resizebox{0.5\columnwidth}{!}{%
\begin{tabular}{l|l}
\hline
\textbf{TABLE} & \textbf{COLUMN} \\ \hline
\multirow{3}{*}{D\_ITEMS} & ITEMID \\ \cline{2-2} 
 & CATEGORY \\ \cline{2-2} 
 & LABEL \\ \hline
DIAGNOSES\_ICD & ICD9\_CODE \\ \hline
\multirow{2}{*}{D\_ICD\_DIAGNOSES} & LONG\_TITLE \\ \cline{2-2} 
 & SHORT\_TITLE \\ \hline
\multirow{2}{*}{D\_ICD\_PROCEDURES} & LONG\_TITLE \\ \cline{2-2} 
 & SHORT\_TITLE \\ \hline
\multirow{3}{*}{D\_LABITEMS} & FLUID \\ \cline{2-2} 
 & ITEMID \\ \cline{2-2} 
 & LABEL \\ \hline
\multirow{9}{*}{INPUTEVENTS\_MV} & ORDERCATEGORYNAME \\ \cline{2-2} 
 & RATE \\ \cline{2-2} 
 & AMOUNT \\ \cline{2-2} 
 & STARTTIME \\ \cline{2-2} 
 & ENDTIME \\ \cline{2-2} 
 & RATEUOM \\ \cline{2-2} 
 & ORIGINALAMOUNT \\ \cline{2-2} 
 & ORIGINALRATE \\ \cline{2-2} 
 & AMOUNTUOM \\ \hline
\multirow{4}{*}{CHARTEVENTS} & CHARTTIME \\ \cline{2-2} 
 & VALUE \\ \cline{2-2} 
 & VALUENUM \\ \cline{2-2} 
 & VALUEUOM \\ \hline
\multirow{7}{*}{MICROBIOLOGYEVENTS} & CHARTTIME \\ \cline{2-2} 
 & SPEC\_ITEMID \\ \cline{2-2} 
 & SPEC\_TYPE\_DESC \\ \cline{2-2} 
 & ORG\_ITEMID \\ \cline{2-2} 
 & ORG\_NAME \\ \cline{2-2} 
 & INTERPRETATION \\ \cline{2-2} 
 & AB\_NAME \\ \hline
\multirow{6}{*}{INPUTEVENTS\_CV} & RATE \\ \cline{2-2} 
 & AMOUNT \\ \cline{2-2} 
 & CHARTTIME \\ \cline{2-2} 
 & RATEUOM \\ \cline{2-2} 
 & ORIGINALROUTE \\ \cline{2-2} 
 & AMOUNTUOM \\ \hline
\multirow{3}{*}{OUTPUTEVENTS} & CHARTTIME \\ \cline{2-2} 
 & VALUE \\ \cline{2-2} 
 & VALUEUOM \\ \hline
\multirow{5}{*}{LABEVENTS} & CHARTTIME \\ \cline{2-2} 
 & VALUE \\ \cline{2-2} 
 & VALUENUM \\ \cline{2-2} 
 & FLAG \\ \cline{2-2} 
 & VALUEUOM \\ \hline
PROCEDURES\_ICD & ICD9\_CODE \\ \hline
\multirow{9}{*}{PRESCRIPTIONS} & STARTDATE \\ \cline{2-2} 
 & ENDDATE \\ \cline{2-2} 
 & DRUG \\ \cline{2-2} 
 & PROD\_STRENGTH \\ \cline{2-2} 
 & DOSE\_VAL\_RX \\ \cline{2-2} 
 & DOSE\_UNIT\_RX \\ \cline{2-2} 
 & FORM\_VAL\_DISP \\ \cline{2-2} 
 & FORM\_UNIT\_DISP \\ \cline{2-2} 
 & ROUTE \\ \hline
\multirow{3}{*}{PROCEDUREEVENTS\_MV} & STARTTIME \\ \cline{2-2} 
 & ENDTIME \\ \cline{2-2} 
 & LOCATION \\ \hline
\end{tabular}%
}
\vspace{3mm}
\label{apptab:table_columns}
\end{table}

\paragraph{Clinical Interpreted Information}
Clinical notes and tables capture the same patient data in two distinct formats: unstructured narrative and standardized tabular forms. While previous benchmarks like EHRCon addressed these differences through lexical mappings (\eg abbreviations and brand names), actual clinical reasoning demands a deeper interpretive bridge. For example, confirming a diagnosis of `anemia' in the notes requires the knowledge to interpret the `hemoglobin (Hgb)' levels recorded in the tables. To bridge this gap, this study integrates authoritative clinical references with expert insights. By incorporating this interpretive layer, it enables a more precise and consistent review that aligns with real-world clinical practices. The clinical references used in this process are summarized in Table~\ref{tab:clinical_ref}.

\subsubsection{Statistics}
Table~\ref{tab:ehrcon2-stats} presents the key statistics of \dtname. The dataset consists of 105 clinical notes, split into 83 notes for the test set and 22 notes for the validation set, including 38 discharge summaries, 33 physician notes, and 34 nursing notes, with a total of 8,042 annotated entities from MIMIC-III. On average, each note contains 76.59 entities and 1,953 tokens. The annotations include 5,847 consistent and 2,195 inconsistent labels, providing a reliable ground truth for consistency verification. Among the three note types, discharge summaries are the most complex, with the longest average length of 2,789 tokens and the largest number of inconsistent labels (1,456). This reflects the difficulty of maintaining factual consistency in longitudinal clinical narratives.

\subsubsection{Fine-grained Analysis of Note–Table Discrepancies}
To better understand the inconsistencies identified in \dtname, we analyze their patterns across clinical notes and structured tables.

\begin{figure*}[t]
\centering

\begin{minipage}[t]{0.48\textwidth}
    \centering
    \includegraphics[width=\linewidth,height=5cm,keepaspectratio]{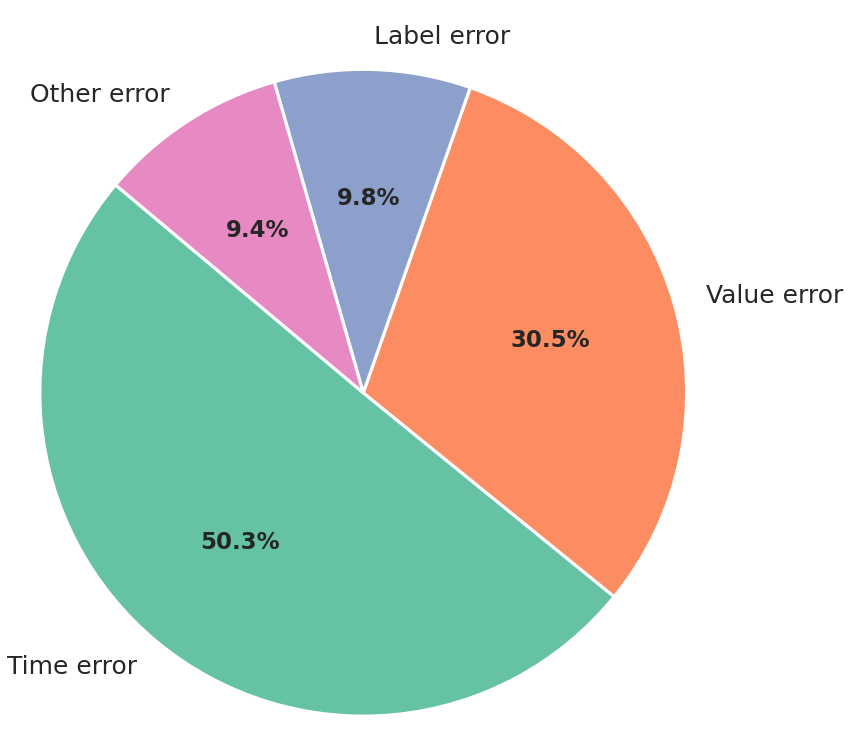}
    \captionof{figure}{Distribution of column-level inconsistency types, with time errors being the most prevalent.}
    \label{app_fig:incon_col_cate}
\end{minipage}
\hfill
\begin{minipage}[t]{0.48\textwidth}
    \centering
    \includegraphics[width=\linewidth,height=5cm,keepaspectratio]{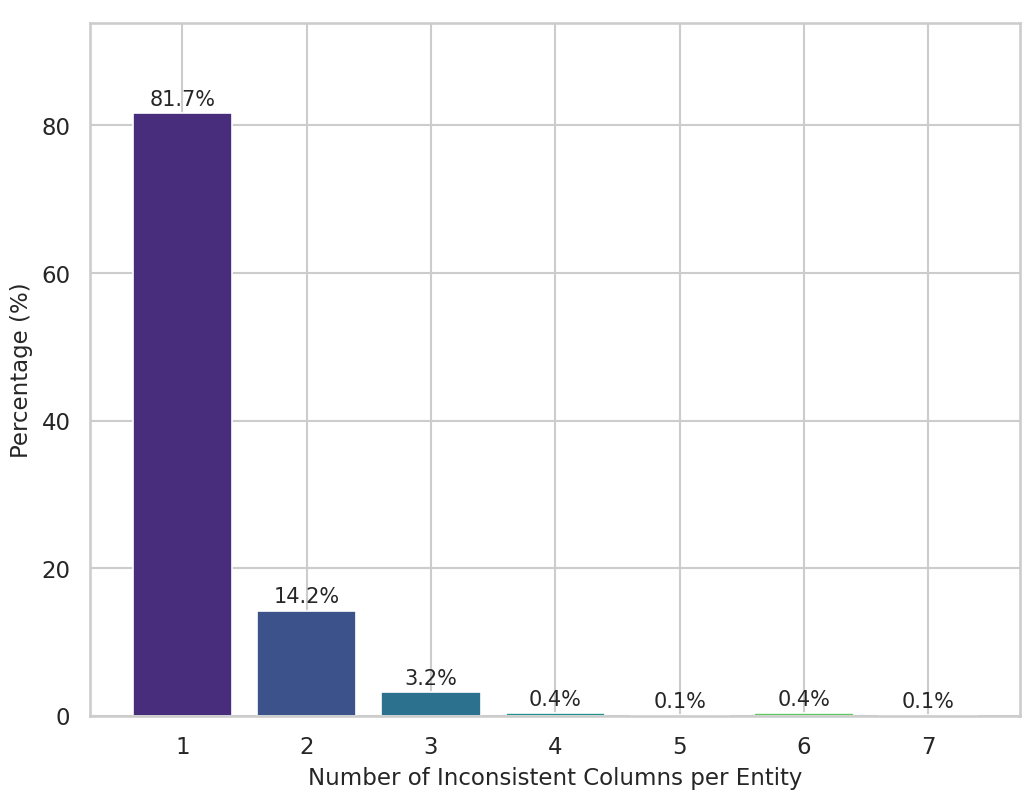}
    \captionof{figure}{Distribution of the number of inconsistent columns per entity, showing most inconsistencies occur in a single column.}
    \label{app_fig:number_incon}
\end{minipage}

\vspace{2mm}

\begin{minipage}[t]{0.48\textwidth}
    \centering
    \includegraphics[width=\linewidth,height=5cm,keepaspectratio]{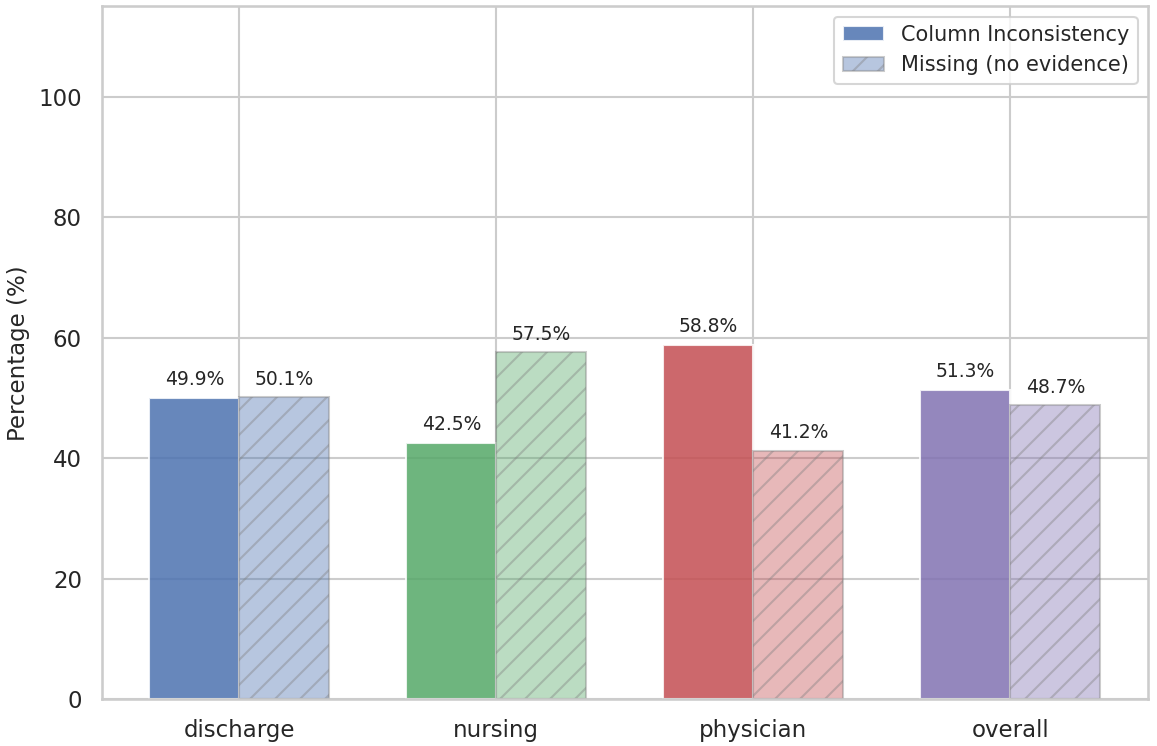}
    \captionof{figure}{Comparison of column inconsistency and missing evidence across note types, illustrating different patterns of discrepancy depending on the note category.}
    \label{app_fig:incon_om}
\end{minipage}
\hfill
\begin{minipage}[t]{0.48\textwidth}
    \centering
    \includegraphics[width=\linewidth,height=5cm,keepaspectratio]{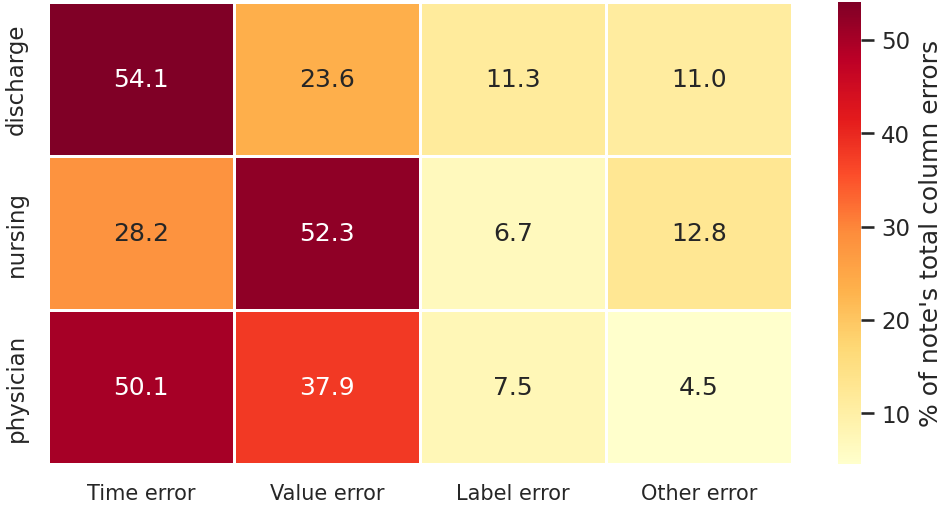}
    \captionof{figure}{Distribution of inconsistency types by note type, highlighting time errors in discharge and physician notes and value errors in nursing notes.}
    \label{app_fig:incon_per_note}
\end{minipage}

\end{figure*}

\paragraph{Column-level Inconsistency Patterns}~
As shown in Figure~\ref{app_fig:incon_col_cate}, the most prevalent type of column-level inconsistency is time error, accounting for 50.3\% of all inconsistent cases. This indicates that discrepancies between clinical notes and structured tables frequently arise from mismatches in the timing or duration of clinical events. Temporal information plays a critical role in interpreting the sequence of examinations, medications, procedures, and patient status changes~\cite{styler2014temporal}. Therefore, inconsistencies in the time column can hinder the reconstruction of a patient’s clinical trajectory and the determination of whether specific treatments occurred before or after particular events~\cite{zhang2018text}. Value error is the second most common type, accounting for 30.5\% of inconsistencies. This category includes clinically important information such as laboratory values, vital signs, medication dosages, and output measurements. Even when the same clinical entity appears in both the note and the table, discrepancies in its associated time or value may lead to different clinical interpretations. These findings highlight the importance of detecting both temporal and value-level inconsistencies to support reliable interpretation of clinical data.

\paragraph{Mismatch Patterns by Number of Columns}~
Figure~\ref{app_fig:number_incon} presents the number of inconsistent columns observed for each entity. Most inconsistencies are confined to a single column, representing 81.7\% of inconsistent entities. Cases involving two and three inconsistent columns account for 14.2\% and 3.2\%, respectively, while inconsistencies spanning four or more columns are relatively rare. This pattern indicates that note--table discrepancies are more likely to arise from partial attribute-level mismatches than from complete disagreement at the entity level.

\paragraph{Patterns of Inconsistency and Missing Evidence}~
We further compare column inconsistency with missing evidence, as shown in Figure~\ref{app_fig:incon_om}. The two discrepancy types occur at similar rates, accounting for 51.3\% and 48.7\% of cases, respectively. This result suggests that note--table discrepancies are not limited to cases where structured table values conflict with narrative documentation. A comparable proportion of discrepancies also arises when information described in clinical notes lacks corresponding structured evidence. Notably, the distribution of these two discrepancy types varies across note categories. In discharge summaries, column inconsistency and missing evidence appear in nearly equal proportions (49.9\% and 50.1\%, respectively), suggesting that discharge summaries may contain both information that conflicts with structured tables and information that cannot be verified within them. In contrast, nursing notes show a higher proportion of missing evidence (57.5\%), indicating that bedside observations and nursing interventions are often not fully captured in structured tables. Conversely, physician notes exhibit a higher proportion of column inconsistency (58.8\%), implying that discrepancies in physician documentation are more likely to arise from attribute-level conflicts between note narratives and existing structured records, rather than from the absence of supporting evidence. If such omissions or inconsistencies persist, downstream systems that rely solely on structured data may fail to capture important aspects of the patient’s condition or treatment trajectory, or may instead depend on incorrect evidence. Therefore, it is essential to verify both whether the information described in clinical notes is adequately supported by structured tables and whether existing structured evidence is fully consistent with the narrative documentation.

\paragraph{Inconsistency Patterns by Note Type}~
Figure~\ref{app_fig:incon_per_note} shows that the distribution of column-level inconsistency differs across note types. In discharge summaries, time errors account for the largest proportion, at 54.1\%. Because discharge summaries provide a comprehensive overview of diagnoses, tests, treatments, and clinical course at the time of discharge, temporal inconsistencies can affect the interpretation of the patient’s overall hospitalization and treatment history~\cite{patel2019communication}. Moreover, discharge summaries are used to communicate the hospitalization course during post-discharge care or patient transfer~\cite{mathiesen2024effect}, making temporal accuracy particularly important for downstream clinical interpretation. In nursing notes, value errors are the most prevalent inconsistency type, accounting for 52.3\%. Nursing notes typically include bedside-level information such as patient observations, vital signs, intake and output, pain assessments, and responses to interventions. Because these data are directly related to patient monitoring and nursing care, value-level inconsistencies can lead to incorrect assessments of patient status when structured tables are used as evidence. In physician notes, time errors and value errors constitute the majority of inconsistencies, accounting for 50.1\% and 37.9\%, respectively. Physician notes document daily patient progress, interpretations of test results, treatment responses, and medication changes. Therefore, both temporal and value information are central to verifying whether clinical decisions and treatment trajectories are accurately aligned with evidence from structured tables.

\section{Annotation Protocol}
\label{app:annotation_protocol}
\subsection{Task}
\begin{figure*}[t]
    \centering
    \includegraphics[width=\textwidth]{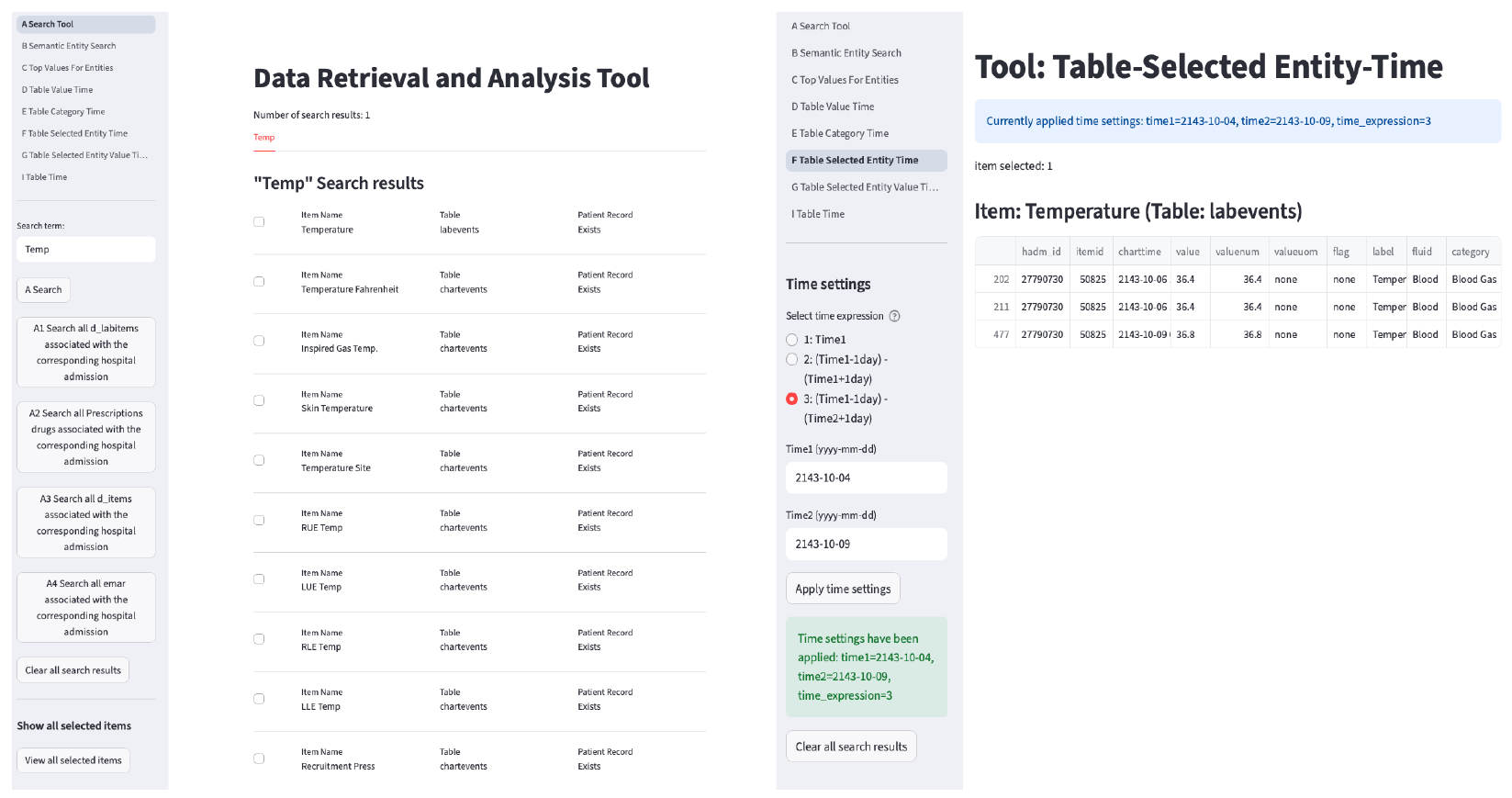}
    \vspace{-5mm}
    \caption{A screenshot of Streamlit provided to labelers for annotation.
}
    \vspace{-3mm}
    \label{fig:screenshot}
\end{figure*}

Identify entities within clinical notes and compare their associated values with actual tables to detect discrepancies between the notes and the tables. 
For efficient labeling, we used Streamlit\footnote{\url{https://streamlit.io/}}, and a screenshot of the interface is shown in Figure~\ref{fig:screenshot}.

\subsection{Annotation Protocols}
\subsubsection{Definition of Clinical Note}
To ensure accurate labeling, it is important to understand the types of clinical notes used in our study. These include the Discharge Summary, Physician Note, and Nursing Note.
\paragraph{Discharge Summary}
The Discharge Summary is written at the time of the patient's discharge, summarizing the events during their hospitalization. It may include information from before admission, such as past medical history, as well as details from after discharge. For example, information about "admission medications" may be recorded to continue treatment for medications the patient was taking prior to hospitalization.
\paragraph{Physician Note}
The Physician Note is written by the physician during daily rounds. It describes the patient's condition and outlines the next steps for diagnosis and treatment.
\paragraph{Nursing Note}
The Nursing Note is written by a nurse and documents the patient’s condition. These notes are often recorded multiple times a day to provide ongoing updates.

\subsection{Definition of Entity}
The goal of entity extraction in this study is to extract all entities that can be matched to the item names from 14 clinical tables, including D\_ITEMS, DIAGNOSES\_ICD, D\_ICD\_DIAGNOSES, D\_ICD\_PROCEDURES, D\_LABITEMS, INPUTEVENTS\_MV, CHARTEVENTS, MICROBIOLOGYEVENTS, INPUTEVENTS\_CV, OUTPUTEVENTS, LABEVENTS, PROCEDURES\_ICD, PRESCRIPTIONS, and PROCEDUREEVENTS\_MV.

\paragraph{Entity Group Considerations} 
In particular, when general drug categories such as ``Antibiotics'' or ``Beta-blockers'' or test panels like ``ABG'' or ``Chem-7'' appear in the clinical notes, they are expanded into detailed items for analysis. For example, if ``Chem-7'' is mentioned, the entity values from the LABEVENTS table, such as Sodium, Potassium, Chloride, Bicarbonate, Blood Urea Nitrogen, Creatinine, and Glucose, should be compared with each respective value. To find these detailed items, searches are limited to four established sources: MedlinePlus~\footnote{\url{https://MedlinePlus.gov/}}, Cleveland Clinic~\footnote{\url{https://my.clevelandclinic.org/}}, Mayo Clinic~\footnote{\url{https://www.mayoclinic.org/}}, and UpToDate~\footnote{{\url{https://www.uptodate.com/contents/search}}}. If the search results are insufficient or additional medical knowledge is required, please consult with a physician.

\paragraph{Entity Mapping in Clinical Notes}
In clinical notes, information is often written as free text, which means the entities might not always be clearly listed or categorized in the tables. For example, a note that mentions ``headaches'' might correspond to different entries in the database, such as ``pain location'' or ``type of pain.'' Even if these entries aren’t explicitly labeled, it's essential to match the data to the correct item in the tables whenever possible. To map entities accurately, it’s important to understand how each entity is stored in the database. Since information in clinical notes can sometimes be incomplete or unclear, you need to interpret it carefully and know how to connect it correctly. If something is unclear, it’s important to consult with a physician to ensure the mapping is done correctly.

\begin{figure*}[t]
    \centering
    \includegraphics[width=0.6\textwidth]{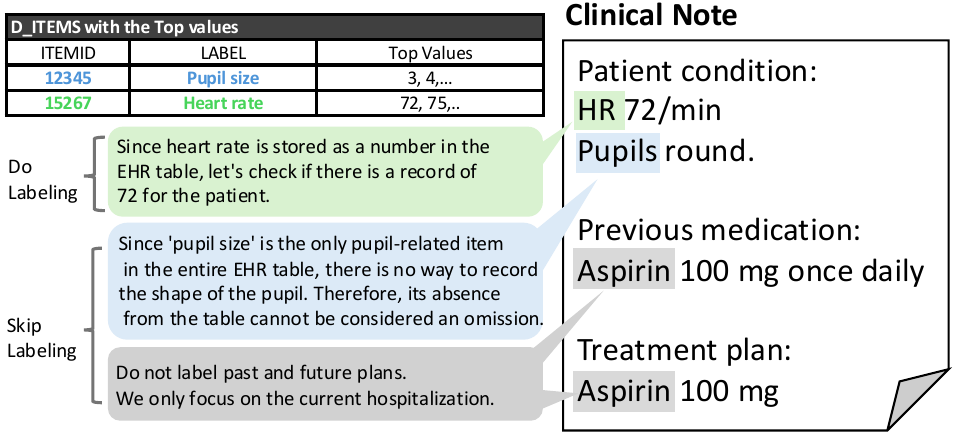}
    \caption{Scope of \dtname.}
    \vspace{-3mm}
    \label{fig:entity_scope}
\end{figure*}

\paragraph{Entity Scope}
To prevent incorrect discrepancies, we do not extract the following types of information as entities (see Figure~\ref{fig:entity_scope}):
\begin{itemize}
    \item \textbf{Past Information}: The past medical history, family history, previous hospital admissions, emergency room visits, and medications administered during hospitalization recorded in clinical notes are all considered past information. Since this information is not stored in the tables, it is not extracted as entities.
    \item \textbf{Future Plans}: Information such as `discharge plans' or `next steps to be taken' mentioned in discharge summaries or physician notes are categorized as future plans. These plans may not be executed due to changes in the patient's condition, and therefore, are not considered discrepancies if not carried out. Future plans are not extracted as entities, including those recorded as `plan' in nursing notes.
    \item \textbf{Information Unrelated to Database}: If the information recorded in clinical notes is not linked to a specific record or item in the tables, it is not extracted as an entity. Additionally, information such as transfer details, which are not related to the tables we handle, or items that only mention total amounts without specifying what those amounts refer to, are also not extracted.
\end{itemize}

\paragraph{Insurance-related Entities}
Items recorded under ``Discharge Diagnoses'' or ``Major Surgical/Invasive Procedures'' in the discharge summary are typically written for insurance claim purposes by hospitals. Entities listed under these items should be limited to discrepancy checks related to insurance claims, specifically for PROCEDURES\_ICD, DIAGNOSES\_ICD, D\_ICD\_DIAGNOSES, and D\_ICD\_PROCEDURES.

\subsection{Definition of Entity Attribution}
\paragraph{Time}
Structured tables in EHR are essentially time-series data, making the timing of events crucial for understanding a patient's clinical journey. When reviewing clinical notes, it is important to determine the time an event occurred. For example, if a note says ``3 days after surgery,'' the surgery date should be extracted, converted into a standard time format, and then used to calculate the exact date that is 3 days after the surgery. The key is to carefully consider the context of the note and extract the most accurate time possible. While the guidelines below are helpful, the main goal is to interpret the context of the note and determine the most appropriate timestamp.

\begin{itemize}
\item \textbf{Handling Ambiguous Time Information}: If explicit time information is not provided, the time is inferred based on the note's nature. For example, a discharge summary typically summarizes the patient's admission and discharge records. If the exact time is unclear, the entities in the discharge summary should be checked against the tables within the patient's hospitalization period, and if they match, it can be considered consistent. Physician or nursing notes are usually written daily, so in this case, the charting date is treated as the event date. That is, if the entity information is recorded in the tables on the charting date, it can be considered consistent.
\item \textbf{Mapping Time Based on Admission and Discharge Dates}: When specified as ``admission date'' and ``discharge date,'' the time expression may vary depending on the physician. Therefore, a reasonable medical range should be applied, for instance, one day before admission and one day before discharge. Consistency checks should be conducted within this time frame.
\item \textbf{Medication Timing}: For medication-related tables, such as inputevents or prescription tables, both the start and end times are recorded. If both the start and end times are explicitly provided, these should be checked against the tables' startdate or starttime and enddate or endtime fields to ensure they match accurately. If a duration is recorded in the clinical note, verify that the medication was administered correctly during the specified period.
\item \textbf{Verifying Time-Related Terms in EHR}: When terms like ``morning'' or ``midnight'' are used in a clinical note, verify whether the records exist accurately within the corresponding time frame.
\end{itemize}

\begin{itemize}
\item \textbf{Note}: While these guidelines should be followed, the main goal is to understand the context of the note and extract the most accurate timestamp possible.
\end{itemize}

\subsection{Discrepancy Detection Process}
The core of this task is to compare the entities extracted from clinical notes with the table records to check for discrepancies. To do this, the following three main approaches can be used:

\paragraph{Matching Based on Common Sense}
    Even if an entity in the clinical note does not exactly match the one in the tables, it can still be considered a match based on common sense. For example, the clinical note may mention ``hand,'' while the tables record ``finger.'' Since both refer to the same part of the body, they can be considered a match.
    
\paragraph{Matching Based on Medical Knowledge}
    Clinical notes describe a patient’s condition in free text, while the tables use a standardized format. For instance, if the note says `edema 2+,' it might be recorded in the tables as `palpable edema.' In such cases, it is essential to understand the medical meaning of these terms (\eg 2+ and palpable) and refer to authoritative sources such as UpToDate, MedlinePlus, Mayo Clinic, and Cleveland Clinic to make an accurate match. If these sources do not provide sufficient information, it is important to consult with a physician for clinical interpretation.
    
\paragraph{Exact Matching}
    Some clinical note entries may be copied from the tables based on medical practices. In this case, the entities in the clinical note may match the table records exactly. For example, if the note mentions a WBC count of 10.0, the same value may appear in the tables. However, EHR system errors could lead to discrepancies, so it’s crucial to carefully check for consistency.

\subsection{Single or Multiple Row Matching}
\paragraph{Single Row Matching}
    Some entities can be compared with a single row in the tables to check for discrepancies. This applies to events that occur at a specific point in time. For example, if the note says ``WBC stable,'' and the tables show a stable value, it can be considered a match. In this case, as the event happens at a single point in time, finding this record in the tables once would be enough to confirm the match. However, time should also be carefully considered in these cases.
    
\paragraph{Multiple Row Matching}
    Clinical notes often describe a patient’s condition over time. When this happens, the trend across multiple records in the tables needs to be checked. For instance, if the note says, ``No fever from hospital day 3 to 5, but fever started on day 6,'' the table records must show no fever from days 3-5, and the fever should be recorded starting from day 6.
    
\begin{itemize}
\item \textbf{Note}: Values like blood pressure (BP) can be represented as follows: 100/60 (80) - 140/100 (120). In this notation, the value in parentheses represents the mean value, while the part before the hyphen indicates the minimum blood pressure, and the part after the hyphen represents the maximum blood pressure range. This notation is used when recording a patient's blood pressure measurement to provide both the average value along with the minimum and maximum values. When comparing with table data, the mean value should be searched as the mean BP in the table, while the minimum and maximum values should be considered as the BP measurement range.
\end{itemize}

\subsection{Example}
\begin{itemize}
\item WBC 20.0 *
    \begin{itemize}
    \item At this point, you need to confirm that the numeric value of WBC is 20.0, and that the FLAG column correctly shows the value as ``ABNORMAL'' in the table.
    \end{itemize}
\end{itemize}
\begin{itemize}
\item Anemia on admission
    \begin{itemize}
    \item Here, you need to check that the hemoglobin value (valuenum) listed in the table falls within the range indicating anemia, from the day before admission to the day after admission.
    \end{itemize}
\end{itemize}
\begin{itemize}
\item Sputum culture identifying Streptococcus sensitive to Cefazolinex.
    \begin{itemize}
    \item In this case, the table should show that the specimen is ``sputum'' and the organism is ``Streptococcus''. Additionally, you need to confirm that the sensitivity test result (interpretation) for ``Cefazolinex'' (ab\_name) shows ``sensitive.''
    \end{itemize}
\end{itemize}

\subsection{Consistency Check}
Labeling should be strictly limited to consistent and inconsistent. A label of consistent indicates that all information described in the clinical note is accurately aligned with the table. If even a single column contains conflicting information or if there is no supporting record in the table, the case should be labeled as inconsistent. 
In addition, please specify the rows in the table that serve as evidence for your consistent or inconsistent decision. Notes for Consistency Checking: 
\begin{itemize}
\item \textbf{Use of medical knowledge}: If medical knowledge is used to make a judgment, please provide the corresponding reference. Examples of medical knowledge used in \dtname are provided in Table~\ref{tab:clinical_ref}.
\end{itemize}
\begin{itemize}
\item \textbf{Use of commonsense knowledge}: If commonsense knowledge is used, please explicitly state that commonsense knowledge was applied. For example, if the note states that the patient injured their right arm, while the table records an injury to the right finger, this may be considered consistent based on commonsense knowledge.
\end{itemize}

\begin{longtable}{p{3cm} p{2cm} p{7.5cm}}
\caption{Clinical Reference Table} \label{tab:clinical_ref} \\

\hline
\textbf{Subject} & \textbf{Reference} & \textbf{Content} \\ \hline
\endfirsthead

\hline
\textbf{Subject} & \textbf{Reference} & \textbf{Content} \\ \hline
\endhead

Normal Heart Rate for Adults (BPM) & Mayo Clinic & A normal resting heart rate for adults is 60--100 beats per minute; well-trained athletes may have lower rates \\ \hline
Regular Temperature & Mayo Clinic & Around 37\textdegree{}C is normal, and 38\textdegree{}C or higher is considered a fever. \\ \hline
Stool Color & Mayo Clinic & Stool color is mainly influenced by diet and bile, and colors such as brown, green, and yellow are generally considered normal, whereas bright red or black stool may indicate possible bleeding. \\ \hline
WBC Normal Range & MedlinePlus & The normal range is approximately 4,500 to 11,000 per microliter; lower levels may indicate weakened immunity or underlying disease, while higher levels may suggest infection, inflammation, or cancer. \\ \hline
Levofloxacin & MedlinePlus & Levofloxacin is an antibiotic used to treat bacterial infections, not viruses like colds or flu, and carries risks of serious side effects that require caution. \\ \hline
Respiratory Range & Cleveland Clinic & RR is normally 12--18 breaths per minute, and values below 12 or above 25 at rest are considered abnormal. \\ \hline
Melena & Cleveland Clinic & Melena (black stool) is a symptom of internal bleeding, usually in your upper gastrointestinal (GI) tract. The blood turns black as it travels through your digestive system before coming out in your poop. \\ \hline
Tachypneic & Cleveland Clinic & Tachypnea is quick, shallow breathing. This makes you feel like you're not getting enough air. This symptom can affect anyone at any age and is common among newborns and people with respiratory conditions. Treating the underlying cause prevents this symptom. \\ \hline
Normal Sinus Rhythm & Uptodate & NSR is a normal rhythm originating from the sinoatrial node, while sinus tachycardia has the same origin but with a heart rate above 100 bpm. \\ \hline
Hypertension Criteria & MedlinePlus & Hypertension is when blood pressure is consistently 130/80 mmHg or higher. \\ \hline
Furosemide & MedlinePlus & Furosemide is a diuretic (``water pill'') used to treat high blood pressure and reduce excess fluid (edema) in the body. \\ \hline
Blood Oxygen Level & Cleveland Clinic & Normal blood oxygen saturation is 95--100\%; below 92\% requires caution, and 88\% or lower is an emergency level. \\ \hline
Clobetasol Ointment & Cleveland Clinic & Clobetasol propionate is an ointment you can rub on your skin to treat eczema and psoriasis. It reduces swelling, redness, itching and rashes caused by these skin conditions. It's a type of topical steroid medication. Brand names of this medication are Cormax, Embeline and Temovate. \\ \hline
T3 normal range & Cleveland Clinic & The normal T3 range in adults is 79 to 165 ng/dL for total T3 and 2.3 to 4.1 pg/mL for free T3. \\ \hline
Atrial fibrillation (AFib) & Mayo Clinic & Atrial fibrillation, also called AFib. This is the most common type of tachycardia. Chaotic, irregular electrical signals start in the upper chambers of the heart, called the atria. \\ \hline
Beta Blockers & Cleveland Clinic & Beta-blockers are a class of medicines most commonly used for problems involving your heart and circulatory system. They can also help treat conditions related to your brain and nervous system. Beta-blockers work by slowing down certain types of cell activity. This can help manage your blood pressure, heart rate and more. \\ \hline
\end{longtable}

\section{Table-Exploration Tools}
\label{app:table_exploration_tools}
This process produced eight table-exploration tools that support efficient exploration of complex EHR databases and are organized into three functional categories:

\subsection{Entity-to-Table Item Alignment}
The tools in this category support the alignment of entities mentioned in clinical notes with corresponding items in structured tables. The same clinical information may appear under different names or levels of abstraction (\eg ``White Blood Cells'' and ``WBC''), so these tools retrieve potentially relevant table items based on both lexical similarity and conceptual relatedness.

\begin{figure*}[t]
    \centering
    \includegraphics[width=0.7\textwidth]{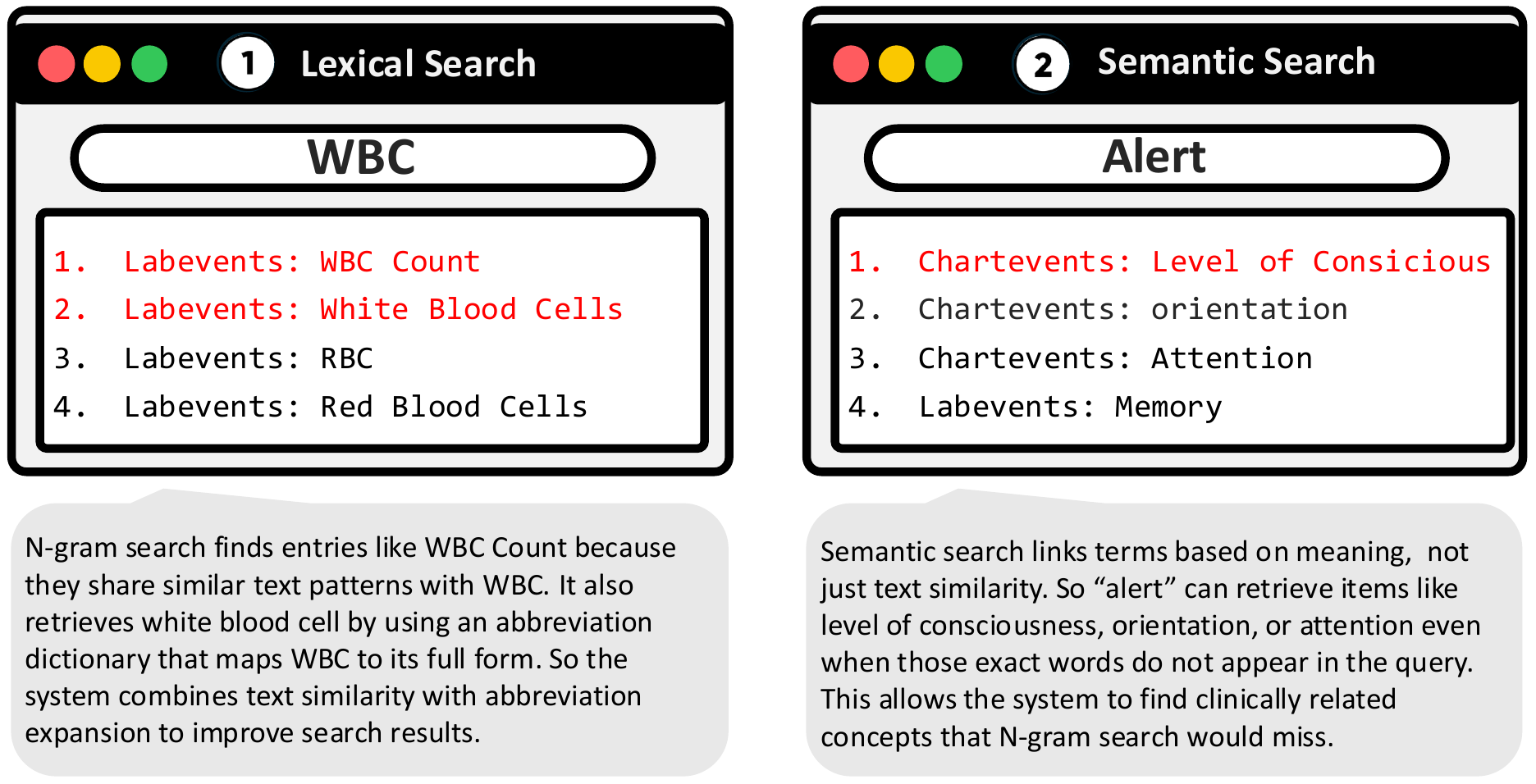}
    \vspace{-2mm}
    \caption{Conceptual illustration comparing N-gram–based lexical search and semantic search, highlighting how string similarity versus conceptual relevance leads to different retrieval results.
}
    \vspace{-2mm}
    \label{fig:app_tool1}
\end{figure*}

\paragraph{Lexical\_Search}~~This tool employs an N-gram–based approach to retrieve items from structured tables that are lexically similar to a given query entity (see Figure~\ref{fig:app_tool1}-(1)). Rather than relying solely on simple string matching, it incorporates the C4-WSRS medical abbreviation dataset~\cite{rajkomar2022deciphering}, allowing abbreviations such as ``WBC'' or ``BP'' to retrieve their full forms, ``White Blood Cell'' and ``Blood Pressure.'' This enables more accurate retrieval by accounting not only for surface-level text similarity but also for abbreviation expansion.

\paragraph{Semantic\_Search}~~This tool supports semantic search to capture similarities that are difficult to detect with lexical methods alone (see Figure~\ref{fig:app_tool1}-(2)). It leverages the \textit{Gemini 2.5 Flash} model to identify items that are semantically similar to a given query entity by analyzing contextual meaning rather than surface-level text. For example, the term ``alert,'' which describes a patient’s condition, is conceptually related to ``level of consciousness,'' but this relationship may not be identified through N-gram matching. This tool addresses this limitation by considering context and meaning, linking expressions that differ lexically but represent the same or similar clinical concepts.

\begin{figure*}[t]
    \centering
    \includegraphics[width=0.7\textwidth]{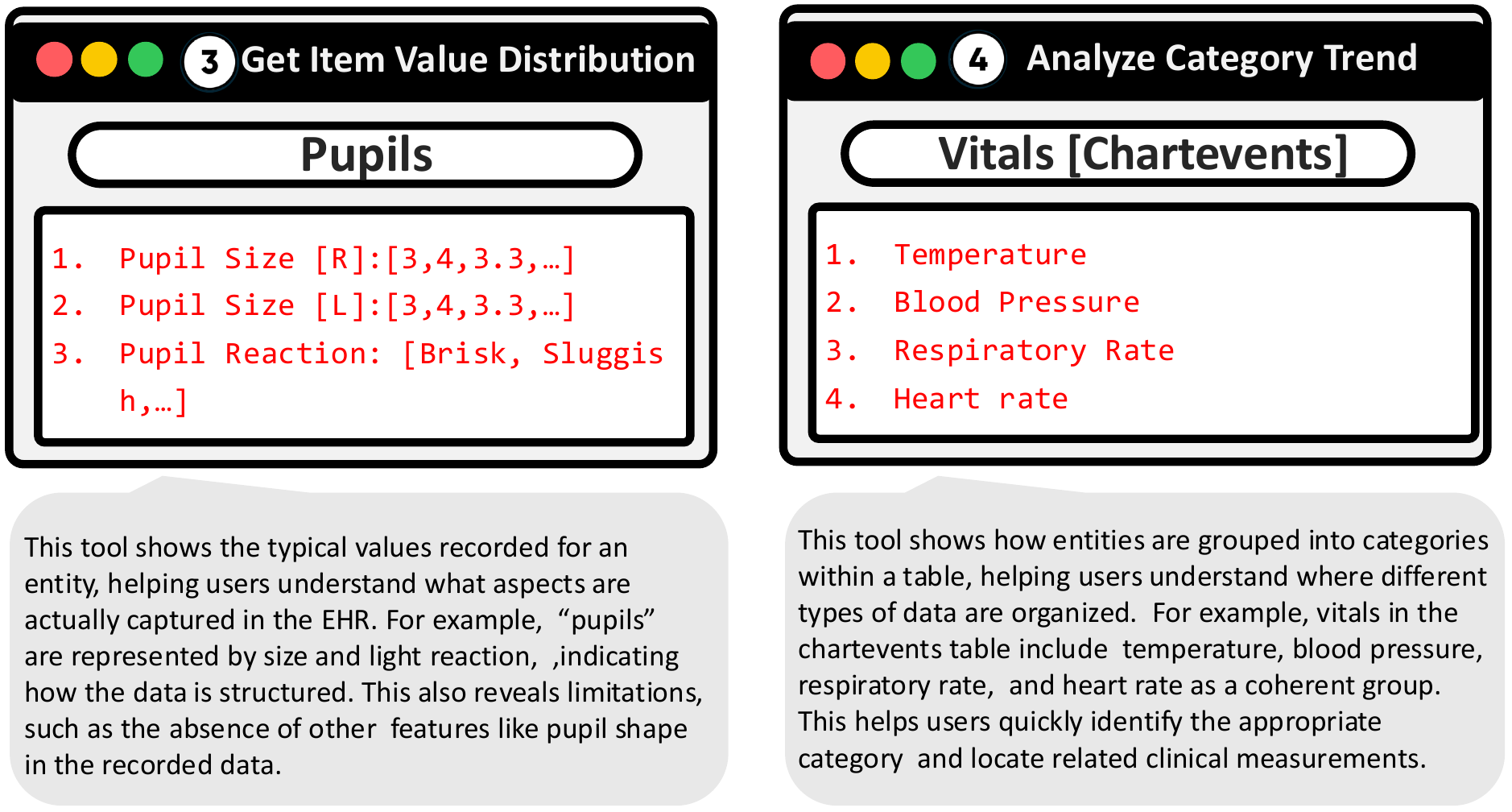}
    \vspace{-2mm}
    \caption{Conceptual illustration comparing Get Item Value Distribution and Analyze Category Trend.
}
    \vspace{-2mm}
    \label{fig:app_tool2}
\end{figure*}

\begin{figure*}[t]
    \centering
    \includegraphics[width=0.7\textwidth]{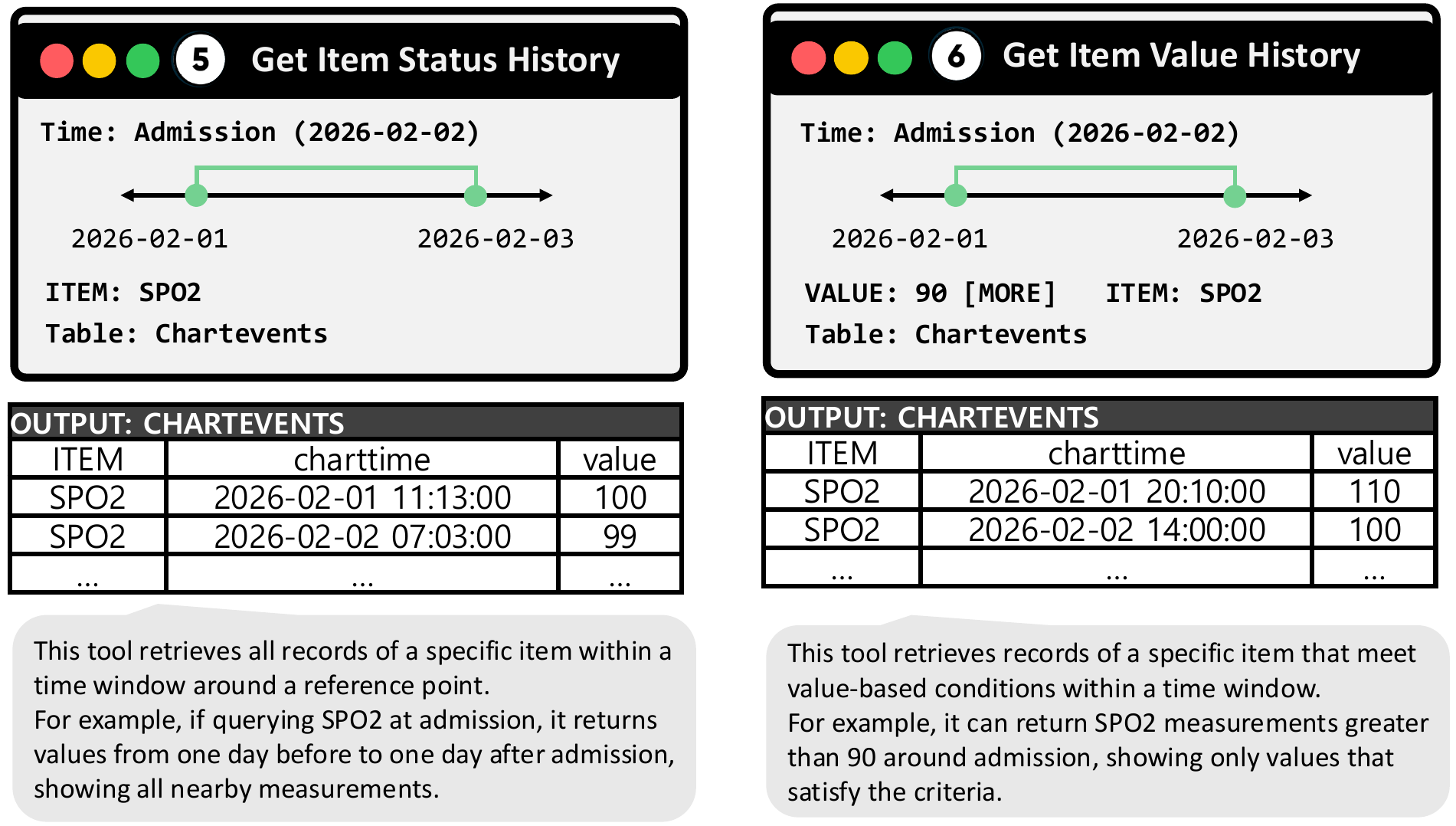}
    \vspace{-2mm}
    \caption{Conceptual illustration comparing Get Item Status History and Get Item Value History.
}
    \vspace{-2mm}
    \label{fig:app_tool3}
\end{figure*}

\begin{figure*}[t]
    \centering
    \includegraphics[width=0.7\textwidth]{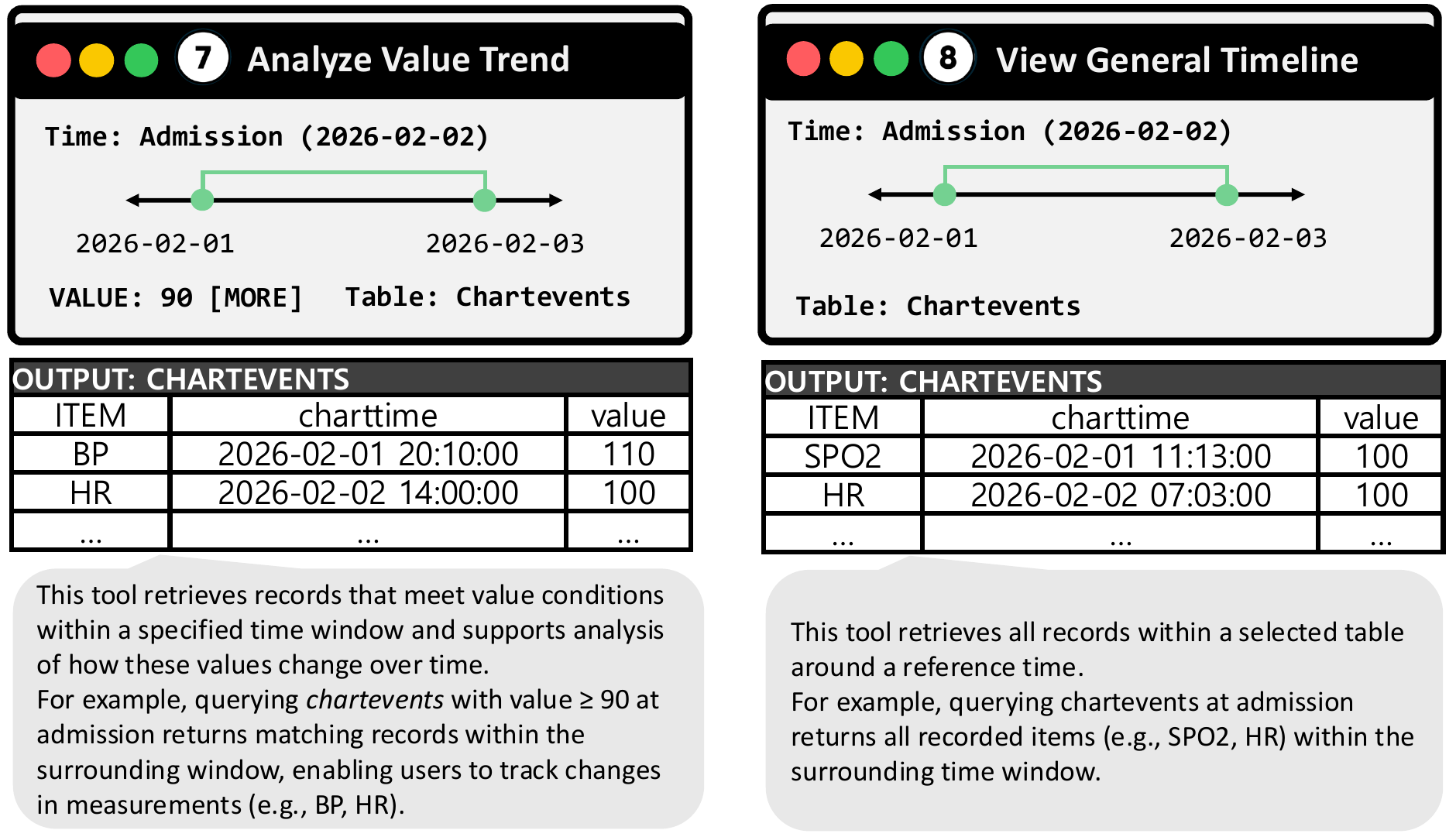}
    \vspace{-2mm}
    \caption{Conceptual illustration comparing Analyze Value Trend and View General Timeline.
}
    \vspace{-2mm}
    \label{fig:app_tool4}
\end{figure*}

\subsection{Database Exploration and Value Profiling}
The tools in this category support exploration of the EHR database schema and content. Since clinical concepts may be distributed across multiple tables, the tools support exploration of relevant table groups and summarize typical values for each item, enabling annotators to quickly interpret the role of different fields.

\paragraph{Get\_Item\_Value\_Distribution}~~This tool provides insight into the distribution of values associated with specific items within tables (see Figure~\ref{fig:app_tool2}-(3)). It allows users to view the top K most frequent values for each item, helping to characterize the nature of the data. In this study, K was set to 10. For instance, if ``SBP'' values frequently appear as 110, 120, or 130, this indicates that the item represents continuous numerical data. This tool enables users to determine whether an item is numerical, categorical, or follows a specific pattern, and to assess whether the retrieved item appropriately reflects the information recorded in clinical notes.

\paragraph{Analyze\_Category\_Trend}~~This tool helps users understand the structure of items across multiple tables in a database (see Figure~\ref{fig:app_tool2}-(4)). Since items are distributed across tables based on their characteristics, it is not always intuitive to determine where a specific item resides. This tool analyzes how each table is organized into categories and what items belong to each category, providing insight into the location and context of entities. For example, the chartevents table may include categories such as Labs and General. The Labs category contains items like ``anion gap'' or ``CK-MB,'' while the General category includes items related to consciousness, such as ``Level of Consciousness'' or ``Oriented.'' This helps users more efficiently identify the appropriate table for a given entity.

\subsection{Temporal and Conditional Record Retrieval}
The tools in this category support verification of clinical statements that involve temporal changes or specific conditions. The tools allow annotators to retrieve records from structured tables based on time windows and value constraints, enabling inspection of whether the structured data support trends or events described in clinical notes.

\paragraph{Get\_Item\_Status\_History}~~This tool enables users to examine entity information over time (see Figure~\ref{fig:app_tool3}-(5)). Time can be specified in three different ways to refine the search. First, users can query based on an exact standard timestamp, such as ``06-24,'' to check whether records exist at a specific point in time. This approach is suitable when explicit time information is directly recorded in the data. Second, this tool supports searches based on time expressions, such as ``admission,'' which are described narratively in clinical notes. Because these expressions do not correspond to precise timestamps and must be interpreted from context, the search is performed by defining a time window around the inferred point, typically extending from one day before to one day after the reference time, in order to retrieve relevant surrounding records. Third, this tool allows searches based on duration. Users can define a start time (time1) and an end time (time2), and for more robust retrieval, the search range is typically expanded to include the period from one day before the start time to one day after the end time, ensuring that all relevant data within the interval are captured. For example, when a user searches (Admission, spo2) using Get\_Item\_Status\_History, the tool retrieves spo2 results recorded during that admission.

\paragraph{Get\_Item\_Value\_History}~~This tool supports more fine-grained queries than Get\_Item\_Status\_History (see Figure~\ref{fig:app_tool3}-(6)). It allows users to specify not only the item but also its associated values as search conditions. By using operators such as ``more,'' ``less,” and ``between,'' users can define value ranges more precisely and retrieve data that meet specific criteria. For example, when a user searches (Admission, spo2, 90[more]) using Get\_Item\_Value\_History, the tool retrieves only spo2 results greater than 90 recorded during that admission.

\paragraph{Analyze\_Value\_Trend}~~This tool enables the analysis of value trends over time, independent of specific items (see Figure~\ref{fig:app_tool4}-(7)). Rather than focusing on values at a single time point, it examines how values evolve across time, capturing patterns such as increases, decreases, or stability. Users can specify a time range to analyze trends within a particular interval, allowing for a more contextual understanding of value changes. In addition to absolute values, this tool considers dynamic characteristics such as the rate of change and variability. It is designed to move beyond item-specific status queries and instead support a broader understanding of temporal value patterns across the data. For example, when a user searches (Admission, 90[more]) using Analyze\_Value\_Trend, the tool retrieves all results greater than 90 recorded during that admission.

\paragraph{View\_General\_Timeline}~~This tool allows users to retrieve all records associated with a specific time point within a selected table (see Figure~\ref{fig:app_tool4}-(8)). For example, if ``2026-05-03'' is set as the reference time for the chartevents table, all patient records documented in the chartevents table on that date can be retrieved.

\section{Inter Annotator Agreement}
\label{app:IAA}
After completing the annotation process, we measured inter-annotator agreement by comparing the annotations produced by two annotators for each note. For entity recognition, we defined recall and precision based on the proportion of overlapping entities identified by the two annotators, and computed the F1 score as their harmonic mean. The resulting F1 score was 0.897, indicating a high level of agreement.
In addition, for entities extracted by both annotators, we compared their consistency labels and found an agreement rate of 0.888. These high agreement scores support the overall quality and reliability of the annotation.

\section{Detailed Implementation of \frname}
\label{app:detail_framework}
\subsection{Entity Extraction}
\label{app:entity_extraction}
\frname identifies entities in clinical notes that can be verified using EHR data. It follows a two-step approach: (1) entities corresponding to items recorded in the patient's structured tables ($E_{patient}$), and (2) entities defined in the EHR schema but absent from the patient's structured records ($E_{ontology}$). The final anchor entity set is defined as $E_{extract}=E_{patient}\cup E_{ontology}$.
\subsubsection{Patient-specific Extraction}
\label{app:patient_specific}
\begin{figure*}[t]
    \centering
    \includegraphics[width=0.8\textwidth]{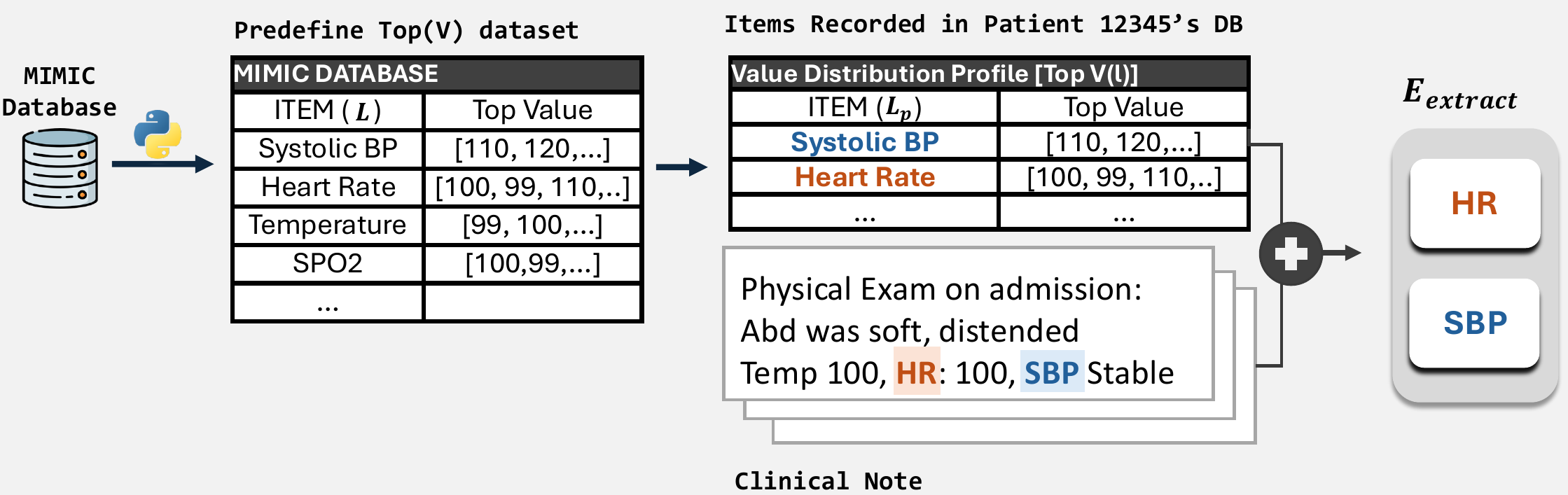}
    \vspace{-1mm}
    \caption{Overview of patient-specific entity extraction.
}
    \vspace{-2mm}
    \label{fig:app_patient}
\end{figure*}

As illustrated in Figure~\ref{fig:app_patient}, we first predefine the global item set $L$ and the corresponding value distribution profile $TopV(l)$ for each item using the full MIMIC-III~\cite{PhysioNet-mimiciii-1.4} database. For a given patient $P$, we then construct a patient-specific subset $L_P \subseteq L$ by selecting only the items recorded in the patient’s structured tables. Notably, the value distribution $TopV(l)$ is not recomputed per patient but directly reused from the globally defined statistics. For each clinical note segment $x_t$, the LLM is provided with $x_t$, $L_P$, and $\{TopV(l)\}_{l \in L_P}$, and extracts entity names that correspond to items in $L_P$. The final patient-specific entity set is obtained by aggregating the extracted entities across all segments, i.e., $E_{\mathrm{patient}} = \bigcup_{x_t \in X} E^{\mathrm{pat}}(x_t)$.

\subsubsection{Ontology-Guided Extraction}
\label{app:ontology_guided}

\begin{figure*}[t]
    \centering
    \includegraphics[width=\textwidth]{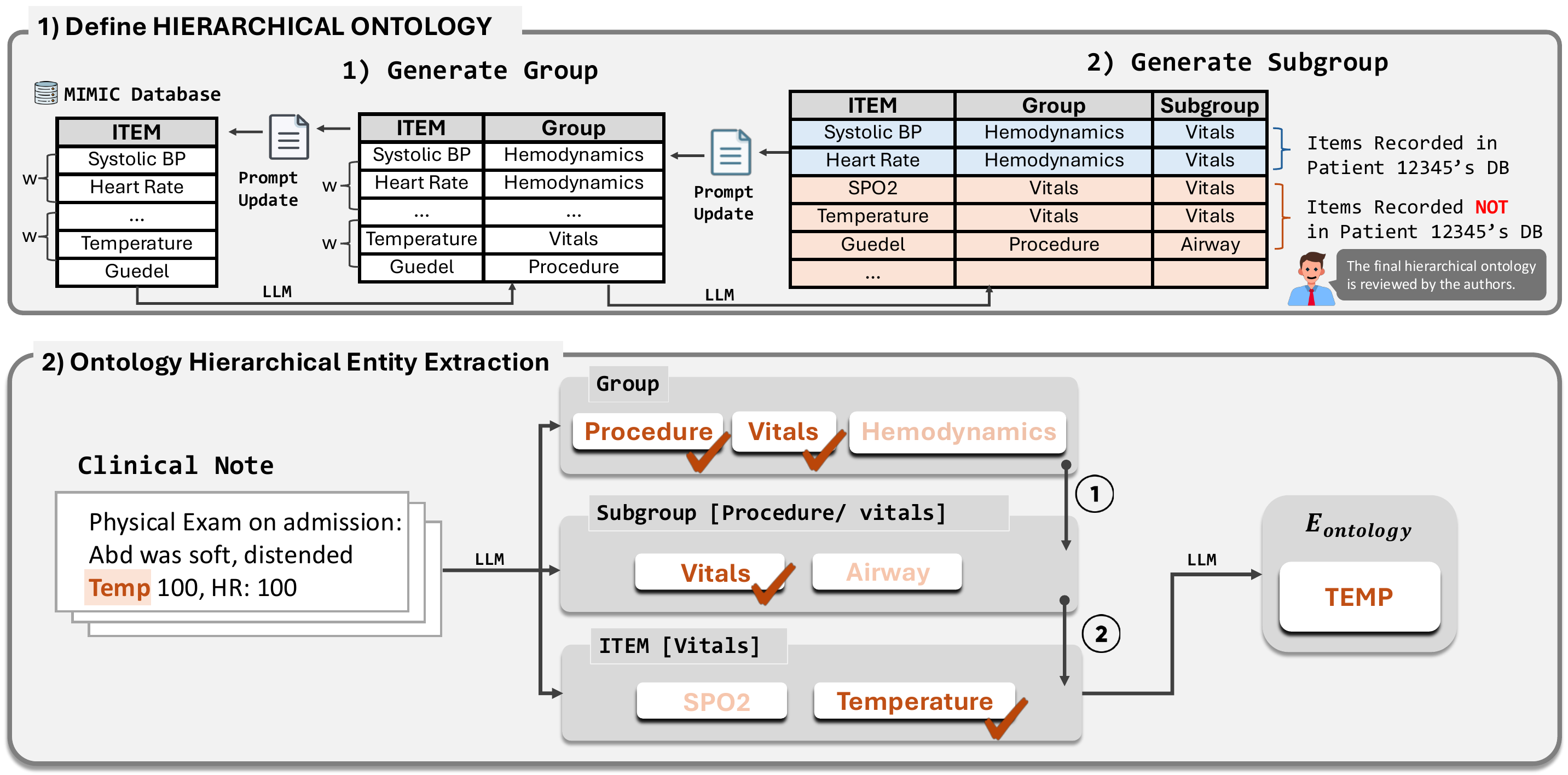}
    \vspace{-5mm}
    \caption{Overview of ontology-guided entity extraction.
}
    \vspace{-2mm}
    \label{fig:app_ont}
\end{figure*}

Figure~\ref{fig:app_ont} illustrates the construction of the predefined hierarchical ontology and its use in ontology-guided entity extraction. This section first describes how the predefined hierarchical ontology is constructed and then explains how the constructed ontology is used to extract ontology-based entities from clinical notes.

\paragraph{Predefined Hierarchical Ontology Construction}
In this study, we automatically construct a hierarchical ontology using an LLM, based on the full set of items appearing in the structured tables. Since real-world EHR tables contain more than 10,000 items, it is impractical to provide the entire item set to the LLM at once. Therefore, we divide the items into smaller batches using a sliding window of size $w$ and process them sequentially. At each step, the LLM receives the $w$ items in the current window and assigns each item to an appropriate group. The prompt also includes the list of groups generated from previous windows. This allows the LLM to map newly observed items to existing groups, or to create a new group when none of the existing groups is suitable. By repeating this prompt-update process over the entire item set, we progressively construct the group-level ontology.
For example, suppose a window contains items such as \textit{Systolic BP}, \textit{Heart Rate}, \textit{Temperature}, and \textit{Guedel}. The LLM may assign \textit{Systolic BP} and \textit{Heart Rate} to the \textit{Hemodynamics} group, \textit{Temperature} to the \textit{Vitals} group, and \textit{Guedel} to the \textit{Procedure} group. When new items are processed in subsequent windows, the LLM refers to the previously generated groups and either assigns the new items to existing groups or creates additional groups when necessary.

After constructing the group-level ontology, we build the subgroup-level ontology using the same sliding-window strategy. In this stage, the LLM receives item--group pairs rather than item names alone. That is, each input includes both an item and its assigned group, such as \textit{Systolic BP}--\textit{Hemodynamics}. The LLM then maps each item--group pair to an appropriate subgroup, or creates a new subgroup if no existing subgroup is suitable. The generated subgroup list is also included in subsequent prompts, allowing the subgroup-level ontology to be progressively expanded over the full item set.
For example, \textit{Systolic BP}--\textit{Hemodynamics} and \textit{Heart Rate}--\textit{Hemodynamics} may be assigned to the \textit{Vitals} subgroup, whereas \textit{Guedel}--\textit{Procedure} may be assigned to the \textit{Airway} subgroup. The final hierarchical ontology is reviewed by the authors to identify and correct inappropriate mappings or redundant categories. The prompts used for group- and subgroup-level ontology construction are provided in Figure~\ref{map_prmpt_gr_sr}. The full list of groups and subgroups is provided in Table~\ref{apptab:group_subgroup}.

\paragraph{Ontology-Guided Entity Extraction}~Given the predefined hierarchical ontology $O$ and segmented clinical notes $x_t$, LLM extract the ontology-based entity set $E_{\mathrm{ontology}}$. The goal of this step is to recover entities that are mentioned in clinical notes but are absent from the corresponding patient's structured records. Specifically, for each clinical note segment, the LLM first selects the groups that are relevant to the segment. Within the selected groups, it then selects the most relevant subgroups. Finally, item-level entity extraction is performed using only the candidate items associated with the selected subgroups. In other words, instead of searching over the entire EHR item set, the ontology guides the LLM to progressively narrow the candidate space from groups to subgroups and then to item-level entities. For example, suppose a clinical note segment contains the phrase ``Temp: 100''. The LLM first recognizes that ``Temp'' refers to a vital-sign-related mention and may select relevant groups such as \textit{Vitals} or \textit{Hemodynamics}. It then selects the relevant subgroup within the selected groups, such as the \textit{Vitals} subgroup. Finally, extraction is performed using only the item candidates associated with that subgroup. Since this subgroup contains the item \textit{Temperature}, the mention ``Temp'' is mapped to \textit{Temperature} in the predefined ontology. Therefore, \textit{Temp} is added to the ontology-based entity set $E_{\mathrm{ontology}}$.

{
\footnotesize
\renewcommand{\arraystretch}{1.3}
\begin{longtable}{p{0.42\linewidth}|p{0.50\linewidth}}
\caption{\normalsize Groups and subgroups used in the dataset.}
\label{apptab:group_subgroup} \\
\hline
\textbf{GROUP} & \textbf{SUB GROUP} \\ \hline
\endfirsthead
\hline
\textbf{GROUP} & \textbf{SUB GROUP} \\ \hline
\endhead
\hline
\endfoot
\endlastfoot

PROCEDURE & ACCESS \& LINES \\ \cline{2-2}
& AIRWAY \& VENTILATION \\ \cline{2-2}
& RENAL \& DIALYSIS \\ \cline{2-2}
& CARDIAC \\ \cline{2-2}
& DIAGNOSTICS (IMAGING/\allowbreak LABS/\allowbreak CULTURES/\allowbreak SWABS/\allowbreak ECG\textperiodcentered{}EEG) \\ \cline{2-2}
& CARE FLOW, COMMS, SAFETY ADMIN \\ \cline{2-2}
& GI\&IR \\ \cline{2-2}
& NEURO \& ICP \\ \cline{2-2}
& OB/GYN \\ \cline{2-2}
& MSK \\ \cline{2-2}
& GENERAL/PLASTICS/ENT/EYE/DENTAL \\ \hline

MEDICATIONS, FLUIDS \& NUTRITION & FLUIDS \& NUTRITION \\ \cline{2-2}
& CARDIO \& HEMATOLOGY \\ \cline{2-2}
& INFECTIOUS DISEASES \& IMMUNOLOGY \\ \cline{2-2}
& CNS, PAIN \& SEDATION \\ \cline{2-2}
& ENDOCRINE, RENAL \& GI \\ \cline{2-2}
& RESPIRATORY, ENT/OPHTH \& DERM \\ \cline{2-2}
& FORMULATIONS, DEVICES, STUDY CODES \& AMBIGUOUS \\ \hline

OUTPUT AND REMOVAL & DIALYSIS\_ULTRAFILTRATION \\ \cline{2-2}
& URINARY\_OUTPUT \\ \cline{2-2}
& PLEURAL\_CHEST\_TUBES \\ \cline{2-2}
& ABDOMINAL\_PARACENTESIS \\ \cline{2-2}
& CSF\_DRAINAGE \\ \cline{2-2}
& GI\_TUBES\_OUTPUT \\ \cline{2-2}
& WOUND\_LOCAL\_DRAINS \\ \cline{2-2}
& BLOODLOSS\_APHERESIS\_OPS \\ \cline{2-2}
& DRAIN\_DEVICES \\ \cline{2-2}
& BOWEL\_OSTOMY\_OUTPUT \\ \cline{2-2}
& EMESIS\_ORAL\_GASTRIC \\ \cline{2-2}
& HEPATOBILIARY\_PANCREATIC\_DRAINS \\ \cline{2-2}
& LOCATION-BASED OUTPUT \\ \cline{2-2}
& CARDIAC OUTPUT \\ \cline{2-2}
& ADJUSTMENT/AGGREGATED VALUES \\ \cline{2-2}
& AMBIGUOUS VALUES \\ \hline

MICROBIOLOGY & SPECIMEN \\ \cline{2-2}
& ORGANISM \\ \cline{2-2}
& ANTIBACTERIUM \\ \hline

RESPIRATORY / VENTILATION \& GAS MONITORING & RESPIRATORY / VENTILATION \& GAS MONITORING \\ \hline
HEMODYNAMICS \& VITALS & HEMODYNAMICS \& VITALS \\ \hline
LABS / ABG-CHEMISTRY & LABS / ABG-CHEMISTRY \\ \hline
SKIN / WOUNDS \& DEVICES / SAFETY & SKIN / WOUNDS \& DEVICES / SAFETY \\ \hline
NEURO/PAIN/FUNCTIONAL & NEURO/PAIN/FUNCTIONAL \\ \hline
AMBIGUOUS, ADMINISTRATIVE / ADMISSION \& DEMOGRAPHICS & AMBIGUOUS, ADMINISTRATIVE / ADMISSION \& DEMOGRAPHICS \\ \hline
ALLERGY & ALLERGY \\ \hline

DIAGNOSIS & TUBERCULOSIS \\ \cline{2-2}
& ENTERIC \& FOODBORNE BACTERIAL INFECTIONS \\ \cline{2-2}
& PARASITIC \& VECTOR-BORNE PROTOZOAL/TICK-BORNE DISEASES \\ \cline{2-2}
& SEXUALLY TRANSMITTED INFECTIONS (STIS) \& HIV \\ \cline{2-2}
& OTHER BACTERIAL \& SEPTIC INFECTIONS \\ \cline{2-2}
& RESPIRATORY BACTERIAL DISEASES \\ \cline{2-2}
& ZOONOTIC \& MYCOBACTERIAL DISEASES \\ \cline{2-2}
& VIRAL DISEASES \\ \cline{2-2}
& PRION \& SLOW VIRUS DISEASES \\ \cline{2-2}
& NON-INFECTIOUS / NEOPLASTIC DISEASES \\ \hline

LABORATORY & HEMATOLOGY (CEREBROSPINAL FLUID (CSF)) \\ \cline{2-2}
& HEMATOLOGY (JOINT FLUID) \\ \cline{2-2}
& HEMATOLOGY (OTHER BODY FLUID) \\ \cline{2-2}
& HEMATOLOGY (STOOL) \\ \cline{2-2}
& BLOOD GAS (BLOOD) \\ \cline{2-2}
& CHEMISTRY (ASCITES) \\ \cline{2-2}
& CHEMISTRY (CEREBROSPINAL FLUID (CSF)) \\ \cline{2-2}
& CHEMISTRY (OTHER BODY FLUID) \\ \cline{2-2}
& CHEMISTRY (STOOL) \\ \cline{2-2}
& HEMATOLOGY (BLOOD) \\ \cline{2-2}
& HEMATOLOGY (URINE) \\ \cline{2-2}
& BLOOD GAS (OTHER BODY FLUID) \\ \cline{2-2}
& HEMATOLOGY (PLEURAL) \\ \cline{2-2}
& CHEMISTRY (BLOOD) \\ \cline{2-2}
& CHEMISTRY (JOINT FLUID) \\ \cline{2-2}
& CHEMISTRY (PLEURAL) \\ \cline{2-2}
& CHEMISTRY (URINE) \\ \cline{2-2}
& HEMATOLOGY (ASCITES) \\ \cline{2-2}
& HEMATOLOGY (BONE MARROW) \\ \hline
\end{longtable}
}
\subsection{Validation Cache}
\label{app:validation_cache}
\begin{figure*}[t]
    \centering
    \includegraphics[width=\textwidth]{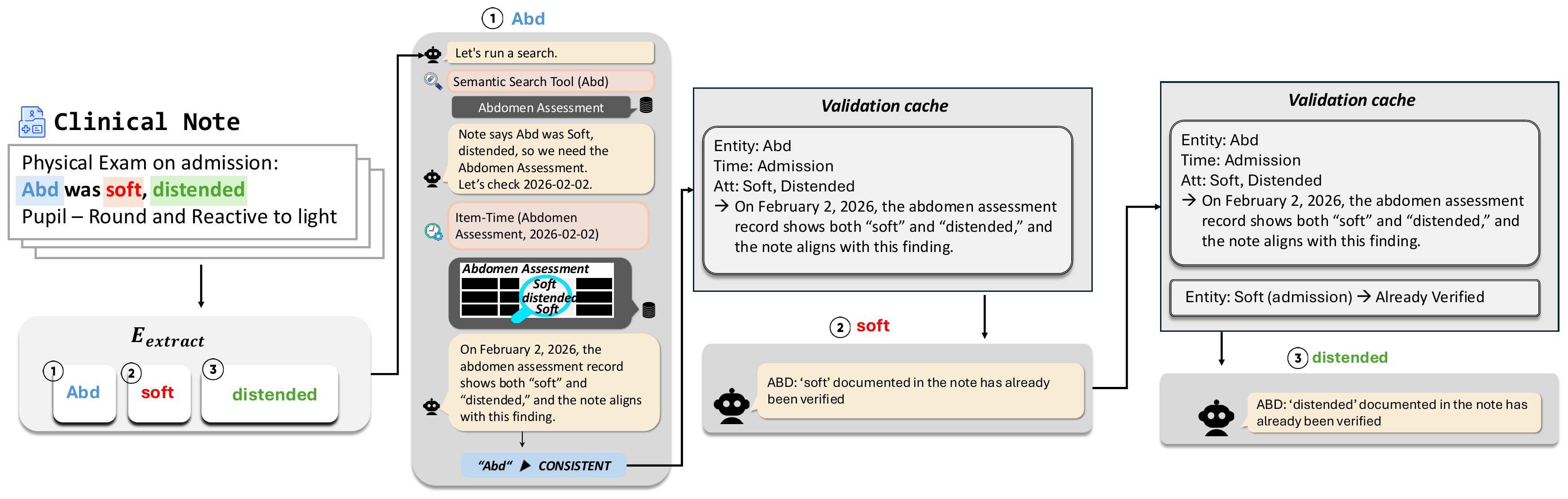}
    \vspace{-2mm}
    \caption{A conceptual illustration of Validation Cache.
}
    \vspace{-2mm}
    \label{fig:cache}
\end{figure*}

In this section, we describe the operational process of the \textsc{validation cache} using the example shown in Figure~\ref{fig:cache}. From the clinical note sentence ``\textit{Abd: soft, distended.}'', the entity extraction step identifies three entities: \textit{Abd}, \textit{soft}, and \textit{distended}. These entities are not independent; rather, they originate from a single abdominal examination. As a result, verifying them independently would lead to repeated retrieval of the same structured data. First, for the entity \textit{Abd}, the LLM invokes a table exploration tool to retrieve the relevant structured record (\eg Abdomen Assessment). The retrieved record indicates that, at the time of admission, the abdomen was documented as “soft” and ``distended.'' This result is stored in the \textsc{validation cache}, which maintains the entity (\textit{Abd}), temporal context (Admission), and associated attributes (Soft, Distended) as reusable evidence for subsequent verification steps. When verifying the entity \textit{soft}, the model first consults the \textsc{validation cache}. Since the cached result already contains the abdominal examination with attributes ``soft'' and ``distended,'' the model can immediately confirm that \textit{soft} has been previously validated. Therefore, the verification is completed without issuing an additional tool call. The same process applies to \textit{distended}, which is also verified through cache lookup alone, without further access to the structured data. In this way, the \textsc{validation cache} enables multiple entities derived from the same clinical observation to share verification results, reducing redundant tool calls and improving computational efficiency. The cache maintains the most recent $m$ validation results in a sliding window, maximizing reuse within temporally localized contexts. We set the validation cache size to five in all experiments.
\subsection{Prompt}
\label{app:prompt}
The prompts used in \frname are presented in Figures~\ref{prompt_note_seg} through~\ref{app:final_veri_prompt}. In addition, to protect patient privacy, all example data values included in the paper have been replaced with fictional ones rather than real data.

\newpage
\begin{figure}[H]
\centering
\input{contents/appendix_/prompt/note_segmentation_prompt}
\caption{Prompt Template for Note Segmentation.}
\label{prompt_note_seg}
\end{figure}

\newpage
\begin{figure}[H]
\centering
\input{contents/appendix_/prompt/time_reference}
\caption{Prompt Template for Extract Time Reference.}
\end{figure}

\newpage
\begin{figure}[H]
\centering
\input{contents/appendix_/prompt/Patient_specific}
\caption{Prompt Template for Patient Specific Entity Extraction.}
\end{figure}

\newpage
\begin{figure}[H]
\centering
\input{contents/appendix_/prompt/map}
\caption{Prompt Template for Mapping Groups and Subgroups.}
\label{map_prmpt_gr_sr}
\end{figure}

\begin{figure}[H]
\centering
\input{contents/appendix_/prompt/ontology_group}
\caption{Prompt Template for Selecting Groups and Subgroups.}
\end{figure}

\newpage
\begin{figure}[H]
\centering
\input{contents/appendix_/prompt/eont}
\caption{Prompt Template for Extracting $E_{\mathrm{ontology}}$.}
\end{figure}

\newpage
\begin{figure}[H]
\centering
\input{contents/appendix_/prompt/scopefilter}
\caption{Prompt Template for Scope Filtering.}
\end{figure}

\newpage
\begin{figure}[H]
\centering
\input{contents/appendix_/prompt/system_tool_calling}
\caption{System Prompt for Tool Calling.}
\end{figure}

\newpage
\begin{figure}[H]
\centering
\input{contents/appendix_/prompt/tool_calling}
\caption{Prompt Template for Tool Calling.}
\end{figure}

\newpage
\begin{figure}[H]
\centering
\input{contents/appendix_/prompt/verification}
\caption{
\label{app:final_veri_prompt}
Prompt Template for Final Verification.}
\end{figure}


\section{LLM-as-a-judge Evaluators}
\label{app:llmjudge}
\subsection{Harsh Evaluator}
\label{app:harsheval}
The \textit{Harsh} evaluator is designed to assess strict agreement with the gold consistency labels. Given a set of predicted entity--value pairs and corresponding gold annotations, the evaluator determines whether the classification is consistent with the gold annotation and assigns a binary label (\textit{Correct} or \textit{Incorrect}).
In this setting, a classification result is considered \textit{correct} only if the assigned consistency label exactly matches the gold label. While the evaluator may internally account for minor variations in entity spans or value expressions when interpreting the classification, the final decision is based strictly on agreement with the gold consistency label.

We implement this evaluator using \textit{Gemini 2.5 Pro}, prompted to act as a deterministic classifier. The model is instructed to output a binary decision reflecting whether the classification is consistent with the gold annotation under a strict interpretation. The full prompt used for this evaluator is provided in Figure~\ref{app:harsh_eval_prompt}.
To validate the reliability of the \textit{Harsh} evaluator, we randomly sample 300 evaluation instances and have them independently reviewed by the authors. The evaluation results are compared against human judgments, achieving 99.46\% agreement. This result indicates that the evaluator reliably reproduces human decisions under well-defined consistency criteria.

\subsection{Lenient Evaluator}
\label{app:lenienteval}
The \textit{Lenient} evaluator is designed to assess the validity of the reasoning underlying the predicted classification, rather than strict agreement with the gold annotation. Unlike the \textit{Harsh} setting, the evaluator focuses on whether the classification is supported by a defensible and clinically appropriate rationale, rather than enforcing surface-level agreement with a single reference annotation.

We employ \textit{Gemini 2.5 Pro} as the evaluator, with prompts emphasizing justification and reasoning validity. The model is instructed to assess whether the predicted classification is supported by a sound reasoning process, rather than strictly matching the gold annotation. The full prompt used for this evaluator is provided in Figure~\ref{app:lenient_eval_prompt}.
To validate this evaluator, we collect a separate set of 200 samples and obtain independent judgments from four clinical practitioners. Agreement is computed individually between the LLM evaluator and each practitioner, and the final score is obtained by averaging across practitioners, resulting in 95.35\% agreement. This demonstrates that the evaluator aligns well with expert judgment in assessing the validity of clinical reasoning, even in cases where multiple interpretations are possible.

\subsection{Evaluator Bias}

To assess potential bias introduced by reliance on a single LLM evaluator, we perform cross-model validation using \textit{GPT-5} as an alternative judge. The same evaluation prompts and decision criteria are applied across both models to ensure consistency. When using \textit{GPT-5} as the evaluator, the measured performance of the framework is comparable to that obtained with \textit{Gemini 2.5 Pro}, as shown in Table~\ref{tab:gpt5_judge}. This indicates that no significant model-specific bias is observed in the evaluation.

\subsection{Evaluation Criteria for Clinical Practitioners}

\subsubsection{Objective}The purpose of this evaluation is to enable clinical experts to conduct a final review of AI-generated assessments verifying consistency between patients’ unstructured clinical notes and structured EHR tables. Reviewers are asked to examine the AI model’s prediction (PREDICT), the reference answer (GOLD), and the AI evaluator’s reasoning comparing the two (RESULT). Based on this review, please determine the final validity of the assessment by marking the Eval field as either Correct or Incorrect.

\subsubsection{Data Structure}
The data provided to reviewers consists of the following fields:
\begin{itemize}
    \item \textbf{PREDICT:} The consistency judgment made by the AI verification model after reading the clinical note and querying the EHR table. This includes the model’s determination of Consistent or Inconsistent, along with its detailed reasoning process.
    \item \textbf{GOLD:} The reference answer, or ground truth, previously established by the research team.
    \item \textbf{RESULT:} The detailed reasoning generated by an AI evaluator, or LLM-as-a-judge, comparing PREDICT with GOLD and providing an initial assessment of whether the prediction is valid.
    \item \textbf{EVAL:} The field to be completed or confirmed by the clinical reviewer. Please determine whether the reasoning in RESULT is clinically appropriate, and mark it as Correct or Incorrect.
\end{itemize}

\subsubsection{Detailed Evaluation Criteria}

\paragraph{Strictness and Clinical Acceptability}~~

\begin{itemize}
    \item \textbf{Evaluation criterion:}~~Even if the text span or specific item name extracted in PREDICT does not exactly match that in GOLD, the prediction should be accepted as correct if it conveys the same meaning in the clinical context and the model’s reasoning is reasonable.

    \item \textbf{Example:}~~If GOLD evaluates “heart rhythm,” but PREDICT logically verifies tachycardia based on the “heart rate” value, this should be marked as Eval: Correct.
\end{itemize}

\paragraph{Acceptance of Omitted Redundant Checks (Validation Cache)}~~

\begin{itemize}
    \item \textbf{Evaluation criterion:}~~For composite indicators such as blood pressure (BP: 94/49), where multiple values are assessed together, the model includes a function that allows it to remember values already checked in previous steps and omit redundant follow-up checks for efficiency, for example: “Consistency check was already completed.”

    \item \textbf{Judgment:} If it is confirmed that the model omitted a redundant check based on a previous verification step, this should be considered normal reasoning rather than an error or omission. In such cases, mark Eval: Correct.

\end{itemize}

\paragraph{Rejection of Clear Medical Errors and Misjudgments}~~
\begin{itemize}
    \item \textbf{Evaluation criterion:}~~The following cases should be marked as incorrect even if the AI evaluator (RESULT) judged them to be correct:
    \begin{itemize}
    \item The model (PREDICT) claims consistency by fabricating EHR records that do not exist, i.e., hallucination.
    \item The model relies on an incorrect medical standard, such as judging an abnormal value as normal.
    \item The comparison logic in RESULT itself contradicts medical facts.
    \item The AI evaluator claims consistency based on an incorrect EHR record.
\end{itemize}
\end{itemize}
\begin{itemize}
    \item \textbf{Judgment:}~~If any such critical error is identified, mark Eval: Incorrect.
\end{itemize}

\begin{figure}[H]
\centering
\input{contents/appendix_/prompt/harsh_eval}
\caption{
\label{app:harsh_eval_prompt}
Prompt Template for Harsh Evaluator.}
\end{figure}

\begin{table*}[!t]
\centering
\caption{Comparison of evaluation results between \textit{Gemini 2.5 Pro} and \textit{GPT-5} under identical settings, showing consistent performance across Lenient and Harsh evaluators.}
\label{tab:gpt5_judge}
\resizebox{\linewidth}{!}{
\begin{tabular}{c|cccccc|cccccc}
\toprule
 & \multicolumn{6}{c}{Lenient} & \multicolumn{6}{c}{Harsh} \\
\cmidrule(lr){2-7} \cmidrule(lr){8-13}
LLM 
& \multicolumn{2}{c}{Discharge} & \multicolumn{2}{c}{Physician} & \multicolumn{2}{c}{Nursing}
& \multicolumn{2}{c}{Discharge} & \multicolumn{2}{c}{Physician} & \multicolumn{2}{c}{Nursing} \\
\cmidrule(lr){2-3} \cmidrule(lr){4-5} \cmidrule(lr){6-7}
\cmidrule(lr){8-9} \cmidrule(lr){10-11} \cmidrule(lr){12-13}
& R & P & R & P & R & P 
& R & P & R & P & R & P \\
\midrule
Gemini 2.5 Pro 
& 75.59 & 66.99 & 82.06 & 78.33 & 75.60 & 73.98 
& 65.56 & 56.18 & 74.06 & 70.21 & 65.43 & 60.76 \\
GPT-5 
& 75.46 & 67.65 & 82.08 & 77.09 & 74.31 & 73.75 
& 65.00 & 55.26 & 74.27 & 68.89 & 63.84 & 59.45 \\
\bottomrule
\end{tabular}
}
\end{table*}

\begin{figure}[H]
\centering
\input{contents/appendix_/prompt/linprompt}
\caption{
\label{app:lenient_eval_prompt}
Prompt Template for Lenient Evaluator.}
\end{figure}

\section{Statistical Evaluation}
\label{app:statistical_eval}
To statistically evaluate the framework, we conducted three runs on the validation set of \dtname using \frname. In addition, to assess the statistical reliability of the LLM-based evaluator, each of the three runs was independently evaluated three times using the LLM-as-a-judge approach. As shown in Table~\ref{tab:detailed_eval}, the results are reported in terms of mean and standard deviation.

\begin{table*}[!t]
\centering
\caption{Detailed evaluation results across runs with per-run averages and standard deviations under Lenient and Harsh settings.}
\label{tab:detailed_eval}
\resizebox{\linewidth}{!}{
\begin{tabular}{cc|cccccc|cccccc}
\toprule
 &  & \multicolumn{6}{c}{Lenient} & \multicolumn{6}{c}{Harsh} \\
\cmidrule(lr){3-8} \cmidrule(lr){9-14}
Run & Eval 
& \multicolumn{2}{c}{Discharge} & \multicolumn{2}{c}{Physician} & \multicolumn{2}{c}{Nursing}
& \multicolumn{2}{c}{Discharge} & \multicolumn{2}{c}{Physician} & \multicolumn{2}{c}{Nursing} \\
\cmidrule(lr){3-4} \cmidrule(lr){5-6} \cmidrule(lr){7-8}
\cmidrule(lr){9-10} \cmidrule(lr){11-12} \cmidrule(lr){13-14}
 & 
& R & P & R & P & R & P 
& R & P & R & P & R & P \\
\midrule
\multirow{5}{*}{1}
& 1 & 75.59 & 66.99 & 82.06 & 78.33 & 75.60 & 73.98 & 65.56 & 56.18 & 74.06 & 70.21 & 65.43 & 60.76 \\
& 2 & 74.83 & 67.80 & 80.78 & 77.42 & 73.66 & 71.45 & 65.36 & 56.63 & 72.82 & 69.40 & 64.78 & 60.92 \\
& 3 & 75.21 & 67.37 & 80.62 & 78.14 & 77.91 & 73.72 & 64.19 & 56.01 & 73.41 & 70.40 & 65.14 & 60.91 \\
\cmidrule(lr){2-14}
& \textbf{Avg} 
& \textbf{75.21} & \textbf{67.39} 
& \textbf{81.15} & \textbf{77.97} 
& \textbf{75.73} & \textbf{73.05} 
& \textbf{65.04} & \textbf{56.27} 
& \textbf{73.43} & \textbf{70.00} 
& \textbf{65.11} & \textbf{60.86} \\
& \textit{Std} 
& 0.38 & 0.40 & 0.79 & 0.48 & 2.13 & 1.39 
& 0.74 & 0.32 & 0.62 & 0.53 & 0.33 & 0.09 \\
\midrule
\multirow{5}{*}{2}
& 1 & 71.62 & 68.42 & 79.72 & 77.74 & 74.39 & 66.32 & 62.73 & 55.27 & 72.24 & 71.53 & 63.04 & 55.31 \\
& 2 & 72.70 & 69.70 & 79.65 & 76.99 & 75.89 & 66.41 & 62.05 & 55.48 & 73.27 & 71.00 & 65.89 & 54.23 \\
& 3 & 73.18 & 68.50 & 80.17 & 77.56 & 75.33 & 68.59 & 63.27 & 55.67 & 72.98 & 70.84 & 64.68 & 56.80 \\
\cmidrule(lr){2-14}
& \textbf{Avg} 
& \textbf{72.50} & \textbf{68.87} 
& \textbf{79.85} & \textbf{77.43} 
& \textbf{75.20} & \textbf{67.11} 
& \textbf{62.68} & \textbf{55.47} 
& \textbf{72.83} & \textbf{71.12} 
& \textbf{64.54} & \textbf{55.45} \\
& \textit{Std} 
& 0.80 & 0.69 & 0.18 & 0.39 & 0.88 & 1.29 
& 0.40 & 0.12 & 0.53 & 0.29 & 1.44 & 1.29 \\
\midrule
\multirow{5}{*}{3}
& 1 & 73.74 & 67.66 & 79.32 & 76.57 & 72.57 & 70.22 & 62.86 & 50.27 & 71.64 & 68.51 & 63.63 & 59.81 \\
& 2 & 74.34 & 67.60 & 79.94 & 76.15 & 72.34 & 69.06 & 62.66 & 55.80 & 71.81 & 67.69 & 62.91 & 59.28 \\
& 3 & 72.22 & 67.35 & 79.34 & 76.53 & 72.23 & 72.38 & 62.37 & 56.07 & 70.79 & 69.46 & 63.89 & 60.15 \\
\cmidrule(lr){2-14}
& \textbf{Avg} 
& \textbf{73.43} & \textbf{67.54} 
& \textbf{79.53} & \textbf{76.42} 
& \textbf{72.38} & \textbf{70.55} 
& \textbf{62.63} & \textbf{54.05} 
& \textbf{71.41} & \textbf{68.55} 
& \textbf{63.47} & \textbf{59.75} \\
& \textit{Std} 
& 1.09 & 0.16 & 0.35 & 0.23 & 0.17 & 1.68 
& 0.25 & 3.27 & 0.54 & 0.89 & 0.51 & 0.44 \\
\bottomrule
\end{tabular}
}
\end{table*}

\subsection{Paired Bootstrap Analysis on the Test Set}
\label{app:bootstrap}

To quantify uncertainty in the performance differences between
\textsc{EHR-Inspector} and CheckEHR, we conduct a paired bootstrap
analysis over the 83 held-out test notes. For each combination of
base LLM, note type, and evaluation criterion, we resample the test
notes with replacement 10,000 times. For each bootstrap replicate,
we compute the paired difference in per-note F1,
$\Delta F_1 =
F_1(\textsc{EHR-Inspector}) - F_1(\textsc{CheckEHR})$.
Both methods are evaluated on the same resampled notes, preserving
the paired structure of the comparison and reducing variability due
to differences in note difficulty. The resulting estimates are
summarized in Table~\ref{tab:bootstrap_test}.

\begin{table*}[t]
\centering
\caption{
Paired bootstrap comparison with CheckEHR on the test set.
Each cell reports the mean paired F1 difference
($\Delta F_1$), followed by the percentile-based 95\% confidence
interval and the corresponding paired-bootstrap $p$-value.
Positive values indicate higher F1 for \textsc{EHR-Inspector}.
}
\label{tab:bootstrap_test}

\small
\setlength{\tabcolsep}{4pt}
\renewcommand{\arraystretch}{1.12}

\begin{tabular}{@{}lcccccc@{}}
\toprule
& \multicolumn{3}{c}{\textbf{Lenient}}
& \multicolumn{3}{c}{\textbf{Harsh}} \\
\cmidrule(lr){2-4}
\cmidrule(lr){5-7}
\textbf{Base LLM}
& \textbf{Discharge}
& \textbf{Physician}
& \textbf{Nursing}
& \textbf{Discharge}
& \textbf{Physician}
& \textbf{Nursing} \\
\midrule

Gemini 2.5 Flash
& \shortstack{$+18.61$ \\ {\scriptsize $[14.47,\,22.77]$} \\ {\scriptsize $p<.001$}}
& \shortstack{$+44.03$ \\ {\scriptsize $[41.26,\,47.18]$} \\ {\scriptsize $p<.001$}}
& \shortstack{$+24.18$ \\ {\scriptsize $[18.11,\,30.65]$} \\ {\scriptsize $p<.001$}}
& \shortstack{$+16.45$ \\ {\scriptsize $[12.77,\,19.77]$} \\ {\scriptsize $p<.001$}}
& \shortstack{$+41.11$ \\ {\scriptsize $[37.68,\,44.27]$} \\ {\scriptsize $p<.001$}}
& \shortstack{$+20.82$ \\ {\scriptsize $[15.25,\,26.17]$} \\ {\scriptsize $p<.001$}}
\\[1.5mm]

Qwen3-32B
& \shortstack{$+23.61$ \\ {\scriptsize $[16.14,\,31.82]$} \\ {\scriptsize $p<.001$}}
& \shortstack{$+24.70$ \\ {\scriptsize $[20.39,\,29.12]$} \\ {\scriptsize $p<.001$}}
& \shortstack{$+27.58$ \\ {\scriptsize $[19.93,\,35.24]$} \\ {\scriptsize $p<.001$}}
& \shortstack{$+21.08$ \\ {\scriptsize $[13.68,\,27.70]$} \\ {\scriptsize $p<.001$}}
& \shortstack{$+21.53$ \\ {\scriptsize $[17.68,\,25.58]$} \\ {\scriptsize $p<.001$}}
& \shortstack{$+21.07$ \\ {\scriptsize $[14.78,\,28.36]$} \\ {\scriptsize $p<.001$}}
\\[1.5mm]

MedGemma-27B
& \shortstack{$+19.68$ \\ {\scriptsize $[12.39,\,27.26]$} \\ {\scriptsize $p<.001$}}
& \shortstack{$+19.73$ \\ {\scriptsize $[14.48,\,25.86]$} \\ {\scriptsize $p<.001$}}
& \shortstack{$+39.35$ \\ {\scriptsize $[32.29,\,45.16]$} \\ {\scriptsize $p<.001$}}
& \shortstack{$+17.27$ \\ {\scriptsize $[9.33,\,25.54]$} \\ {\scriptsize $p<.001$}}
& \shortstack{$+18.34$ \\ {\scriptsize $[12.83,\,23.87]$} \\ {\scriptsize $p<.001$}}
& \shortstack{$+35.34$ \\ {\scriptsize $[29.30,\,41.09]$} \\ {\scriptsize $p<.001$}}
\\[1.5mm]

GPT-OSS-20B
& \shortstack{$+0.29$ \\ {\scriptsize $[-6.60,\,7.09]$} \\ {\scriptsize $p=.930$}}
& \shortstack{$+6.22$ \\ {\scriptsize $[-0.33,\,13.23]$} \\ {\scriptsize $p=.084$}}
& \shortstack{$+5.25$ \\ {\scriptsize $[-2.06,\,12.78]$} \\ {\scriptsize $p=.182$}}
& \shortstack{$+0.81$ \\ {\scriptsize $[-5.31,\,6.53]$} \\ {\scriptsize $p=.776$}}
& \shortstack{$+2.74$ \\ {\scriptsize $[-3.02,\,8.57]$} \\ {\scriptsize $p=.386$}}
& \shortstack{$+3.10$ \\ {\scriptsize $[-4.65,\,10.51]$} \\ {\scriptsize $p=.444$}}
\\

\bottomrule
\end{tabular}
\end{table*}

\section{Error Analysis by Model}
\label{app:error_analysis}
To better understand model-specific failure patterns in \frname, we randomly sampled 100 error cases per model and conducted a manual analysis. Each error was categorized into five types: Premature Conclusion Error, Tool Usage Error, Temporal Reasoning Error, Medical Knowledge Error, and Validation Cache Error.

\paragraph{Premature Conclusion Error}
As shown in Figure~\ref{fig:errortype}, this is the most frequent error across all models. It occurs when a model terminates the verification process before collecting sufficient supporting evidence. In practice, models often stop after limited exploration (\eg semantic search or lexical search) without further table exploration. This behavior indicates a systematic under-search issue and highlights the need for mechanisms that encourage deeper exploration or enforce sufficient evidence gathering.

\paragraph{Tool Usage Error}
This error arises when models fail to use available tools consistently. For example, a model may identify ``Heart Rate'' in a table but subsequently issue a query using an inconsistent term such as ``HR,'' leading to failed retrieval. These errors reflect limitations in query formulation and tool interaction, and are particularly prominent in Gemini-2.5-Flash and Qwen3.

\paragraph{Temporal Reasoning Error}
This error refers to failures in correctly interpreting time-related information in clinical notes. Such notes often contain events that are not chronologically ordered and are densely interwoven, making it difficult to determine when a specific condition or event occurred. Our analysis shows that models still struggle with such temporally complex and unstructured data.

Notably, GPT-OSS exhibits relatively fewer Temporal Reasoning Errors, as well as fewer Tool Usage Errors. However, this is primarily because it shows a significantly higher rate of Premature Conclusion Errors, often terminating before reaching stages where such errors would occur. This suggests that lower downstream error rates do not necessarily indicate stronger reasoning ability, but rather earlier failure in the reasoning pipeline.

\paragraph{Medical Knowledge Error}
This error occurs when models rely on incorrect or insufficient internal medical knowledge. While MedGemma, a medically specialized model, shows fewer such errors compared to general-purpose models, these errors are not entirely eliminated. This indicates that reliance on parametric knowledge alone is insufficient and that incorporating external knowledge retrieval could further improve performance.

\paragraph{Validation Cache Error}
This error involves failures in tracking or reusing intermediate verification states. It is relatively rare overall and almost negligible in stronger models, suggesting that they maintain more reliable intermediate representations during multi-step reasoning.

\section{NER Baselines Implementation}
\label{app:ner_baselines}
To analyze the impact of the entity extraction stage, we compare our method with three NER baselines while keeping the rest of the framework unchanged: Trained NER, BERT Ensemble, and CheckEHR NER. 

\textbf{Trained NER} fine-tunes MedGemma 27B on note–entity pairs from \dtname. After note segmentation in \frname, each note is divided into segments of varying lengths, and the model is trained to extract entities from these segments. To construct the training data, each note is split into chunks containing $n$ lines, where each chunk is paired with the entities corresponding to those lines. We vary $n$ from 2 to the full length of the note, allowing the model to learn from inputs of different granularities. This setup encourages the model to generalize across varying input lengths and to reliably extract the corresponding entity sets.
Fine-tuning is performed for one epoch with a learning rate of $2\times10^{-5}$, an effective batch size of 8 through gradient accumulation, cosine learning-rate scheduling with a warmup ratio of 0.03, and a maximum sequence length of 2048 tokens.

\textbf{BERT Ensemble} aggregates predictions from three biomedical BERT-based models trained on different datasets~\cite{beltagy2019scibert, mattupalli2023clinicalner, uzuner20112010i2b2}. In our implementation, we use BERT-based NER models provided by Hugging Face\footnote{\url{https://huggingface.co}}, spaCy\footnote{\url{https://spacy.io}}, and Stanza\footnote{\url{https://stanfordnlp.github.io/stanza/}}.

\textbf{CheckEHR NER} follows the entity extraction method used in the original CheckEHR pipeline.

\section{Tool Ablation Experiment}
\label{app:tool_ablation}
Detailed ablation results for the tool category are provided in Table~\ref{tab:tool_abl_detail}.

\begin{table*}[!t]
\centering
\caption{Detailed ablation results for tool categories.}
\label{tab:tool_abl_detail}
\resizebox{\linewidth}{!}{
\begin{tabular}{l|cccccc|cccccc}
\toprule
 & \multicolumn{6}{c}{Lenient} & \multicolumn{6}{c}{Harsh} \\
\cmidrule(lr){2-7} \cmidrule(lr){8-13}
Method
& \multicolumn{2}{c}{Discharge} & \multicolumn{2}{c}{Physician} & \multicolumn{2}{c}{Nursing}
& \multicolumn{2}{c}{Discharge} & \multicolumn{2}{c}{Physician} & \multicolumn{2}{c}{Nursing} \\
\cmidrule(lr){2-3} \cmidrule(lr){4-5} \cmidrule(lr){6-7}
\cmidrule(lr){8-9} \cmidrule(lr){10-11} \cmidrule(lr){12-13}
& R & P & R & P & R & P 
& R & P & R & P & R & P \\
\midrule

\frname
& 75.59 & 66.99 & 82.06 & 78.33 & 75.60 & 73.98
& 65.56 & 56.18 & 74.06 & 70.21 & 65.43 & 60.76 \\

- Entity-to-Table Item Alignment
& 77.00 & 69.09 & 84.32 & 80.97 & 77.00 & 71.90
& 67.29 & 57.54 & 77.12 & 73.49 & 64.57 & 59.93 \\

- Database Exploration and Value Profiling
& 77.31 & 67.25 & 80.65 & 77.30 & 74.77 & 72.62
& 65.59 & 56.75 & 71.41 & 68.21 & 64.11 & 62.16 \\

- Temporal and Conditional Record Retrieval
& 65.79 & 60.92 & 80.64 & 76.44 & 74.56 & 75.21
& 57.05 & 51.03 & 72.16 & 67.77 & 62.39 & 60.92 \\

\bottomrule
\end{tabular}
}
\end{table*}

\section{Tool Trace Analysis by Model}
\label{app:tool_trace_model}
\begin{figure*}[t]
    \centering
    \includegraphics[width=\textwidth]{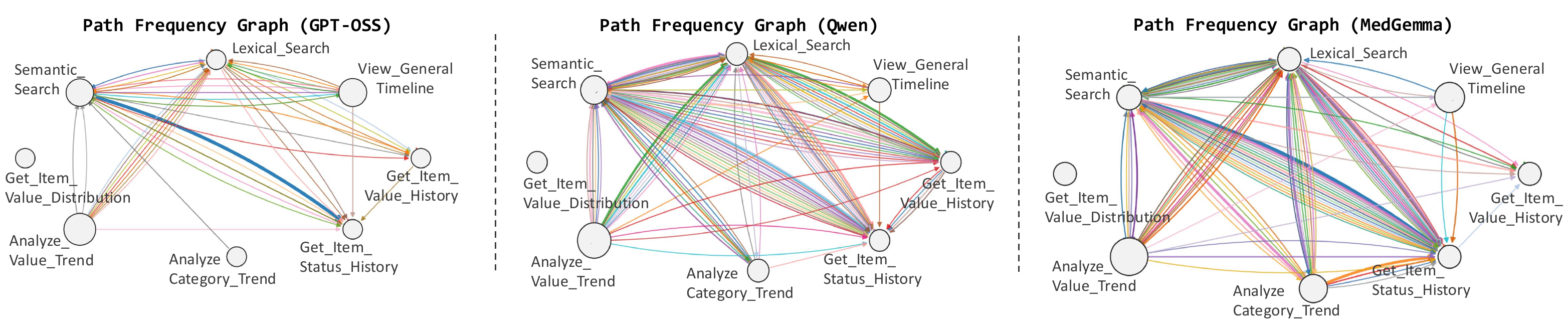}
    \vspace{-2mm}
    \caption{Comparison of path frequency graphs across models.
}
    \vspace{-2mm}
    \label{fig:compare22}
\end{figure*}

Figure~\ref{fig:compare22} compares the tool-usage traces generated by three backbone models, GPT-OSS 20B~\cite{agarwal2025gpt}, Qwen3-32B~\cite{yang2025qwen3}, and MedGemma 27B~\cite{sellergren2025medgemma}, during consistency verification. Overall, all three models exhibit denser and more dispersed transition patterns than the human annotators shown in Figure~\ref{fig:frameworkvshuman}. This suggests that LLM-based agents tend to explore a broader range of tool combinations across search, history retrieval, and trend analysis, rather than converging on a small set of stable tool-usage paths. However, the degree of dispersion differs across models. GPT-OSS shows the most concentrated structure among the three models, with prominent paths centered around Semantic\_Search, Lexical\_Search, Get\_Item\_Status\_History, and Get\_Item\_Value\_History. This indicates that GPT-OSS tends to follow retrieval-to-history verification patterns more consistently than the other models. In contrast, Qwen exhibits the densest and most widely dispersed transition structure, with frequent transitions across nearly all major tools. This suggests that Qwen is more likely to explore a wide range of possible tool combinations rather than converge on a specific verification path. MedGemma also shows a high-density transition structure despite being specialized for the medical domain. In particular, repeated verification patterns are observed around Lexical\_Search, Analyze\_Category\_Trend, and Get\_Item\_Status\_History, where retrieval is followed by trend analysis or history inspection. This indicates that domain-specific knowledge does not necessarily translate into efficient tool-use planning in relational-table-based verification settings. Taken together, although the models differ in their tool-use behaviors, all of them generate broader and less compact tool-usage traces than human annotators.

\section{Prior Knowledge of the MIMIC Schema}
\label{app:prior_mimiciii}
To analyze the extent of Gemini’s internal knowledge of the MIMIC schema, we present the prompt and its corresponding response.

\subsection{Gemini Prompt}

The following content is the response generated by the Gemini model when prompted with: \textit{``Tell me what you know about the EHR database MIMIC without using web search. Describe its table structure and the columns it contains.''}

\subsection{Gemini Response} 
The \textbf{Medical Information Mart for Intensive Care (MIMIC)} is one of the most significant open-access datasets in the world of healthcare informatics. It is maintained by the MIT Laboratory for Computational Physiology and contains de-identified health-related data from patients admitted to critical care units at the Beth Israel Deaconess Medical Center.

While there are several versions (MIMIC-II, MIMIC-III, and the current MIMIC-IV), the structure generally follows a relational database model designed to capture the complex timeline of an ICU stay.

\noindent \textbf{Database Architecture} \\
The database is organized into several modules. The primary structure is built around a few ``core'' tables that define patients and their hospital stays, which then link to various ``event'' tables containing clinical data.

\noindent \textbf{1. Core Tables (The Foundation)} \\
These tables define the ``who'' and ``when'' of the data.
\begin{itemize}
    \item \textbf{PATIENTS:} Contains one row per patient.
    \begin{itemize}
        \item \textit{Columns:} \texttt{subject\_id}, \texttt{gender}, \texttt{dob}, \texttt{dod} (date of death).
    \end{itemize}
    \item \textbf{ADMISSIONS:} Tracks each unique hospital visit.
    \begin{itemize}
        \item \textit{Columns:} \texttt{subject\_id}, \texttt{hadm\_id}, \texttt{admittime}, \texttt{dischtime}, \texttt{admission\_type}, \texttt{insurance}, \texttt{religion}, \texttt{marital\_status}, \texttt{ethnicity}.
    \end{itemize}
    \item \textbf{ICUSTAYS:} Defines stays within an ICU (since one hospital admission can involve multiple ICU transfers).
    \begin{itemize}
        \item \textit{Columns:} \texttt{subject\_id}, \texttt{hadm\_id}, \texttt{icustay\_id}, \texttt{first\_careunit}, \texttt{last\_careunit}, \texttt{intime}, \texttt{outime}, \texttt{los} (length of stay).
    \end{itemize}
\end{itemize}

\noindent \textbf{2. Event Tables (The Clinical Data)} \\
These are the largest tables, containing longitudinal data points timestamped to the patient's stay.
\begin{itemize}
    \item \textbf{CHARTEVENTS:} The largest table. Contains all charted data for a patient (vitals, sensor data, etc.).
    \begin{itemize}
        \item \textit{Columns:} \texttt{subject\_id}, \texttt{hadm\_id}, \texttt{icustay\_id}, \texttt{itemid}, \texttt{charttime}, \texttt{valuenum}, \texttt{valueuom} (unit of measurement).
    \end{itemize}
    \item \textbf{LABEVENTS:} Contains results from laboratory tests (blood work, chemistry).
    \begin{itemize}
        \item \textit{Columns:} \texttt{subject\_id}, \texttt{hadm\_id}, \texttt{itemid}, \texttt{charttime}, \texttt{value}, \texttt{valuenum}, \texttt{flag} (marks ``abnormal'' results).
    \end{itemize}
    \item \textbf{PRESCRIPTIONS:} Medications ordered for the patient.
    \begin{itemize}
        \item \textit{Columns:} \texttt{subject\_id}, \texttt{hadm\_id}, \texttt{startdate}, \texttt{enddate}, \texttt{drug}, \texttt{dose\_val\_rx}, \texttt{dose\_unit\_rx}, \texttt{route}.
    \end{itemize}
    \item \textbf{INPUTEVENTS / OUTPUTEVENTS:} Tracks fluids coming in (IVs, nutrition) and fluids going out (urine, drains).
\end{itemize}

\noindent \textbf{3. Dictionary Tables} \\
To keep the database normalized, specific codes are mapped to human-readable labels in ``D'' tables.
\begin{itemize}
    \item \textbf{D\_ICD\_DIAGNOSES:} Maps ICD codes to diagnosis descriptions.
    \item \textbf{D\_ITEMS:} Maps the \texttt{itemid} found in \texttt{CHARTEVENTS} to what it actually represents (e.g., Heart Rate, GCS, Temperature).
    \item \textbf{D\_LABITEMS:} Maps the \texttt{itemid} in \texttt{LABEVENTS} to the specific lab test name.
\end{itemize}

\noindent \textbf{Key Identifiers} \\
To join these tables effectively, you rely on a hierarchy of IDs:
\begin{enumerate}
    \item \textbf{\texttt{subject\_id}}: Unique to a person.
    \item \textbf{\texttt{hadm\_id}}: Unique to a single hospital admission.
    \item \textbf{\texttt{icustay\_id}}: Unique to a single unit stay within that admission.
\end{enumerate}

\noindent \textbf{Important Note on Privacy} \\
Though the data is incredibly detailed, it is \textbf{de-identified} in compliance with HIPAA. Dates are shifted into the future by a random offset for each patient (though the intervals between events remain chronologically accurate), and patients over the age of 89 have their ages obscured to prevent identification of the ``oldest old.''

\section{Generalization of \frname}
\label{app:ehrcon_res}
To evaluate whether \frname generalizes to different datasets and database schemas, we conduct additional experiments under three settings.

\subsection{Evaluation on EHRCon}
To assess whether \frname generalizes to other datasets, we evaluate it on EHRCon, a \textit{surface-level} consistency verification dataset. The results show that \frname achieves a Lenient F1 score of 86.41, outperforming CheckEHR by 24.99 points. This indicates that \frname generalizes well even when applied to datasets with different task formulations. Detailed results are reported in Table~\ref{tab:app_ehrconv1}.

\subsection{Evaluation on MIMIC-IV}
To evaluate robustness under different note formats and database schemas, we conduct experiments on MIMIC-IV. To this end, we construct a MIMIC-IV validation set under a matched annotation scale, manually annotating seven discharge summaries following the \dtname annotation protocol.
Compared to MIMIC-III, this setting introduces additional complexity, including new tables (\eg \textit{emar} and \textit{emar\_detail}). Under this setting, performance shows a modest decrease. The Lenient F1 score drops from 71.03 to 68.67, driven by changes in Recall (75.59 $\rightarrow$ 68.18) and Precision (66.99 $\rightarrow$ 69.18). The Harsh F1 score also decreases, driven by changes in Recall (65.56 $\rightarrow$ 58.27) and Precision (56.18 $\rightarrow$ 56.72). The MIMIC-IV tables and columns we used are listed in Table~\ref{apptab:table_columns_mimiciv}.

\begin{table}[H]
\centering
\renewcommand{\arraystretch}{1.2} 
\caption{Tables and columns of MIMIC-IV.}
\resizebox{0.5\columnwidth}{!}{%
\begin{tabular}{l|l}
\hline
\textbf{TABLE} & \textbf{COLUMN} \\ \hline
\multirow{3}{*}{D\_ITEMS} & ITEMID \\ \cline{2-2} 
 & CATEGORY \\ \cline{2-2} 
 & LABEL \\ \hline
DIAGNOSES\_ICD & ICD\_CODE \\ \hline
D\_ICD\_DIAGNOSES & LONG\_TITLE \\ \hline
D\_ICD\_PROCEDURES & LONG\_TITLE \\ \hline
\multirow{3}{*}{D\_LABITEMS} & FLUID \\ \cline{2-2} 
 & ITEMID \\ \cline{2-2} 
 & LABEL \\ \hline
\multirow{9}{*}{INPUTEVENTS} & ORDERCATEGORYNAME \\ \cline{2-2} 
 & RATE \\ \cline{2-2} 
 & AMOUNT \\ \cline{2-2} 
 & STARTTIME \\ \cline{2-2} 
 & ENDTIME \\ \cline{2-2} 
 & RATEUOM \\ \cline{2-2} 
 & ORIGINAL\_AMOUNT \\ \cline{2-2} 
 & ORIGINAL\_RATE \\ \cline{2-2} 
 & AMOUNTUOM \\ \hline
\multirow{4}{*}{CHARTEVENTS} & CHARTTIME \\ \cline{2-2} 
 & VALUE \\ \cline{2-2} 
 & VALUENUM \\ \cline{2-2} 
 & VALUEUOM \\ \hline
\multirow{7}{*}{MICROBIOLOGYEVENTS} & CHARTTIME \\ \cline{2-2} 
 & SPEC\_ITEMID \\ \cline{2-2} 
 & SPEC\_TYPE\_DESC \\ \cline{2-2} 
 & ORG\_ITEMID \\ \cline{2-2} 
 & ORG\_NAME \\ \cline{2-2} 
 & INTERPRETATION \\ \cline{2-2} 
 & AB\_NAME \\ \hline
\multirow{3}{*}{EMAR} & CHARTTIME \\ \cline{2-2} 
 & MEDICATION \\ \cline{2-2} 
 & EMAR\_ID \\  \hline
 \multirow{6}{*}{EMAR\_DETAIL} & DOSE\_DUE \\ \cline{2-2} 
& DOSE\_DUE\_UNIT \\ \cline{2-2} 
& DOSE\_GIVEN \\ \cline{2-2} 
& DOSE\_GIVEN\_UNIT \\ \cline{2-2} 
& ROUTE \\ \cline{2-2} 
& EMAR\_ID \\  \hline
\multirow{3}{*}{OUTPUTEVENTS} & CHARTTIME \\ \cline{2-2} 
 & VALUE \\ \cline{2-2} 
 & VALUEUOM \\ \hline
\multirow{5}{*}{LABEVENTS} & CHARTTIME \\ \cline{2-2} 
 & VALUE \\ \cline{2-2} 
 & VALUENUM \\ \cline{2-2} 
 & FLAG \\ \cline{2-2} 
 & VALUEUOM \\ \hline
PROCEDURES\_ICD & ICD\_CODE \\ \hline
\multirow{9}{*}{PRESCRIPTIONS} & STARTTIME \\ \cline{2-2} 
 & STOPTIME \\ \cline{2-2} 
 & DRUG \\ \cline{2-2} 
 & PROD\_STRENGTH \\ \cline{2-2} 
 & DOSE\_VAL\_RX \\ \cline{2-2} 
 & DOSE\_UNIT\_RX \\ \cline{2-2} 
 & FORM\_VAL\_DISP \\ \cline{2-2} 
 & FORM\_UNIT\_DISP \\ \cline{2-2} 
 & ROUTE \\ \hline
\multirow{3}{*}{PROCEDUREEVENTS} & STARTTIME \\ \cline{2-2} 
 & ENDTIME \\ \cline{2-2} 
 & LOCATION \\ \hline
\end{tabular}%
}
\vspace{3mm}
\label{apptab:table_columns_mimiciv}
\end{table}

\subsection{Evaluation on Perturbed MIMIC-III}
To examine whether \frname relies excessively on the base LLM’s prior knowledge of the MIMIC schema, we construct a perturbed database by replacing table and column names with new identifiers. This enables evaluation under an unseen schema. As a result, the Lenient F1 score decreases from 74.44 to 69.73, suggesting that familiarity with the original schema provides some benefit. Nevertheless, \frname still achieves a reasonably high F1 score, indicating robustness to schema variations. Detailed results are reported in Table~\ref{tab:app_perturbation}, and the perturbed table and column names are listed in Table~\ref{apptab:table_columns_purt}.

\begin{table*}[!t]
\centering
\caption{Comparison of \frname\ and CheckEHR on the EHRCon benchmark for note–table consistency verification, evaluated across note types using Lenient and Harsh metrics.}
\label{tab:app_ehrconv1}
\resizebox{\linewidth}{!}{
\begin{tabular}{c|cccccc|cccccc}
\toprule
 & \multicolumn{6}{c}{Lenient} & \multicolumn{6}{c}{Harsh} \\
\cmidrule(lr){2-7} \cmidrule(lr){8-13}
Method 
& \multicolumn{2}{c}{Discharge} & \multicolumn{2}{c}{Physician} & \multicolumn{2}{c}{Nursing}
& \multicolumn{2}{c}{Discharge} & \multicolumn{2}{c}{Physician} & \multicolumn{2}{c}{Nursing} \\
\cmidrule(lr){2-3} \cmidrule(lr){4-5} \cmidrule(lr){6-7}
\cmidrule(lr){8-9} \cmidrule(lr){10-11} \cmidrule(lr){12-13}
& R & P & R & P & R & P 
& R & P & R & P & R & P \\
\midrule
\frname
& 83.85 & 91.72 & 86.02 & 93.09 & 77.42 & 89.85 
& 75.36 & 78.93 & 79.66 & 84.93 & 67.39 & 76.38 \\
CheckEHR 
& 69.79 & 61.65 & 60.65 & 47.29 & 73.64 & 54.86
& 68.72 & 55.74 & 59.22 & 43.94 & 64.87 & 48.40 \\
\bottomrule
\end{tabular}
}
\end{table*}

\begin{table*}[!t]
\centering
\caption{Evaluation results of \frname across two database settings (original vs. perturbed) under identical experimental conditions, reported for both Lenient and Harsh evaluators.}
\label{tab:app_perturbation}
\resizebox{\linewidth}{!}{
\begin{tabular}{c|cccccc|cccccc}
\toprule
 & \multicolumn{6}{c}{Lenient} & \multicolumn{6}{c}{Harsh} \\
\cmidrule(lr){2-7} \cmidrule(lr){8-13}
Database 
& \multicolumn{2}{c}{Discharge} & \multicolumn{2}{c}{Physician} & \multicolumn{2}{c}{Nursing}
& \multicolumn{2}{c}{Discharge} & \multicolumn{2}{c}{Physician} & \multicolumn{2}{c}{Nursing} \\
\cmidrule(lr){2-3} \cmidrule(lr){4-5} \cmidrule(lr){6-7}
\cmidrule(lr){8-9} \cmidrule(lr){10-11} \cmidrule(lr){12-13}
& R & P & R & P & R & P 
& R & P & R & P & R & P \\
\midrule
Original 
& 75.59 & 66.99 & 82.06 & 78.33 & 75.60 & 73.98 
& 65.56 & 56.18 & 74.06 & 70.21 & 65.43 & 60.76 \\
Perturbed 
& 63.90 & 68.62 & 73.06 & 77.60 & 66.51 & 72.10
& 53.82 & 55.69 & 64.19 & 68.48 & 58.56 & 62.15 \\
\bottomrule
\end{tabular}
}
\end{table*}





\begin{center}
\footnotesize
\renewcommand{\arraystretch}{1.3}
\begin{longtable}{p{0.225\linewidth}|p{0.215\linewidth}|p{0.23\linewidth}|p{0.21\linewidth}}
\caption{\normalsize Tables and columns used with perturbation mappings.}
\label{apptab:table_columns_purt} \\        
\hline
\textbf{TABLE} & \textbf{COLUMN} & \textbf{PERTURBED\_TABLE} & \textbf{PERTURBED\_COLUMN} \\ \hline
\endfirsthead
\hline
\textbf{TABLE} & \textbf{COLUMN} & \textbf{PERTURBED\_TABLE} & \textbf{PERTURBED\_COLUMN} \\ \hline
\endhead
\hline
\endfoot
\endlastfoot       
\multirow{6}{*}{CHARTEVENTS} & ROW\_ID & \multirow{6}{*}{VITALS\_TIMESERIES} & ID \\* \cline{2-2}\cline{4-4}
 & ITEMID &  & ITEM\_ID \\* \cline{2-2}\cline{4-4}
 & CHARTTIME &  & OBSERVED\_AT \\* \cline{2-2}\cline{4-4}
 & VALUE &  & VALUE\_TEXT \\* \cline{2-2}\cline{4-4}
 & VALUENUM &  & VALUE\_NUM \\* \cline{2-2}\cline{4-4}
 & VALUEUOM &  & VALUE\_UNIT \\ \hline
\multirow{9}{*}{MICROBIOLOGYEVENTS} & ROW\_ID & \multirow{9}{*}{MICRO\_RESULTS} & ID \\* \cline{2-2}\cline{4-4}
 & CHARTTIME &  & COLLECTED\_AT \\* \cline{2-2}\cline{4-4}
 & SPEC\_ITEMID &  & SPECIMEN\_ITEM\_ID \\* \cline{2-2}\cline{4-4}
 & SPEC\_TYPE\_DESC &  & SPECIMEN\_TYPE \\* \cline{2-2}\cline{4-4}
 & ORG\_ITEMID &  & ORGANISM\_ITEM\_ID \\* \cline{2-2}\cline{4-4}
 & ORG\_NAME &  & ORGANISM\_NAME \\* \cline{2-2}\cline{4-4}
 & AB\_ITEMID &  & ANTIBIOTIC\_ITEM\_ID \\* \cline{2-2}\cline{4-4}
 & AB\_NAME &  & ANTIBIOTIC\_NAME \\* \cline{2-2}\cline{4-4}
 & INTERPRETATION &  & SUSCEPTIBILITY \\ \hline
\multirow{4}{*}{D\_ICD\_DIAGNOSIS} & ROW\_ID & \multirow{4}{*}{ICD\_DIAGNOSIS\_REF} & ID \\* \cline{2-2}\cline{4-4}
 & ICD9\_CODE &  & ICD9CODE \\* \cline{2-2}\cline{4-4}
 & SHORT\_TITLE &  & DX\_SHORT\_NAME \\* \cline{2-2}\cline{4-4}
 & LONG\_TITLE &  & DX\_LONG\_NAME \\ \hline
\multirow{4}{*}{D\_ICD\_PROCEDURES} & ROW\_ID & \multirow{4}{*}{ICD\_PROCEDURE\_REF} & ID \\* \cline{2-2}\cline{4-4}
 & ICD9\_CODE &  & ICD9CODE \\* \cline{2-2}\cline{4-4}
 & SHORT\_TITLE &  & PROC\_SHORT\_NAME \\* \cline{2-2}\cline{4-4}
 & LONG\_TITLE &  & PROC\_LONG\_NAME \\ \hline
\multirow{2}{*}{DIAGNOSIS\_ICD} & ROW\_ID & \multirow{2}{*}{PATIENT\_DIAGNOSES} & ID \\* \cline{2-2}\cline{4-4}
 & ICD9\_CODE &  & ICD\_CODE \\ \hline
\multirow{2}{*}{PROCEDURE\_ICD} & ROW\_ID & \multirow{2}{*}{PATIENT\_PROCEDURES} & ID \\* \cline{2-2}\cline{4-4}
 & ICD9\_CODE &  & ICD\_CODE \\ \hline
\multirow{7}{*}{LABEVENTS} & ROW\_ID & \multirow{7}{*}{LAB\_RESULTS} & ID \\* \cline{2-2}\cline{4-4}
 & ITEMID &  & LAB\_ITEM\_ID \\* \cline{2-2}\cline{4-4}
 & CHARTTIME &  & MEASURED\_AT \\* \cline{2-2}\cline{4-4}
 & VALUE &  & VALUE\_TEXT \\* \cline{2-2}\cline{4-4}
 & VALUENUM &  & VALUE\_NUM \\* \cline{2-2}\cline{4-4}
 & VALUEUOM &  & VALUE\_UNIT \\* \cline{2-2}\cline{4-4}
 & FLAG &  & ABNORMAL\_FLAG \\ \hline
\multirow{4}{*}{D\_LABITEMS} & ROW\_ID & \multirow{4}{*}{LAB\_TEST\_REF} & ID \\* \cline{2-2}\cline{4-4}
 & ITEMID &  & LAB\_ITEM\_ID \\* \cline{2-2}\cline{4-4}
 & LABEL &  & LAB\_TEST\_NAME \\* \cline{2-2}\cline{4-4}
 & FLUID &  & SPECIMEN\_FLUID \\ \hline
\multirow{2}{*}{D\_ITEMS} & ROW\_ID & \multirow{2}{*}{CLINICAL\_ITEM\_REF} & ID \\* \cline{2-2}\cline{4-4}
 & ITEMID &  & ITEM\_ID \\ \hline
\multirow{5}{*}{PROCEDUREEVENTS\_MV} & ROW\_ID & \multirow{5}{*}{PROCEDURES\_MV\_LOG} & ID \\* \cline{2-2}\cline{4-4}
 & STARTTIME &  & START\_AT \\* \cline{2-2}\cline{4-4}
 & ENDTIME &  & END\_AT \\* \cline{2-2}\cline{4-4}
 & ITEMID &  & ITEM\_ID \\* \cline{2-2}\cline{4-4}
 & LOCATION &  & PROCEDURE\_LOCATION \\ \hline
\multirow{10}{*}{PRESCRIPTIONS} & ROW\_ID & \multirow{10}{*}{MEDICATION\_ORDERS} & ID \\* \cline{2-2}\cline{4-4}
 & STARTDATE &  & START\_AT \\* \cline{2-2}\cline{4-4}
 & ENDDATE &  & END\_AT \\* \cline{2-2}\cline{4-4}
 & DRUG &  & DRUG\_NAME \\* \cline{2-2}\cline{4-4}
 & PROD\_STRENGTH &  & PRODUCT\_STRENGTH \\* \cline{2-2}\cline{4-4}
 & DOSE\_VAL\_RX &  & DOSE\_VALUE \\* \cline{2-2}\cline{4-4}
 & DOSE\_UNIT\_RX &  & DOSE\_UNIT \\* \cline{2-2}\cline{4-4}
 & FORM\_VAL\_DISP &  & DISPENSE\_VALUE \\* \cline{2-2}\cline{4-4}
 & FORM\_UNIT\_DISP &  & DISPENSE\_UNIT \\* \cline{2-2}\cline{4-4}
 & ROUTE &  & ADMIN\_ROUTE \\ \hline
\multirow{10}{*}{INPUTEVENTS\_MV} & ROW\_ID & \multirow{10}{*}{FLUID\_INPUTS\_MV} & ID \\* \cline{2-2}\cline{4-4}
 & ITEMID &  & ITEM\_ID \\* \cline{2-2}\cline{4-4}
 & STARTTIME &  & START\_AT \\* \cline{2-2}\cline{4-4}
 & ENDTIME &  & END\_AT \\* \cline{2-2}\cline{4-4}
 & AMOUNT &  & AMOUNT\_VALUE \\* \cline{2-2}\cline{4-4}
 & AMOUNTUOM &  & AMOUNT\_UNIT \\* \cline{2-2}\cline{4-4}
 & RATE &  & RATE\_VALUE \\* \cline{2-2}\cline{4-4}
 & RATEUOM &  & RATE\_UNIT \\* \cline{2-2}\cline{4-4}
 & ORIGINALAMOUNT &  & ORIG\_AMOUNT\_VALUE \\* \cline{2-2}\cline{4-4}
 & ORIGINALRATE &  & ORIG\_RATE\_VALUE \\ \hline
\multirow{12}{*}{INPUTEVENTS\_CV} & ROW\_ID & \multirow{12}{*}{FLUID\_INPUTS\_CV} & ID \\* \cline{2-2}\cline{4-4}
 & ITEMID &  & ITEM\_ID \\* \cline{2-2}\cline{4-4}
 & CHARTTIME &  & CHART\_AT \\* \cline{2-2}\cline{4-4}
 & AMOUNT &  & AMOUNT\_VALUE \\* \cline{2-2}\cline{4-4}
 & AMOUNTUOM &  & AMOUNT\_UNIT \\* \cline{2-2}\cline{4-4}
 & RATE &  & RATE\_VALUE \\* \cline{2-2}\cline{4-4}
 & RATEUOM &  & RATE\_UNIT \\* \cline{2-2}\cline{4-4}
 & ORIGINALAMOUNT &  & ORIG\_AMOUNT\_VALUE \\* \cline{2-2}\cline{4-4}
 & ORIGINALAMOUNTUOM &  & ORIG\_AMOUNT\_UNIT \\* \cline{2-2}\cline{4-4}
 & ORIGINALROUTE &  & ORIG\_ROUTE \\* \cline{2-2}\cline{4-4}
 & ORIGINALRATE &  & ORIG\_RATE\_VALUE \\* \cline{2-2}\cline{4-4}
 & ORIGINALRATEUOM &  & ORIG\_RATE\_UNIT \\ \hline
\multirow{5}{*}{OUTPUTEVENTS} & ROW\_ID & \multirow{5}{*}{OUTPUT\_FLUIDS} & ID \\* \cline{2-2}\cline{4-4}
 & ITEMID &  & ITEM\_ID \\* \cline{2-2}\cline{4-4}
 & CHARTTIME &  & RECORDED\_AT \\* \cline{2-2}\cline{4-4}
 & VALUE &  & VALUE \\* \cline{2-2}\cline{4-4}
 & VALUEUOM &  & VALUE\_UNIT \\ \hline
\multirow{2}{*}{DIAGNOSES\_ICD} & ROW\_ID & \multirow{2}{*}{DIAGNOSIS\_ICD\_CODES} & ID \\* \cline{2-2}\cline{4-4}
 & ICD9\_CODE &  & ICD9CODE \\ \hline
\multirow{2}{*}{PROCEDURES\_ICD} & ROW\_ID & \multirow{2}{*}{PROCEDURE\_ICD\_CODES} & ID \\* \cline{2-2}\cline{4-4}
 & ICD9\_CODE &  & ICD9CODE \\ \hline
\multirow{5}{*}{D\_LABITEMS} & ROW\_ID & \multirow{5}{*}{LAB\_TEST\_INFO} & ID \\* \cline{2-2}\cline{4-4}
 & ITEMID &  & ITEM\_ID \\* \cline{2-2}\cline{4-4}
 & LABEL &  & TEST\_NAME \\* \cline{2-2}\cline{4-4}
 & FLUID &  & SAMPLE\_FLUID \\* \cline{2-2}\cline{4-4}
 & CATEGORY &  & TEST\_CATEGORY \\ \hline
\multirow{10}{*}{D\_ITEMS} & ROW\_ID & \multirow{10}{*}{ITEM\_INFORMATION} & ID \\* \cline{2-2}\cline{4-4}
 & ITEMID &  & ITEM\_ID \\* \cline{2-2}\cline{4-4}
 & LABEL &  & ITEM\_NAME \\* \cline{2-2}\cline{4-4}
 & ABBREVIATION &  & ITEM\_ABBREVIATION \\* \cline{2-2}\cline{4-4}
 & DBSOURCE &  & DATASOURCE \\* \cline{2-2}\cline{4-4}
 & LINKSTO &  & LINKED\_TABLE \\* \cline{2-2}\cline{4-4}
 & CATEGORY &  & ITEM\_CATEGORY \\* \cline{2-2}\cline{4-4}
 & UNITNAME &  & UNIT\_NAME \\* \cline{2-2}\cline{4-4}
 & PARAM\_TYPE &  & PARAMETER\_TYPE \\* \cline{2-2}\cline{4-4}
 & CONCEPTID &  & CONCEPT\_ID \\ \hline
\end{longtable}
\end{center}

\section{Sample data of \dtname}
\label{app:sample data}
\begin{figure*}[t]
    \centering
    \includegraphics[width=\textwidth]{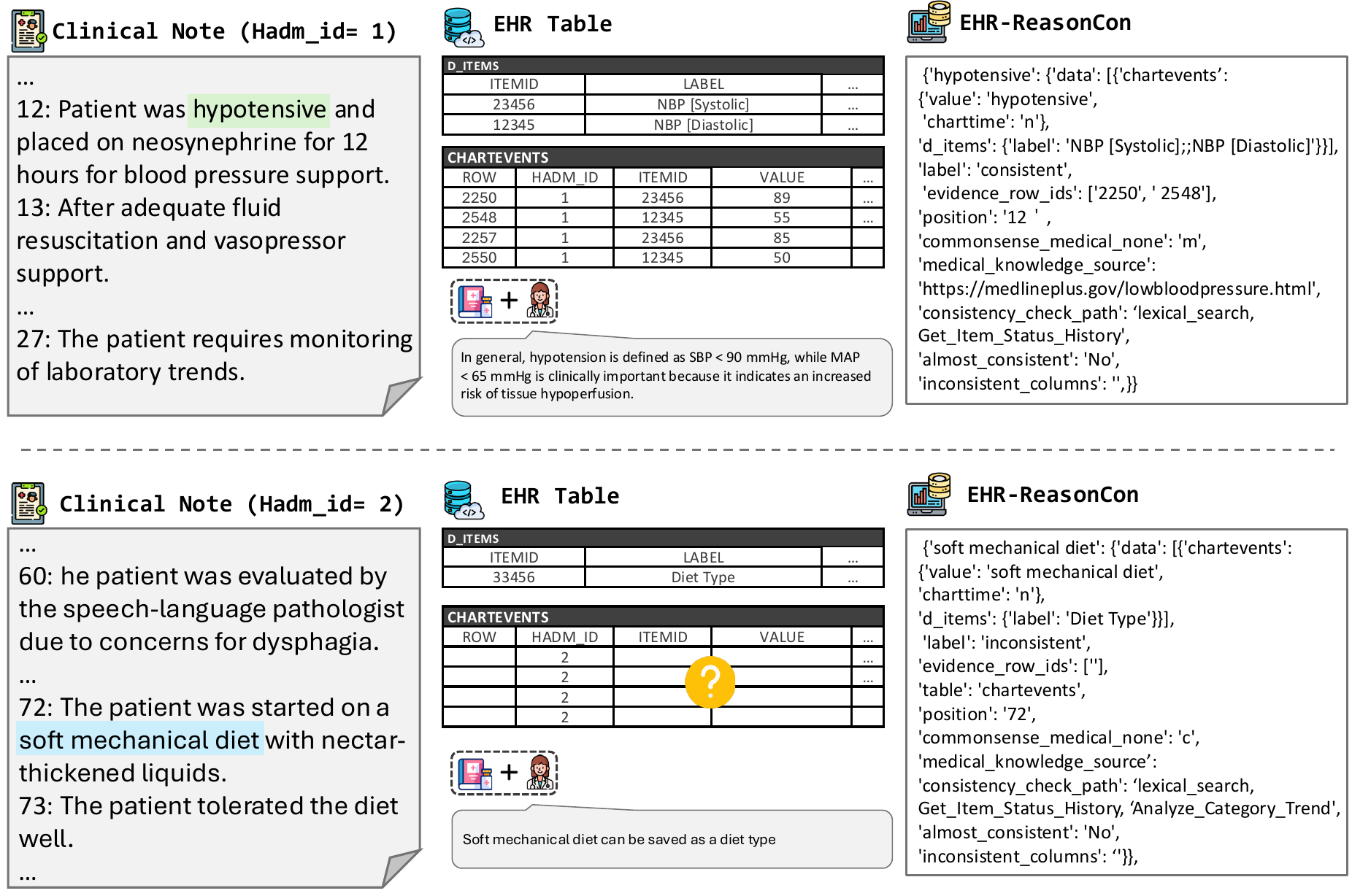}
    \vspace{-2mm}
    \caption{Sample cases from \dtname. The \textit{Hypotensive} example illustrates a consistent case, whereas the \textit{soft mechanical diet} example demonstrates an inconsistent case caused by omission. \texttt{Commonsense\_medical\_none} indicates whether consistency verification relied on commonsense reasoning (c), medical knowledge (m), or no deep reasoning. When medical knowledge is used, the corresponding reference is recorded in the \texttt{medical\_knowledge\_source} field. The \texttt{consistency\_check\_path} denotes the annotator-labeled reasoning trace, and \texttt{evidence\_row\_ids} records the row IDs in the structured tables that support the consistency or inconsistency judgment. To prevent patient identification, some content in the notes and tables was added, removed, or modified.
}
    \vspace{-2mm}
    \label{fig:data_Sample}
\end{figure*}

The sample data of \dtname is described in Figure~\ref{fig:data_Sample}.

\section{Experiment Setup Details}
\label{exp_setup_detail}
\subsection{Dataset Resources}
\begin{itemize}[leftmargin=1.3em, itemsep=2pt]
\item \textbf{MIMIC III~\cite{PhysioNet-mimiciii-1.4}}
\begin{itemize}[leftmargin=1.3em, itemsep=2pt]
\item URL: https://physionet.org/content/mimiciii/1.4/
\item License: PhysioNet Credentialed Health Data License 1.5.0
\end{itemize}
\item \textbf{MIMIC IV~\cite{johnson2020mimic}}
\begin{itemize}[leftmargin=1.3em, itemsep=2pt]
\item URL: https://physionet.org/content/mimiciv/3.1/
\item License: PhysioNet Credentialed Health Data License 1.5.0
\end{itemize}
\item \textbf{EHRCon~\cite{kwon2024ehrcon}}
\begin{itemize}[leftmargin=1.3em, itemsep=2pt]
\item URL: https://physionet.org/content/ehrcon-consistency-of-notes/1.0.0/
\item License: PhysioNet Credentialed Health Data License 1.5.0
\end{itemize}
\end{itemize}

\subsection{Model Resources}
This study categorizes the evaluated models into two primary types: proprietary and open-source. The proprietary lineup consists of Gemini-2.5-Pro, Gemini-2.5-Flash, and GPT-5. Each was accessed via its respective provider's API and used strictly within their given terms of service. For all experiments involving GPT-5 and Gemini 2.5, GPT-5 was accessed via HIPAA-compliant Azure OpenAI services, and Gemini 2.5 models were accessed via Google Vertex AI.
Table~\ref{tab:closed_source_cost} summarizes the estimated average inference cost per note, in U.S. dollars, for each proprietary model. The specific model versions are as follows:

\begin{itemize}[leftmargin=1.3em, itemsep=2pt]
\item \textbf{Gemini-2.5-Flash~\cite{comanici2025gemini}}
\item \textbf{Gemini-2.5-Pro~\cite{comanici2025gemini}}
\item \textbf{GPT-5~\cite{singh2025openai}}
\end{itemize}

\begin{table}[t]
\centering
\caption{Estimated inference costs for proprietary models.}
\label{tab:closed_source_cost}
\begin{tabular}{l l r}
\toprule
Model & Task & Average Cost per Note \\
\midrule
Gemini-2.5-Flash   & \frname & \$2.5 \\
Gemini-2.5-Pro & Evaluation & \$0.42  \\
GPT-5        & Evaluation & \$0.48 \\
\bottomrule
\end{tabular}
\end{table}

The open-source models were obtained from Hugging Face~\cite{wolf-etal-2020-transformers}. The corresponding Hugging Face repositories and licenses for each model are listed below:
\begin{itemize}[leftmargin=1.3em, itemsep=2pt]
\item \textbf{Qwen3-32B~\cite{yang2025qwen3}}
\begin{itemize}[leftmargin=1.3em, itemsep=2pt]
\item Hugging Face Path: Qwen/Qwen3-32B
\item License: Apache license 2.0 
\end{itemize}
\item \textbf{GPT-OSS 20B~\cite{agarwal2025gpt}}
\begin{itemize}[leftmargin=1.3em, itemsep=2pt]
\item Hugging Face Path: openai/gpt-oss-20b
\item License: Apache license 2.0
\end{itemize}
\item \textbf{MedGemma 27B~\cite{sellergren2025medgemma}}
\begin{itemize}[leftmargin=1.3em, itemsep=2pt]
\item Hugging Face Path: google/medgemma-27b-it
\item License: health-ai-developer-foundations
\end{itemize}
\end{itemize}

\subsection{Experiment Details}
\paragraph{Dataset}~We split the 105 clinical notes into 83 for the test set and 22 for the validation set. The main experiments were conducted on the test set, and the validation set was used to develop \frname. For MIMIC-IV, we additionally annotated seven discharge summary notes and used them in our experiments.

\paragraph{Compute Requirements and Hyperparameters}
On average, open-source models required approximately 9 hours per note to complete our task, while Gemini-2.5-Flash required approximately 3 hours per note. This is substantially shorter than human annotation, which required approximately 12 hours per note on average, even when performed by annotators familiar with EHRs. For model-based evaluation, Gemini-2.5-Pro required approximately 1 hour per note, and GPT-5 also required approximately 1 hour per note on average. For open-source models, inference was performed on a single NVIDIA A100 GPU with a maximum generation length of 2048 tokens, temperature set to 0.5, and top-p set to 0.95. Gemini models were run using the default inference settings.

\paragraph{Additional NER Models and Fine-tuning Details.}~Additional NER models were used for the entity extraction ablation study. Specifically, the BERT Ensemble baseline aggregates predictions from biomedical BERT-based NER models, while the Trained NER baseline fine-tunes MedGemma 27B on note--entity pairs from \dtname. The additional model resources, fine-tuning hyperparameters are as follows:
\begin{itemize}[leftmargin=1.3em, itemsep=2pt]
\item \textbf{Clinical Bert Model}
\begin{itemize}[leftmargin=1.3em, itemsep=2pt]
\item Hugging Face Path: blaze999/clinical-ner
\item License: mit
\end{itemize}
\item \textbf{Spacy}
\begin{itemize}[leftmargin=1.3em, itemsep=2pt]
\item https://spacy.io/
\end{itemize}
\item \textbf{Stanza}
\begin{itemize}[leftmargin=1.3em, itemsep=2pt]
\item https://stanfordnlp.github.io/stanza/
\end{itemize}
\item \textbf{MedGemma 27B~\cite{sellergren2025medgemma}}
\begin{itemize}[leftmargin=1.3em, itemsep=2pt]
\item Hugging Face Path: google/medgemma-27b-it
\item Hyperparameter:
\begin{itemize}[leftmargin=1.3em, itemsep=2pt]
\item Epoch: 1
\item Learning Rate: $2\times10^{-5}$
\item Batch Size: 8
\item Cosine learning-rate scheduling with a warmup ratio of 0.03
\item Maximum sequence length of 2048 tokens
\end{itemize}
\end{itemize}
\end{itemize}

\section{Author statement}
\label{app:statment}
Any legal or ethical issues, including potential rights violations related to \dtname, are the sole responsibility of the authors.




\newpage
\section*{NeurIPS Paper Checklist}


\begin{enumerate}

\item {\bf Claims}
    \item[] Question: Do the main claims made in the abstract and introduction accurately reflect the paper's contributions and scope?
    \item[] Answer: \answerYes{} 
    \item[] Justification: As discussed in Section~\ref{intro}, the claims presented in the abstract and introduction faithfully represent the paper’s contributions and scope, clearly describing the research aims, methodology, and findings.

\item {\bf Limitations}
    \item[] Question: Does the paper discuss the limitations of the work performed by the authors?
    \item[] Answer: \answerYes{} 
    \item[] Justification: The paper addresses its limitations in the conclusion and limitations section (Section~\ref{limitation}).

\item {\bf Theory assumptions and proofs}
    \item[] Question: For each theoretical result, does the paper provide the full set of assumptions and a complete (and correct) proof?
    \item[] Answer: \answerNA{} 
    \item[] Justification: This paper focuses on benchmark and framework construction, along with empirical evaluation, and does not present theoretical results or formal proofs.

    \item {\bf Experimental result reproducibility}
    \item[] Question: Does the paper fully disclose all the information needed to reproduce the main experimental results of the paper to the extent that it affects the main claims and/or conclusions of the paper (regardless of whether the code and data are provided or not)?
    \item[] Answer: \answerYes{} 
    \item[] Justification: Appendix~\ref{exp_setup_detail} provides the model paths, computational resources, and detailed experimental settings required to reproduce the main results. It also includes links to the experimental code, the proposed framework, and related resources.

\item {\bf Open access to data and code}
    \item[] Question: Does the paper provide open access to the data and code, with sufficient instructions to faithfully reproduce the main experimental results, as described in supplemental material?
    \item[] Answer: \answerYes{} 
    \item[] Justification: The EHR-ReasonCon dataset (v1.0.0) is available on PhysioNet~\url{https://physionet.org/content/ehr-reasoncon/1.0.0/}, and the implementation of our proposed EHR-INSPECTOR framework is publicly available on GitHub~\url{https://github.com/dustn1259/EHR-ReasonCon}. Since the dataset is derived from MIMIC-III, access to EHR-ReasonCon and MIMIC-III requires PhysioNet credentialing and completion of the corresponding data use agreements. The GitHub repository provides detailed instructions for environment setup, dependency installation, data preparation, and execution of the three-stage pipeline and evaluation, enabling reproduction of the proposed method's experimental results.

\item {\bf Experimental setting/details}
    \item[] Question: Does the paper specify all the training and test details (e.g., data splits, hyperparameters, how they were chosen, type of optimizer) necessary to understand the results?
    \item[] Answer: \answerYes{} 
    \item[] Justification: Detailed information about the experimental setup can be found in Appendix~\ref{exp_setup_detail}.

\item {\bf Experiment statistical significance}
    \item[] Question: Does the paper report error bars suitably and correctly defined or other appropriate information about the statistical significance of the experiments?
    \item[] Answer: \answerYes{} 
    \item[] Justification: We statistically evaluated the framework through three validation-set runs of \dtname using \frname, with each run independently assessed three times by the LLM-as-a-judge evaluator; the mean and standard deviation results are provided in Appendix~\ref{app:statistical_eval}.

\item {\bf Experiments compute resources}
    \item[] Question: For each experiment, does the paper provide sufficient information on the computer resources (type of compute workers, memory, time of execution) needed to reproduce the experiments?
    \item[] Answer: \answerYes{} 
    \item[] Justification: Details on the experimental compute resources can be found in Appendix~\ref{exp_setup_detail}.
    
\item {\bf Code of ethics}
    \item[] Question: Does the research conducted in the paper conform, in every respect, with the NeurIPS Code of Ethics \url{https://neurips.cc/public/EthicsGuidelines}?
    \item[] Answer: \answerYes{} 
    \item[] Justification: The research presented in this paper complies with the NeurIPS Code of Ethics.

\item {\bf Broader impacts}
    \item[] Question: Does the paper discuss both potential positive societal impacts and negative societal impacts of the work performed?
    \item[] Answer: \answerNA{} 
    \item[] Justification: Our dataset is derived exclusively from MIMIC-III and MIMIC-IV, both of which are de-identified and distributed under the PhysioNet credentialed-access license. Because these data have been anonymized and are publicly available to approved users, we do not identify direct negative societal impacts arising from their use.
    
\item {\bf Safeguards}
    \item[] Question: Does the paper describe safeguards that have been put in place for responsible release of data or models that have a high risk for misuse (e.g., pre-trained language models, image generators, or scraped datasets)?
    \item[] Answer: \answerNA{} 
    \item[] Justification: This research is based on the MIMIC-III and MIMIC-IV datasets, which have been fully de-identified and are distributed through PhysioNet in accordance with its data-use requirements. Given that the study does not involve sensitive misuse-prone data collection methods, such as web scraping, nor generative or otherwise high-risk models, we do not deem further protective measures to be required.

\item {\bf Licenses for existing assets}
    \item[] Question: Are the creators or original owners of assets (e.g., code, data, models), used in the paper, properly credited and are the license and terms of use explicitly mentioned and properly respected?
    \item[] Answer: \answerYes{} 
    \item[] Justification: All datasets and models used in this paper have been appropriately cited.

\item {\bf New assets}
    \item[] Question: Are new assets introduced in the paper well documented and is the documentation provided alongside the assets?
    \item[] Answer: \answerYes{} 
    \item[] Justification: Detailed descriptions of the dataset construction process are provided in Section~\ref{sec3} and Appendix~\ref{app:annotation_protocol}.

\item {\bf Crowdsourcing and research with human subjects}
    \item[] Question: For crowdsourcing experiments and research with human subjects, does the paper include the full text of instructions given to participants and screenshots, if applicable, as well as details about compensation (if any)? 
    \item[] Answer: \answerYes{} 
    \item[] Justification: The annotation protocol followed by the human annotators was developed in consultation with practitioners and is described in Appendix ~\ref{app:annotation_protocol}, which also includes a screenshot of the annotation tool used during the annotation process.

\item {\bf Institutional review board (IRB) approvals or equivalent for research with human subjects}
    \item[] Question: Does the paper describe potential risks incurred by study participants, whether such risks were disclosed to the subjects, and whether Institutional Review Board (IRB) approvals (or an equivalent approval/review based on the requirements of your country or institution) were obtained?
    \item[] Answer: \answerNA{} 
    \item[] Justification: This study uses MIMIC-III and MIMIC-IV, both of which were approved by the Institutional Review Boards (IRBs) of Beth Israel Deaconess Medical Center in Boston, MA, and the Massachusetts Institute of Technology in Cambridge, MA.

\item {\bf Declaration of LLM usage}
    \item[] Question: Does the paper describe the usage of LLMs if it is an important, original, or non-standard component of the core methods in this research? Note that if the LLM is used only for writing, editing, or formatting purposes and does \emph{not} impact the core methodology, scientific rigor, or originality of the research, declaration is not required.
    \item[] Answer: \answerYes{} 
    \item[] Justification: The LLM-based framework proposed in this study, \frname, consists of an LLM-based pipeline. A detailed description is provided in Section~\ref{sec4} and Appendix~\ref{app:detail_framework}.

\end{enumerate}

\end{document}